%% file: main.tex
\pdfoutput=1
\documentclass[10pt]{article} %

\usepackage[preprint]{tmlr}

\makeatletter
\@ifundefined{mathdefault}{}{}
\@ifundefined{mathit}{\def\mathit#1{\mathnormal{#1}}}{}
\@ifundefined{textminus}{}{}
\makeatother
\usepackage{algorithm}
\usepackage{algpseudocode}
\usepackage[utf8]{inputenc} %
\usepackage[T1]{fontenc}    %
\usepackage{hyperref}       %
\usepackage{url}            %
\usepackage{booktabs}       %
\usepackage{nicefrac}       %
\usepackage{microtype}      %
\usepackage{xcolor}         %
\usepackage{tcolorbox}
\usepackage{graphicx}
\usepackage{subcaption}
\usepackage{caption}
\usepackage{tabularx,booktabs}
\usepackage{comment}
\usepackage{amsthm} %
\usepackage{wrapfig}
\newtheorem{theorem}{Theorem}
\newtheorem{lemma}{Lemma}
\newtheorem{proposition}{Proposition}
\newtheorem{corollary}{Corollary}
\newtheorem{remark}{Remark}
\usepackage[nolist]{acronym}
\usepackage{float}
\usepackage{color}
\usepackage{multicol}
\usepackage{multirow}
\usepackage{amsmath,amssymb}
\usepackage{amsfonts}
\usepackage{textcomp}
\usepackage{psfrag}
\usepackage{multimedia}
\usepackage{fancybox}
\usepackage{mathtools}
\usepackage{pgfplots}    %
\usepackage{adjustbox}
\usepackage{tikz}
\usepackage{amsmath}
\newtheorem{assumption}{Assumption}

\allowdisplaybreaks
\usepackage{standalone} %
\usetikzlibrary{positioning, arrows, shapes.geometric}
\usetikzlibrary{arrows.meta, positioning, shapes}
\usetikzlibrary{calc, fit, backgrounds}

\title{Gradient-Based Co-Design of Systems and Reinforcement Learning Policies with Differentiable Simulators}

\author{\name Akhil S.~Anand \email akhil.s.anand@ntnu.no \\
      \addr Department of Engineering Cybernetics \\
      Norwegian University of Science and Technology (NTNU) \\
      Trondheim, Norway
      \AND
      \name Xun Shen \\
      \addr Graduate School of Engineering \\
      Tokyo University of Agriculture and Technology \\
      Tokyo, Japan
      \AND
      \name Marc Peter Deisenroth \\
      \addr UCL Centre for Artificial Intelligence \\
      University College London \\
      London, United Kingdom
      \AND
      \name Sebastien Gros \\
      \addr Department of Engineering Cybernetics \\
      Norwegian University of Science and Technology (NTNU) \\
      Trondheim, Norway
      }

\begin{document}
\input{acronyms.tex}

\maketitle  %

\begin{abstract}

Co-design is the problem of jointly optimizing the design of a system and the policy that operates it. Neither can be chosen independently: a system design is only as good as the policy trained for it, and the optimal policy changes with the design. When that policy is learned by reinforcement learning (RL), evaluating a candidate design requires a full policy training run, and the design is usually optimized by black-box search such as Bayesian optimization (BO) or evolution strategies (ES). These methods scale poorly with the design dimension and ignore the differentiable structure of modern simulators and, in particular, the sensitivity of the optimal policy with respect to the design. We instead derive a gradient-based co-design method, where the design gradient decomposes into a direct effect of changing the design at a fixed policy and an indirect effect through the design-dependent optimal policy. The indirect effect requires the sensitivity of the converged policy to the design, which we obtain by applying the implicit function theorem to the stationarity condition of the policy gradient. Based on this sensitivity, we develop a proof-of-concept co-design algorithm with stochastic value gradient (SVG)-style policy optimization, prove that it converges to an approximately stationary design, and demonstrate on a differentiable networked-system benchmark with high-dimensional design spaces that it consistently outperforms BO and ES.
\end{abstract}

\input{sections/introduction}

\input{figures/fig_overview}
\input{sections/related_work}
\input{sections/background}

\input{sections/codesign_grad}
\input{sections/shac_sensitivity}
\input{sections/example}

\input{sections/discussion}
\input{sections/conclusion}

\makeatletter
\if@accepted
\subsubsection*{Acknowledgments}
This work was supported by the Research Council of Norway through the Norwegian Centre on AI for Decisions (\href{https://aid-centre.no}{aid-centre.no}) under Grant 357485.
\fi
\makeatother

\bibliography{mbrl}
\bibliographystyle{tmlr}

\appendix
\input{sections/appendix_codesign_grad}
\input{sections/appendix_shac_sensitivity}
\input{sections/convergence_analysis}
\input{sections/appendix_implementation}
\input{sections/appendix_example}

\end{document}

%% file: acronyms.tex
\begin{acronym}
    \acro{sdm}[SDM]{Sequential Decision Making}
   \acro{mdp}[MDP]{Markov Decision Process}
   \acro{rl}[RL]{Reinforcement Learning}
   \acro{sl}[SL]{Supervised Learning}
   \acro{ssl}[SSL]{Self-Supervised Learning}
   \acro{dp}[DP]{Dynamic Programming}
   \acro{adp}[ADP]{Approximate Dynamic Programming}
   \acro{sp}[SP]{Stochastic Programming}
   \acro{lp}[LP]{Linear Programming}
   \acro{mpc}[MPC]{Model Predictive Control}
   \acro{empc}[EMPC]{Economic Model Predictive Control}
   \acro{td}[TD]{Temporal Difference}
   \acro{dpg}[DPG]{Deterministic Policy Gradient}
   \acro{spg}[SPG]{Stochastic Policy Gradient}
   \acro{mbrl}[MBRL]{Model-based Reinforcement Learning}
   \acro{dt}[DT]{Digital Twin}
   \acro{ml}[ML]{Machine Learning}
   \acro{lstm}[LSTM]{Long Short-Term Memory}
   \acro{rnn}[RNN]{Recurrent Neural Network }
   \acro{dnn}[DNN]{Deep Neural Network }
   \acro{dnns}[DNNs]{Deep Neural Networks}
   \acro{gp}[GP]{Gaussian Processes}
   \acro{ai}[AI]{Artificial Intelligence}
   \acro{mle}[MLE]{Maximum Likelihood Estimation}
   \acro{lqr}[LQR]{Linear Quadratic Regulator}
   \acro{gpu}[GPU]{Graphics processing unit}
   \acro{tpu}[TPU]{Tensor processing unit}
   \acro{gans}[GANs]{Generative Adversarial Networks}
   \acro{cnns}[CNNs]{Convolutional Neural Networks}
   \acro{ls}[LS]{Least Squares}
   \acro{mse}[MSE]{Mean Squared Error}
    \acro{map}[MAP]{Maximum A Posteriori}
    \acro{gmm}[GMM]{Gaussian Mixture Model}
    \acro{hmm}[HMM]{Hidden Markov Model}
    \acro{elbo}[ELBO]{Evidence Lower Bound}
    \acro{vae}[VAE]{Variational Autoencoder}
    \acro{gan}[GAN]{Generative Adversarial Network}
    \acro{ode}[ODE]{Ordinary Differential Equations}
    \acro{ude}[UDE]{Universal Differential Equations}
    \acro{ham}[HAM]{Hybrid Analysis and Modeling}
    \acro{vi}[VI]{Variational Inference}
    \acro{em}[EM]{Expectation-Maximization}
    \acro{lstdq}[LSTDQ]{Least Square Temporal Difference Q-learning}
    \acro{rldp}[RLDP]{Reinforcement Learning of Decision Processes}
    \acro{or}[OR]{Operations Research}
    \acro{mppi}[MPPI]{Model Predictive Path Integral}
    \acro{cem}[CEM]{Cross Entropy Method}
    \acro{ift}[IFT]{Implicit Function Theorem}
    \acro{ppo}[PPO]{Proximal Policy Optimization}
    \acro{sac}[SAC]{Soft Actor Critic}
    \acro{llm}[LLM]{Large Language Model}
    \acro{nn}[NN]{Neural Network}
    \acro{lr}[LR]{Likelihood-ratio Gradient}
    \acro{pd}[PD]{Pathwise Derivative}
     \acro{svg}[SVG]{Stochastic Value Gradients}
     \acro{ddpg}[DDPG]{Deep Deterministic Policy Gradient}
     \acro{td3}[TD3]{Twin Delayed Deep Deterministic Policy Gradient}
    \acro{shac}[SHAC]{Short-Horizon Actor-Critic}
    \acro{ad}[AD]{Automatic Differentiation}
    \acro{jvp}[JVP]{Jacobian-vector product}
    \acro{vjp}[VJP]{vector-Jacobian product}
    \acro{gmres}[GMRES]{Generalized Minimal Residual}
    \acro{bo}[BO]{Bayesian Optimization}
    \acro{es}[ES]{Evolution Strategy}
\end{acronym}

%% file: sections/introduction.tex
\section{Introduction}\label{sec:intro}

Co-design is the problem of jointly optimizing a system design and its control (decision) policy to achieve the best performance on a given task. It is a long-standing problem studied across domains including control theory, operations research, and robotics~\citep{herber2017nested,tobin1988sensitivity,cheney2014unshackling,gupta2021derl}. This problem arises whenever an operational system must be configured for a task, such as sizing the capacities of a supply network, expanding a power grid, or choosing the morphology of a robot. The quality of a design is strongly dependent on the performance of the policy optimized for that design. Hence, the design cannot be evaluated independently of the policy: changing the system design changes the environment on which the policy is trained and evaluated, and therefore also changes the optimized policy itself. This naturally leads to a bilevel formulation of co-design, in which the outer level optimizes the design and the inner level optimizes the policy. The same bilevel structure underlies meta-learning and hyperparameter optimization\citep{rajeswaran2019imaml,lorraine2020optimizing}. Co-design differs in that the design changes the environment itself, so it affects the outer level objective directly and not only through the policy. The common approach treats the outer level as black-box search with \ac{bo} or \ac{es}, wrapped around an inner policy optimization level~\citep{gupta2021derl,bhatia2021evolutiongym,schaff2019jointly}. While this can be effective in many cases, it ignores the structure of the problem, specifically the influence of changing design on the policy.

In this work, we consider that the system is represented by a differentiable simulator and the inner-level policy optimization is performed using \ac{rl}. This setting is increasingly relevant with the rise of differentiable simulators~\citep{hu2020difftaichi,freeman2021brax} and \ac{pd}-based \ac{rl} methods such as \ac{svg}~\citep{heess2015svg} and \ac{shac}~\citep{xu2022shac}, which exploit simulator derivatives for accelerated policy training. Differentiable simulators provide a differentiable structure for the policy-optimization problem, which can be leveraged for design optimization. Because the \ac{pd} estimator differentiates through the sampled action, the approach is restricted to continuous action space settings with a differentiable policy, and, analogously, a continuous design parameterization. Discrete action or design cases can often be relaxed through smooth surrogate parameterizations~\citep{jang2017categorical}. Effective co-design requires understanding how changes in the design affect the task objective directly, and how they affect it indirectly by changing the policy optimized for that design. Understanding this indirect effect requires sensitivity analysis of the underlying \ac{rl} scheme. Recent works have shown that such sensitivity analysis is possible for \ac{ppo}-based policy optimization, by applying implicit differentiation for sim-to-real \ac{rl}~\citep{anand2026bilevel} and for co-design~\citep{dai2026stackelberg}, the latter combined with a likelihood-ratio surrogate of the policy-gradient objective.

In this paper, we formulate co-design as a bilevel \ac{rl} problem, where the inner level optimizes a control policy and the outer level optimizes the design parameters. Both levels use \ac{pd}-based \ac{rl}, with \ac{shac} in our practical demonstrations. The key contributions of this work are: (i) we derive a sensitivity analysis for \ac{pd}-based \ac{rl} by applying the \ac{ift} to the stationarity condition of the optimized policy; (ii) we use this sensitivity to derive the outer-level co-design policy gradient and develop a proof-of-concept algorithm for co-design; (iii) we conduct a convergence analysis of the proposed algorithm, and (iv) we evaluate the proposed algorithm on illustrative examples with high-dimensional design spaces.

%% file: figures/fig_overview.tex
\providecommand{\codesigngradeq}{$\nabla_\theta J^{\mathrm{D}}=\underbrace{\textcolor{orange!80!black}{\partial_\theta J^{\mathrm{D}}}}_{\text{direct}}+\underbrace{\textcolor{blue!70!black}{\nabla_\phi J^{\mathrm{D}}\,\kappa_\theta}}_{\text{indirect}}$}
\begin{figure}[t]
\centering
\begin{minipage}[b]{0.68\textwidth}
\centering
\resizebox{\linewidth}{!}{%
\begin{tikzpicture}[>=Latex,
  box/.style={draw, rounded corners, align=center, inner sep=5pt, font=\small},
  env/.style={box, fill=black!4}, proc/.style={box, fill=blue!6},
  eval/.style={box, fill=black!4}, sens/.style={box, fill=blue!6, font=\footnotesize},
  gbox/.style={box, fill=red!4}]
\node[env] (env) at (0,0) {Differentiable environment\\$(\mathcal{P}_\theta,\,R_\theta)$};
\node[proc] (rl) at (0,-2.4) {Policy optimization\\(SVG)};
\node[eval] (jd) at (4.6,-1.0) {Policy eval.\\under $J^{\mathrm{D}}$};
\node[sens] (kappa) at (5.5,-2.7) {$\kappa_\theta=\nabla_\theta\phi^\star$ (IFT)};
\node[gbox] (grad) at (8.6,-1.0) {\codesigngradeq};
\draw[->] ([xshift=8mm]env.south) to[out=-60,in=60] node[right,font=\scriptsize]{$s_t,\,r_t$} ([xshift=8mm]rl.north);
\draw[->] ([xshift=-8mm]rl.north) to[out=120,in=-120] node[left,font=\scriptsize]{$a_t\!\sim\!\pi_\phi$} ([xshift=-8mm]env.south);
\draw[->] ([yshift=3mm]rl.east) to[out=0,in=180] node[pos=0.62,below right=-1pt,font=\scriptsize]{$\phi^\star(\theta)$} (jd.west);
\draw[->] (jd) -- (grad);
\draw[->,blue,dashed] (rl.east |- kappa.west) -- node[pos=0.5,below,font=\scriptsize]{$\nabla_\phi J^\pi\!=\!0$} (kappa.west);
\draw[->,blue] (kappa.east) to[out=0,in=-90] (grad.south);
\draw[->,thick,red!70!black,rounded corners=8pt] (grad.north) |- node[pos=0.75,above,font=\scriptsize,text=red!70!black]{design update (modifies $\mathcal{P}_\theta,\,R_\theta$)} (env.east);
\begin{scope}[shift={(-3.95,-1.3)}]
\draw[rounded corners, fill=black!2] (-1.7,-1.05) rectangle (1.7,0.95);
\coordinate (rollSW) at (-1.7,-1.05);
\coordinate (rollNE) at (1.7,0.95);
\node[font=\scriptsize] at (0,0.7) {sampled rollouts $\tau(\omega)$};
\draw[->,gray!70] (-1.45,-0.78) -- (1.5,-0.78) node[pos=0.97,above,font=\tiny,gray!70]{$t$};
\draw[->,gray!70] (-1.45,-0.78) -- (-1.45,0.42) node[pos=0.9,right,font=\tiny,gray!70]{$s_t$};
\draw[blue!80,thin] plot[smooth,tension=0.8] coordinates {(-1.4,-0.5)(-0.7,0.0)(0.1,0.2)(0.8,-0.05)(1.45,0.25)};
\draw[blue!70,thin] plot[smooth,tension=0.8] coordinates {(-1.4,-0.5)(-0.7,-0.2)(0.1,-0.35)(0.8,0.1)(1.45,-0.05)};
\draw[blue!60,thin] plot[smooth,tension=0.8] coordinates {(-1.4,-0.5)(-0.7,-0.4)(0.1,0.05)(0.8,-0.15)(1.45,0.15)};
\draw[blue!50,thin] plot[smooth,tension=0.8] coordinates {(-1.4,-0.5)(-0.7,-0.6)(0.1,-0.25)(0.8,-0.55)(1.45,-0.3)};
\draw[blue!75,thin] plot[smooth,tension=0.9] coordinates {(-1.4,-0.5)(-0.75,0.15)(0.0,-0.1)(0.7,0.3)(1.45,0.1)};
\draw[blue!65,thin] plot[smooth,tension=0.9] coordinates {(-1.4,-0.5)(-0.75,-0.3)(0.0,0.15)(0.7,-0.3)(1.45,0.0)};
\draw[blue!55,thin] plot[smooth,tension=0.9] coordinates {(-1.4,-0.5)(-0.75,-0.55)(0.0,-0.65)(0.7,-0.4)(1.45,-0.6)};
\draw[blue!45,thin] plot[smooth,tension=0.9] coordinates {(-1.4,-0.5)(-0.75,-0.15)(0.0,-0.5)(0.7,0.0)(1.45,-0.2)};
\draw[blue!35,thin] plot[smooth,tension=0.9] coordinates {(-1.4,-0.5)(-0.75,-0.7)(0.0,-0.15)(0.7,0.15)(1.45,-0.45)};
\fill[black] (-1.4,-0.5) circle (0.8pt);
\end{scope}
\begin{scope}[on background layer]
\node[fill=blue!4, draw=blue!25, rounded corners=5pt, inner sep=7pt,
      fit=(env)(rl)(rollSW)(rollNE)(env.north -| rollSW)(kappa.south -| rollSW)] (innerzone) {};
\node[fill=red!3, draw=red!20, rounded corners=5pt, inner sep=7pt,
      fit=(jd)(kappa)(grad)(jd.west |- env.north)(jd.west |- kappa.south)] (outerzone) {};
\end{scope}
\node[font=\scriptsize\itshape, text=blue!50!black, anchor=south west]
  at ([xshift=1mm,yshift=0.5mm]innerzone.south west) {inner level: policy optimization};
\node[font=\scriptsize\itshape, text=red!50!black, anchor=south east]
  at ([xshift=-1mm,yshift=0.5mm]outerzone.south east) {outer level: design gradient};
\end{tikzpicture}}\\[3pt]
{\small (a)}
\end{minipage}\hfill
\begin{minipage}[b]{0.28\textwidth}
\centering
\resizebox{\linewidth}{!}{%
\begin{tikzpicture}[>=Latex, yscale=0.78,
  lbl/.style={fill=white, fill opacity=0.85, text opacity=1, inner sep=1pt, rounded corners=1pt}]
\begin{scope}[shift={(3.9,3.05)}, rotate=25]
  \foreach \r/\shade in {2.0109/5, 1.51/10, 1.005/16, 0.50/23}
    \filldraw[fill=gray!\shade, draw=gray!45, line width=0.4pt] (0,0) ellipse ({\r} and {0.72*\r});
  \fill[gray!70] (0,0) circle (1.4pt);
\end{scope}
\node[gray!65!black, font=\footnotesize, anchor=west] at (4.6,1.25) {$J^{\mathrm{D}}$ contours};
\draw[->] (0,0)--(6.2,0) node[right, font=\small] {$\theta$};
\draw[->] (0,0)--(0,4.75) node[above, font=\small] {$\phi$};
\draw[gray!60, densely dotted] (2.4,0)--(2.4,1.8)--(0,1.8);
\node[font=\footnotesize, below] at (2.4,0) {$\theta_0$};
\node[font=\footnotesize, left]  at (0,1.8) {$\phi^\star(\theta_0)$};
\draw[blue, very thick] (0.2,0.75) .. controls (1.0,1.05) and (1.8,1.35) .. (2.4,1.8)
                                   .. controls (3.0,2.25) and (3.5,3.0) .. (3.85,4.2);
\node[lbl, text=blue, font=\small, anchor=south] at (3.85,4.22) {$\phi^\star(\theta)$};
\draw[gray!55, dashed, line width=0.4pt] (3.5,1.8)--(3.5,2.625)--(2.4,2.625);
\draw[->, orange!80!black, thick] (2.4,1.8) -- (3.5,1.8);
\draw[->, blue, thick] (2.4,1.8) -- (2.4,2.625);
\draw[->, densely dashed, thick] (2.4,1.8) -- (3.5,2.625);
\draw[gray!70] (2.56,1.8)--(2.56,1.96)--(2.4,1.96);
\fill (2.4,1.8) ellipse (1.8pt and 2.3pt);
\node[lbl, text=orange!80!black, font=\footnotesize, anchor=north] at (2.95,1.74) {$\partial_\theta J^{\mathrm{D}}$};
\node[lbl, text=blue, font=\footnotesize, anchor=east] at (2.34,2.22) {$\nabla_\phi J^{\mathrm{D}}\kappa_\theta$};
\node[lbl, font=\footnotesize, anchor=north west] at (3.02,2.2) {$\kappa_\theta$};
\end{tikzpicture}}\\[3pt]
{\small (b)}
\end{minipage}
\caption{\textbf{Gradient-based co-design.} \textbf{(a)}~The co-design loop: the
policy optimization loop (environment $\leftrightarrow$ policy) yields
the converged policy $\phi^\star(\theta)$, evaluated under the design objective $J^{\mathrm{D}}$; the
design gradient, a \textcolor{orange!80!black}{direct} effect plus an \textcolor{blue!70!black}{indirect}
policy-sensitivity effect through $\kappa_\theta=\nabla_\theta\phi^\star(\theta)$, used by another \ac{rl} agent to update the design parameters.
\textbf{(b)}~Moving along the optimal policy manifold $\phi^\star(\theta)$ (slope $\kappa_\theta$)
decomposes the total design gradient $\nabla_\theta J^{\mathrm{D}}$ into a \textcolor{orange!80!black}{direct}
step at fixed policy ($\partial_\theta J^{\mathrm{D}}$) and an \textcolor{blue!70!black}{indirect} step from
the induced policy shift ($\nabla_\phi J^{\mathrm{D}}\kappa_\theta$).}
\label{fig:overview}
\end{figure}
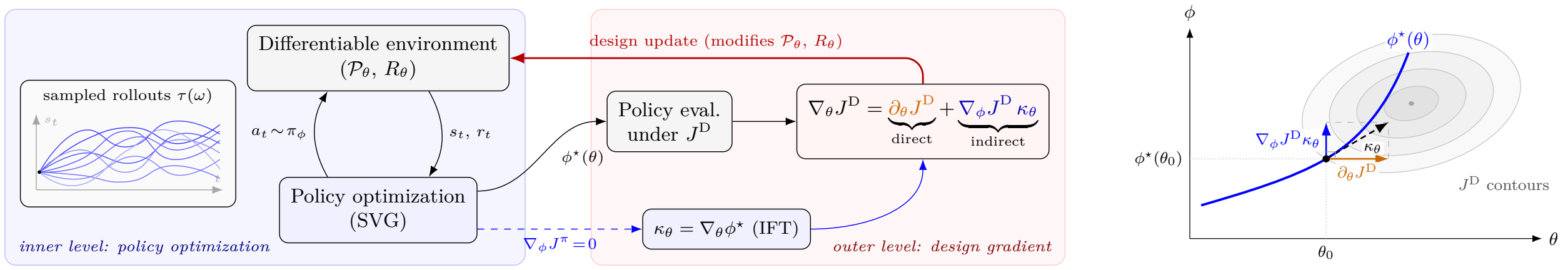

%% file: sections/related_work.tex
\section{Related Work}\label{sec:related_work}
Typically, the co-design problem is cast as a bilevel optimization: an outer-level search over the design space (\ac{es} or \ac{bo}) wrapped around an inner-level policy optimization scheme \citep{herber2017nested,cheney2014unshackling,gupta2021derl,bhatia2021evolutiongym,schaff2019jointly}. These methods search for a global optimum but ignore the structure of the problem, in particular how the design implicitly shapes the policy, and they scale poorly with the design dimension. We instead exploit the differentiable structure of \ac{rl}-based policy optimization in a differentiable simulator, trading global exploration for sample efficiency and scalability.

Recent works optimize design and an \ac{rl} policy jointly on a single shared objective~\citep{spielberg2019learning,yuan2022transform2act}. This has two consequences: the single objective cannot separate the reward that trains the policy from the criterion that evaluates the design, and the design update treats the current policy as fixed, ignoring how the optimal policy shifts with the design. We instead keep the two levels and their objectives separate, and include the gradient of the optimal policy with respect to design parameters by implicit differentiation.

Differentiating through a converged inner optimization via implicit differentiation is standard in bilevel learning~\citep{blondel2022implicit}, e.g., in hyperparameter optimization~\citep{lorraine2020optimizing}, meta-learning~\citep{rajeswaran2019imaml}, model-based \ac{rl}~\citep{nikishin2022control}, and sim-to-real \ac{rl}~\citep{anand2026bilevel}. A similar approach is used for co-design in a model-free setting: the policy is trained with \ac{ppo}, and the co-design gradient is obtained from a surrogate policy-gradient objective with a Fisher/Hessian approximation over a non-differentiable morphology interface~\citep{dai2026stackelberg}. We instead consider a fully differentiable setting with \ac{pd}-based (\ac{svg}-style) policy optimization and derive the corresponding exact co-design gradient using implicit differentiation. We do not compare directly against co-design methods such as Transform2Act~\citep{yuan2022transform2act} or Stackelberg \ac{ppo}~\citep{dai2026stackelberg}, since they target model-free \ac{rl} with discrete design spaces and non-differentiable simulators and we consider a fully differentiable setting with model-based \ac{rl}.

%% file: sections/background.tex
\section{Background}\label{sec:background}
\subsection{Markov Decision Processes and Reinforcement Learning}
\label{sec:bg_mdp}
Formally, the \ac{rl} problem is modeled as a discounted \ac{mdp} \((\mathcal{S}, \mathcal{A}, \mathcal{P}, R, \gamma)\)~\citep{sutton2018reinforcement}, where \(\mathcal{S}\) is the state space, \(\mathcal{A}\) the action space, \(\mathcal{P}(s_{t+1}\mid s_t,a_t)\) the transition kernel giving the probability of reaching \(s_{t+1}\) from \((s_t,a_t)\), \(R(s,a):\mathcal{S}\times\mathcal{A}\to\mathbb{R}\) the scalar reward function, and \(\gamma\in(0,1)\) the discount factor. A policy \(\pi(a\mid s)\) maps each state to a distribution over actions; together with an initial-state distribution \(p(s_0)\) it induces a trajectory distribution \(p_\pi\). The objective of \ac{rl} is to find the optimal policy \(\pi^{\star}\) maximizing the expected discounted return \({J(\pi) = \mathbb{E}_{\tau \sim p_\pi}\!\left[ \sum_{t=0}^{\infty} \gamma^t R(s_t, a_t) \right]}\). The state- and action-value functions are \({V^\pi(s) = \mathbb{E}_{\pi}\!\left[ \sum_{t=0}^{\infty} \gamma^t R(s_t, a_t) \mid s_0 = s \right]}\) and \({Q^\pi(s, a) = \mathbb{E}_{\pi}\!\left[ \sum_{t=0}^{\infty} \gamma^t R(s_t, a_t) \mid s_0 = s, a_0 = a \right]}\), with advantage \(A^\pi(s, a)=Q^\pi(s, a) - V^\pi(s)\). The optimal value and action-value functions \(V^{\star},Q^{\star}\) satisfy the Bellman optimality conditions \(V^{\star}(s) = \max_a Q^{\star}(s, a)\) and \(Q^{\star}(s, a) = R(s, a) + \gamma\, \mathbb{E}_{s' \sim \mathcal{P}}\!\left[ V^{\star}(s') \right]\), and the optimal policy is \(\pi^{\star}(s) = \arg\max_a Q^{\star}(s, a)\).

\subsection{Pathwise Policy Gradients: SVG and SHAC}
\label{sec:bg_pd}
A parameterized \ac{rl} policy \(\pi_\phi(a\mid s)\) is optimized by policy-gradient methods using an estimate of \(\nabla_\phi J\). The model-free \ac{lr} (score-function) estimator differentiates only the log-density \(\log\pi_\phi\) and uses only data samples from the dynamics model. When the transition model is differentiable and \emph{reparameterizable}, with exogenous, parameter-independent noise as \(a_t=g_\phi(s_t,\varepsilon_t)\) and \(s_{t+1}=f(s_t,a_t,\xi_t)\), the return \(J(\phi)\) is a deterministic, differentiable function of \(\phi\) for fixed noise. The \ac{pd} estimator, also known as the reparameterization gradient, differentiates it along the sampled trajectory through the Jacobians of \(f\) and \(g_\phi\), giving \(\nabla_\phi J=\mathbb{E}_{\varepsilon,\xi}[\nabla_\phi\sum_t\gamma^t R(s_t,a_t)]\). When the closed-loop dynamics are smooth and non-expansive over the differentiated horizon, this estimator has far lower variance than \ac{lr}, since its variance is governed by path smoothness rather than return magnitude~\citep{mohamed2020monte}.

Differentiating through a long rollout, however, makes the \ac{pd} estimator's variance grow with the horizon and induces exploding or vanishing gradients~\citep{parmas2018pipps}. \ac{svg}~\citep{heess2015svg}, the canonical model-based pathwise value-gradient family, addresses this by bootstrapping a learned critic so the full horizon need not be differentiated. \ac{shac}~\citep{xu2022shac} is a recent \ac{svg}-style algorithm that differentiates only a short rollout of horizon \(H\) and bootstraps the tail with a terminal critic, \(J_H(\phi)=\mathbb{E}[\sum_{t=0}^{H-1}\gamma^t R(s_t,a_t)+\gamma^H V(s_H)]\). In \ac{shac}, the critic is a frozen target at \(s_H\); our co-design gradient builds on these methods.

%% file: sections/codesign_grad.tex
\section{Gradient-based Co-design for RL}\label{sec:codesign}
In this section, we derive the co-design policy gradient, which forms the main contribution of this work. We consider differentiable simulators and policies optimized using \ac{svg}-style \ac{rl} algorithms~\citep{heess2015svg}, which rely on \ac{pd}-based policy gradients. We formalize the co-design problem as a bilevel optimization problem in Section~\ref{sec:codesign_problem}, derive its policy gradient in Section~\ref{sec:outer_gradient_general}, and finally derive the gradient for a practical \ac{shac}-style truncated form with a terminal critic in Section~\ref{sec:outer_gradient_shac}.

\subsection{Differentiable Co-design Problem}\label{sec:codesign_problem}
We consider a family of discounted \acp{mdp} \(
\mathcal{M}_\theta
=
(\mathcal{S},\mathcal{A},{\mathcal P}_\theta,R_\theta,\gamma),
\) where the transition kernel \({\mathcal P}_\theta\) and reward function \(R_\theta\) are parameterized by the design variables \(\theta\), and \(\gamma\in(0,1)\). Throughout this paper, we work with a reparameterized (pathwise) model of the \ac{mdp} under the following assumption.
\begin{assumption}[Differentiable reparameterized simulator]
\label{ass:differentiable_simulator}
The dynamics, policy, and reward admit the reparameterized form: \(s_{t+1}=f_\theta(s_t,a_t,\xi_t)\), \(a_t=g_\phi(s_t,\varepsilon_t)\), and \(r_t=R_\theta(s_t,a_t)\), with exogenous noise \(\xi_t\sim p(\xi)\) and \(\varepsilon_t\sim p(\varepsilon)\) whose laws do not depend on \((\phi,\theta)\), and with \(f_\theta\), \(g_\phi\), and \(R_\theta\) differentiable in \((s,a,\theta,\phi)\).
\end{assumption}

Assumption~\ref{ass:differentiable_simulator} holds for a broad class of systems with smooth, differentiable dynamics and reparameterizable noise. Additive or multiplicative Gaussian disturbances used to model uncertainty are one common example; more generally, the assumption accommodates any exogenous noise whose sampling can be reparameterized independently of \((\phi,\theta)\). Even non-smooth effects such as contact, switching, or saturation can be replaced by smooth surrogates (soft contact, \(\tanh\)/softplus saturations) that preserve differentiability. Under Assumption~\ref{ass:differentiable_simulator}, a sampled trajectory \(\tau=(s_0,a_0,s_1,a_1,\ldots)\) is a deterministic, differentiable function of \((\phi,\theta)\) for fixed noise \((\xi_{0:\infty},\varepsilon_{0:\infty})\). Methods, such as \ac{svg} and \ac{shac}, exploit this differentiable structure, which also enables the co-design gradient derived in the next section.

\paragraph{Policy optimization:} We collect the exogenous noise in \(\omega=(s_0,\xi_{0:\infty},\varepsilon_{0:\infty})\sim\mu\), with \(\mu\) independent of \((\phi,\theta)\), so the return is a deterministic, differentiable function of \((\phi,\theta,\omega)\). For a fixed design \(\theta\), the reparameterized form of \emph{policy objective} is
\begin{equation}
\label{eq:J_policy}
J^{\pi}(\phi,\theta)
=
\mathbb{E}_{\omega\sim\mu}\!\left[
\sum_{t=0}^{\infty}\gamma^t R^{\pi}_\theta(s_t,a_t)
\right],
\end{equation}
and the corresponding optimal policy parameters \(\phi^\star(\theta)\) for a given \(\theta\) are given by,
\begin{equation}
\label{eq:inner_rl}
\phi^\star(\theta)
\in
\arg\max_{\phi} J^{\pi}(\phi,\theta).
\end{equation}
Changing the design changes the induced \ac{mdp} and hence its optimal policy. In practice \(\phi^\star(\theta)\) is approximated by a \ac{pd}-based method (\ac{shac} in our experiments); for brevity, we write the policy reward as \(R_\theta\).

\paragraph{Design objective:} in co-design, the policy and the design may be optimized against \emph{different} objectives. For example, the design reward could have an additional capital cost term that penalizes expensive designs. We therefore distinguish the policy reward \(R^{\pi}_\theta(s,a)\), used to \emph{train the policy}, from a design reward \(R^{\mathrm{D}}_\theta(s,a)\), used to \emph{evaluate the design}. For a policy \(\pi_\phi\) and a design \(\theta\), the design objective is
\begin{equation}
\label{eq:J_design}
J^{\mathrm{D}}(\phi,\theta)
=
\mathbb{E}_{\omega\sim\mu}\!\left[
\sum_{t=0}^{\infty}\gamma^t R^{\mathrm{D}}_\theta(s_t,a_t)
\right]
-
\rho(\theta),
\end{equation}
where \(\rho(\theta)\geq 0\) is a design-cost term (e.g., a capital cost of the design) that depends on \(\theta\) alone. In co-design this objective is evaluated at the inner solution, \(J^{\mathrm{D}}(\phi^\star(\theta),\theta)\).

\paragraph{Co-design problem:} the objective of co-design is to identify the design parameters \(\theta^\star\) that maximize the design objective~\ref{eq:J_design}. This can be formulated as a bilevel optimization problem:
\begin{equation}
\label{eq:bilevel_codesign}
\theta^\star
\in
\arg\max_{\theta} J^{\mathrm{D}}(\phi^\star(\theta),\theta),
\qquad
\text{s.t.}\quad
\phi^\star(\theta)\in\arg\max_{\phi} J^{\pi}(\phi,\theta)\,.
\end{equation}
If we can compute the gradient of the design objective~\ref{eq:J_design} with respect to \(\theta\), we can solve for \(\theta^\star\) using gradient-based methods.  The next section derives this co-design policy gradient.

\subsection{Co-design Policy Gradient}
\label{sec:outer_gradient_general}
From here on we drop the policy argument whenever it is at the inner solution, writing \(J^{\mathrm{D}}(\theta)\triangleq J^{\mathrm{D}}(\phi^\star(\theta),\theta)\) and likewise for the truncated objectives of Section~\ref{sec:outer_gradient_shac}; the number of arguments distinguishes the two. Throughout, \(\nabla_x\) of a vector-valued map denotes its Jacobian (rows indexed by outputs, columns by inputs); the gradient of a scalar is accordingly a row vector, so \(\kappa_\theta=\nabla_\theta\phi^\star(\theta)\in\mathbb{R}^{\dim\phi\times\dim\theta}\) and products such as \(\nabla_\phi J^{\mathrm{D}}\,\kappa_\theta\) conform. We write \(\partial_\theta\) for partial derivatives that hold \(\phi^\star\) fixed and \(\nabla_\theta\) for total derivatives through \(\phi^\star(\theta)\); implementation-level expressions identify a scalar's gradient with the column vector returned by reverse-mode \ac{ad}, writing the contraction as \(\kappa_\theta^\top u\). Here \(\phi\) denotes a free policy parameter and \(\phi^\star(\theta)\) the inner solution at design \(\theta\): derivatives in the policy are taken with respect to the free argument \(\phi\) and then evaluated at \(\phi^\star(\theta)\), whereas \(\kappa_\theta\) differentiates the solution map itself. Under Assumption~\ref{ass:differentiable_simulator}, the gradient of \(J^{\mathrm{D}}(\theta)\)~\eqref{eq:J_design} is given by the chain rule as:
\begin{equation}
\label{eq:outer_grad_chainrule}
\nabla_\theta J^{\mathrm{D}}(\theta)
=
\underbrace{
\partial_\theta J^{\mathrm{D}}(\phi^\star(\theta),\theta)
}_{\text{direct design effect}}
+
\underbrace{
\nabla_\phi J^{\mathrm{D}}(\phi^\star(\theta),\theta)\,\kappa_\theta
}_{\text{policy-sensitivity effect}}.
\end{equation}
where \(\kappa_\theta \triangleq \nabla_\theta \phi^\star(\theta)\) represents the sensitivity of the optimal policy to design parameters. The \emph{direct} term holds the policy fixed and captures the influence of \(\theta\) through the dynamics \(f_\theta\), the design reward \(R^{\mathrm{D}}_\theta\), and the design cost \(\rho\). The \emph{policy-sensitivity} term captures the design's indirect effect on the objective via the policy: changing \(\theta\) changes \(\phi^\star(\theta)\) via \(\kappa_\theta\), which in turn changes \(J^{\mathrm{D}}\). The sensitivity \(\kappa_\theta\) is the most difficult term to obtain here, and its derivation is given in Section~\ref{appendix:shac_sensitivity_full}. Figure~\ref{fig:overview}(b) shows the two effects geometrically: \(\kappa_\theta\) is the slope of the manifold \(\phi^\star(\theta)\), while \(\nabla_\phi J^{\mathrm{D}}\) measures how the design objective changes when the policy moves off that manifold.
\begin{lemma}[Envelope property]
\label{lem:envelope}
Suppose the design and policy objectives match, \(R^{\mathrm{D}}_\theta=R^{\pi}_\theta\) with \(\rho\) depending on \(\theta\) alone, so that \(J^{\mathrm{D}}(\phi,\theta)=J^{\pi}(\phi,\theta)-\rho(\theta)\), and the inner policy is exactly stationary, \(\nabla_\phi J^{\pi}(\phi^\star(\theta),\theta)=0\). Then \(\nabla_\phi J^{\mathrm{D}}(\phi^\star(\theta),\theta)=0\), so the policy-sensitivity term in~\eqref{eq:outer_grad_chainrule} vanishes, leaving only the direct design effect.
\end{lemma}
\begin{proof}
Since \(\rho\) is independent of \(\phi\), \(\nabla_\phi J^{\mathrm{D}}(\phi,\theta)=\nabla_\phi J^{\pi}(\phi,\theta)\) for all \((\phi,\theta)\); evaluated at \(\phi^\star(\theta)\) this vanishes by stationarity, eliminating the policy-sensitivity term in~\eqref{eq:outer_grad_chainrule}.
\end{proof}
Lemma~\ref{lem:envelope} is the classical envelope theorem~\citep{milgrom2002envelope}: at an inner optimum, the policy shift induced by a design change has no first-order effect on the very objective the policy optimizes, so the design gradient may ignore the policy's re-adaptation. The simplification is strictly first-order and does not extend to second-order sensitivities.

\begin{proposition}[Pathwise co-design gradient]
\label{prop:pd_codesign_gradient}
Let Assumption~\ref{ass:differentiable_simulator} hold, let \(\phi^\star(\theta)\) be differentiable with \(\kappa_\theta=\nabla_\theta\phi^\star(\theta)\), and let \(h_t(\theta,\omega)\triangleq\gamma^t R^{\mathrm{D}}_\theta(s_t,a_t)\) denote the discounted per-step design reward along the trajectory generated by \((\phi^\star(\theta),\theta,\omega)\). Assuming for some neighborhood \(\Theta_0\) of \(\theta\),\(
\sum_{t=0}^{\infty}\;\mathbb{E}_{\omega\sim\mu}\Big[\,\sup_{\theta'\in\Theta_0}\big\|\nabla_{\theta'}h_t(\theta',\omega)\big\|\Big]<\infty\). Then \(\nabla_\theta\) interchanges with \(\mathbb{E}_{\omega\sim\mu}\) and with the infinite-horizon sum by dominated convergence, and the co-design gradient is
\begin{equation}
\label{eq:pd_codesign_gradient}
\nabla_\theta J^{\mathrm{D}}(\theta)
=
\mathbb{E}_{\omega\sim\mu}\!\left[
\sum_{t=0}^{\infty}\gamma^t\Big(
\partial_\theta R^{\mathrm{D}}_\theta(s_t,a_t)
+\nabla_s R^{\mathrm{D}}_\theta(s_t,a_t)\,S_t
+\nabla_a R^{\mathrm{D}}_\theta(s_t,a_t)\,A_t
\Big)
\right]
-\nabla_\theta\rho(\theta),
\end{equation}
where the total state and action sensitivities \(S_t=\mathrm{d}s_t/\mathrm{d}\theta\) and \(A_t=\mathrm{d}a_t/\mathrm{d}\theta\) satisfy:
\begin{align}
\label{eq:pd_action_rec_main}
A_t &= \nabla_s g_{\phi^\star(\theta)}(s_t,\varepsilon_t)\,S_t
+\nabla_\phi g_\phi(s_t,\varepsilon_t)\big|_{\phi^\star(\theta)}\,\kappa_\theta,\\
\label{eq:pd_state_rec_main}
S_{t+1} &= \partial_\theta f_\theta(s_t,a_t,\xi_t)
+\nabla_s f_\theta(s_t,a_t,\xi_t)\,S_t
+\nabla_a f_\theta(s_t,a_t,\xi_t)\,A_t,
\end{align}
initialized at \(S_0=\mathrm{d}s_0/\mathrm{d}\theta\) (with \(S_0=0\) when \(p(s_0)\) is design-independent).
\end{proposition}

The proof of Proposition~\ref{prop:pd_codesign_gradient}, together with the equivalent reverse-mode (adjoint / backpropagation-through-time) form, is given in Appendix~\ref{app:pd_codesign_gradient}. The reverse-mode form makes the direct vs policy-sensitivity  split~\eqref{eq:outer_grad_chainrule} explicit.
In implementation, the sensitivities \(S_t,A_t\) and the equivalent adjoints are computed by a single reverse-mode automatic-differentiation pass through the unrolled simulator evaluates \(\nabla_\theta J^{\mathrm{D}}_\theta\) given the policy sensitivity \(\kappa_\theta\) as an input. \(\kappa_\theta\) can not be computed through automatic differentiation; it is obtained by the implicit-function solve of Section~\ref{appendix:shac_sensitivity_full}.

Evaluating~\eqref{eq:pd_codesign_gradient}  by differentiating a full or long rollout is problematic. Backpropagation through many steps composes a long product of state Jacobians \(\nabla_s f_\theta\), so the sensitivities \(S_t\) (and the adjoints) tend to explode or vanish, and its variance grows with the horizon, rendering long pathwise gradients unreliable.  This motivates, \ac{svg}-style objective, trading a small bias for a large reduction in variance.

\subsection{SVG-style Pathwise Co-design Gradient}
\label{sec:outer_gradient_shac}
As in \ac{svg} and \ac{shac}, to avoid issues related to differentiating through a long horizon, we replace the full rollout in~\eqref{eq:pd_codesign_gradient} by a short differentiable rollout of length \(H\) whose tail is bootstrapped by a learned critic as
\begin{equation}
\label{eq:svg_objective}
J_H^{\mathrm{D}}(\phi,\theta)
=
\mathbb{E}_{\omega_H}\!\left[
\sum_{t=0}^{H-1}\gamma^t R^{\mathrm{D}}_\theta(s_t,a_t)
+\gamma^H V^{\mathrm{D}}_\theta(s_H)
\right]
-\rho(\theta),
\qquad
\omega_H=(s_0,\xi_{0:H-1},\varepsilon_{0:H-1})\,,
\end{equation}
where \(V^{\mathrm{D}}_\theta(s)\) is a terminal critic that depends on the design \(\theta\) through the dynamics, the design reward, and the policy.
Differentiating only \(H\) steps in~\eqref{eq:svg_objective} keeps the Jacobian product short, controlling variance, while the critic accounts for the truncated horizon. The policy is trained on the analogous \ac{svg}-style truncated objective, with its own terminal critic \(V^{\pi}_\theta\),
\begin{equation}
\label{eq:Jsvg_inner}
J^{\pi}_H(\phi,\theta)
=
\mathbb{E}_{\omega_H}\!\left[
\sum_{t=0}^{H-1}\gamma^t R^{\pi}_\theta(s_t,a_t)
+\gamma^H V^{\pi}_\theta(s_H)
\right],
\end{equation}
whose maximizer over \(\phi\) is the inner solution \(\phi^\star(\theta)\); its sensitivity \(\kappa_\theta\) is derived in Section~\ref{appendix:shac_sensitivity_full}.

\begin{proposition}[\ac{svg}-style co-design gradient]
\label{prop:svg_codesign_gradient}
Under Assumption~\ref{ass:differentiable_simulator}, with a differentiable terminal critic \(V^{\mathrm{D}}_\theta\) and \(\kappa_\theta=\nabla_\theta\phi^\star(\theta)\), the gradient of the design objective~\eqref{eq:svg_objective} is
\begin{align}
\label{eq:svg_codesign_gradient}
\nabla_\theta J_H^{\mathrm{D}}(\theta)
=
\mathbb{E}_{\omega_H}\Bigg[
&\sum_{t=0}^{H-1}\gamma^t\Big(
\partial_\theta R^{\mathrm{D}}_\theta(s_t,a_t)
+\nabla_s R^{\mathrm{D}}_\theta(s_t,a_t)\,S_t
+\nabla_a R^{\mathrm{D}}_\theta(s_t,a_t)\,A_t
\Big)
\nonumber\\[-2pt]
&+\gamma^H\Big(
{\partial_\theta V^{\mathrm{D}}_\theta(s_H)}
+\nabla_s V^{\mathrm{D}}_\theta(s_H)\,S_H
\Big)
\Bigg]
-\nabla_\theta\rho(\theta),
\end{align}
with \(S_t,A_t\) given by the forward recursions~\eqref{eq:pd_action_rec_main}--\eqref{eq:pd_state_rec_main}.
\end{proposition}
 Proof of Proposition~\ref{prop:svg_codesign_gradient} and equivalent adjoint mode form is given in Appendix~\ref{app:svg_codesign_gradient}. In reverse mode the gradient is generated by the terminal costate \(\lambda_H=\gamma^H\nabla_s V^{\mathrm{D}}_\theta(s_H)\) together with the term \(\gamma^H\partial_\theta V^{\mathrm{D}}_\theta(s_H)\), and decomposes into a direct design effect and a policy-sensitivity effect proportional to \(\kappa_\theta\), exactly as in~\eqref{eq:outer_grad_chainrule}.

\paragraph{Estimating the gradient of the critic.}
The term \(\partial_\theta V^{\mathrm{D}}_\theta(s_H)\) can be estimated without unrolling the discarded tail by implicitly differentiating the Bellman fixed point of the design value function, \(V^{\mathrm{D}}_\theta(s)=\mathbb{E}_{\varepsilon}[R^{\mathrm{D}}_\theta(s,a)+\gamma\,\mathbb{E}_{\xi}V^{\mathrm{D}}_\theta(s')]\). This is formalized in the following lemma.

\begin{lemma}[Critic sensitivity]
\label{lem:critic_design_sensitivity}
Under Assumption~\ref{ass:differentiable_simulator}, the design sensitivity \(w_\theta(s)\triangleq\partial_\theta V^{\mathrm{D}}_\theta(s)\) satisfies the linear Bellman fixed point
\begin{equation}
\label{eq:critic_sensitivity_bellman}
w_\theta(s)=b_\theta(s)+\gamma\,\mathbb{E}_{\varepsilon,\xi}\!\left[w_\theta(s')\right],
\qquad\text{equivalently}\qquad
w_\theta=(I-\gamma P^{\pi}_\theta)^{-1}b_\theta,
\end{equation}
with closed-loop transition operator \((P^{\pi}_\theta h)(s)\triangleq\mathbb{E}_{\varepsilon,\xi}[h(s')]\) and per-step design-sensitivity source \(b_\theta(s)=\mathbb{E}_{\varepsilon}\big[\partial_\theta R^{\mathrm{D}}_\theta+\nabla_a R^{\mathrm{D}}_\theta\,\delta a+\gamma\,\mathbb{E}_{\xi}[\nabla_s V^{\mathrm{D}}_\theta(s')\,\delta s']\big]\), where \(\delta a=\nabla_\phi g_\phi|_{\phi^\star(\theta)}\kappa_\theta\) is the design sensitivity of the action at fixed state, the first-order response of the action to \(\theta\) through \(\phi^\star\) alone, and \(\delta s'=\partial_\theta f_\theta+\nabla_a f_\theta\,\delta a\) is the induced one-step state sensitivity. The solve is well posed for \(\gamma<1\).
\end{lemma}

Thus \(w_\theta=\partial_\theta V^{\mathrm{D}}_\theta\)  can be learned by a \emph{critic sensitivity} function approximator \(W_\zeta(s)\approx\partial_\theta V^{\mathrm{D}}_\theta(s)\), trained by the same temporal-difference regression as the value critic and reusing the rollouts, the policy sensitivity \(\kappa_\theta\), and the Jacobians already computed for~\eqref{eq:svg_codesign_gradient}. Substituting \(W_\zeta(s_H)\) for \(\partial_\theta V^{\mathrm{D}}_\theta(s_H)\) recovers the full \ac{svg} co-design gradient, while \(W_\zeta\equiv0\) recovers the \ac{shac} semi-gradient (Corollary~\ref{cor:shac_semigradient}). The construction follows the stochastic value-gradient recursion of \citet{heess2015svg}; the proof is given in Appendix~\ref{app:critic_design}.

Alternatively, unrolling the fixed point~\eqref{eq:critic_sensitivity_bellman} gives the Monte Carlo form
\begin{equation}
\label{eq:critic_sensitivity_mc}
w_\theta(s_0)=\mathbb{E}\Big[\textstyle\sum_{t\ge0}\gamma^t\,b_\theta(s_t)\Big],
\end{equation}
which can be estimated through the backward recursion: \(w_t=b_\theta(s_t)+\gamma\,w_{t+1}\) for every state of a rolled-out episode at once, with no extra function approximator. As \(b_\theta\) uses the one-step \(\delta s'\) and the value-critic gradient \(\nabla_s V^{\mathrm{D}}_\theta\) rather than a chained state Jacobian, this sum stays bounded-variance over a long tail, unlike the full pathwise gradient~\eqref{eq:pd_codesign_gradient}. The \(n\)-step target \(\sum_{k=0}^{n-1}\gamma^k b_\theta(s_{t+k})+\gamma^n W_\zeta(s_{t+n})\) recovers the temporal-difference estimator~\eqref{eq:critic_sensitivity_bellman} for \(n=1\), while \(n\to\infty\) recovers the Monte Carlo estimator~\eqref{eq:critic_sensitivity_mc}.

Either critic sensitivity estimator has a cost: the complexity of learning \(W_\zeta\) and its approximation error for~\eqref{eq:critic_sensitivity_bellman}, or variance in the estimate for~\eqref{eq:critic_sensitivity_mc}. Algorithms such as \ac{shac} instead find it simpler and more stable to freeze the terminal critic to a state-only target, avoiding differentiating the critic with respect to \(\theta\) altogether. This is formalized in the following corollary.

\begin{corollary}[\ac{shac}-style semi-gradient]
\label{cor:shac_semigradient}
Let \(J^{\mathrm{D},+}_H\) denote the objective~\eqref{eq:svg_objective} with the terminal critic replaced by a frozen, state-only target \(V^{+}(s)\), as in \ac{shac}. Then \(\partial_\theta V^{+}\equiv0\) by construction, and the exact gradient of this surrogate is the \ac{shac}-style semi-gradient
\begin{equation}
\label{eq:shac_semigradient}
\nabla_\theta J_H^{\mathrm{D},+}(\theta)
=
\mathbb{E}_{\omega_H}\!\left[
\sum_{t=0}^{H-1}\gamma^t\big(
\partial_\theta R^{\mathrm{D}}_\theta
+\nabla_s R^{\mathrm{D}}_\theta\,S_t
+\nabla_a R^{\mathrm{D}}_\theta\,A_t
\big)
+\gamma^H\nabla_s V^{+}(s_H)\,S_H
\right]
-\nabla_\theta\rho(\theta).
\end{equation}
\end{corollary}
Corollary~\ref{cor:shac_semigradient} follows directly from Proposition~\ref{prop:svg_codesign_gradient}. The semi-gradient differs from the true gradient~\eqref{eq:svg_codesign_gradient} exactly by the omitted critic term \(\gamma^H\,\mathbb{E}_{\omega_H}\!\big[\partial_\theta V^{\mathrm{D}}_\theta(s_H)\big]\)---the quantity that \(W_\zeta\) estimates (Lemma~\ref{lem:critic_design_sensitivity}).

\paragraph{Implementation details.}
 The \emph{direct} design term in~\eqref{eq:svg_codesign_gradient} is a single reverse-mode \ac{ad} pass through the differentiable rollout: one differentiates the truncated return with respect to \(\theta\) while holding \(\phi=\phi^\star\) fixed, and \ac{ad} assembles the per-step reward, dynamics, and terminal-value contributions, i.e.\ the costates of Appendix~\ref{app:svg_codesign_gradient}. The \ac{shac} semi-gradient of Corollary~\ref{cor:shac_semigradient} is computed the same way on the surrogate \(J^{\mathrm{D},+}_H\): the state-only critic carries no design argument, so \(\partial_\theta V^{+}\equiv0\) by construction, and a \texttt{stop\_gradient} on the critic parameters additionally holds it fixed as a bootstrap target during the actor/design update. The \emph{policy-sensitivity} effect, only \emph{partly} uses \ac{ad}: the vector \(\nabla_\phi J^{\mathrm{D}}(\phi^\star,\theta)\) is one \ac{ad} pass, but contracting it with \(\kappa_\theta\) requires the inner-solution sensitivity of Section~\ref{appendix:shac_sensitivity_full}, which is \emph{not} available through \ac{ad}.

%% file: sections/shac_sensitivity.tex
\subsection{Sensitivity of the SVG-style Inner Solution}
\label{appendix:shac_sensitivity_full}

The co-design gradients of Propositions~\ref{prop:pd_codesign_gradient} and~\ref{prop:svg_codesign_gradient} require the policy sensitivity \(\kappa_\theta=\nabla_\theta\phi^\star(\theta)\). Since the policy is optimized using an \ac{svg}-style objective~\eqref{eq:Jsvg_inner} in practice, we derive the corresponding sensitivity \(\kappa_\theta\) here.

The locally converged actor \(\phi^\star(\theta)\), the solution of~\eqref{eq:inner_rl} of the objective \(J^{\pi}_H(\phi,\theta)\)~\eqref{eq:Jsvg_inner} satisfies the condition:
\begin{equation}
\label{eq:inner_stationarity}
F(\phi,\theta)\triangleq\nabla_\phi J^{\pi}_H(\phi,\theta)=0 .
\end{equation}

Applying \ac{ift} to condition~\eqref{eq:inner_stationarity} provides the policy sensitivity as given in the following proposition.

\begin{lemma}[Inner-solution sensitivity]
\label{lem:inner_sensitivity}
Assuming \(F\) is continuously differentiable near \((\phi^\star(\theta),\theta)\), the stationary point is locally isolated, and the policy Hessian \(\nabla^2_{\phi\phi}J^{\pi}_H(\phi^\star(\theta),\theta)\) is nonsingular. Then sensitivity of the policy \(\phi^\star(\theta)\) with respect to the design parameters \(\theta\) is given by:
\begin{equation}
\label{eq:inner_ift}
\kappa_\theta \triangleq \nabla_\theta\phi^\star(\theta)
=-\big(\nabla^2_{\phi\phi}J^{\pi}_H(\phi^\star(\theta),\theta)\big)^{-1}\,\nabla^2_{\theta\phi}J^{\pi}_H(\phi^\star(\theta),\theta)\,,
\end{equation}
where \(\nabla^2_{\phi\phi}J^{\pi}_H\) represents the policy Hessian holding \(\theta\) fixed and \(\nabla^2_{\theta\phi}J^{\pi}_H\) the cross-term of the full Hessian of \(J^{\pi}_H\) with respect to \((\phi,\theta)\). The cross-term carries the explicit design dependence of the dynamics and reward (\(\partial_\theta f_\theta,\ \partial_\theta R_\theta\)) together with the explicit critic--design coupling \(\gamma^H\partial^2_{\theta s}V^{\pi}_\theta(s_H)\).
\end{lemma}

The following corollary provides the sensitivity of the \ac{shac}-style semi-gradient, which we use in practice.
\begin{corollary}[\ac{shac} semi-gradient sensitivity]
\label{cor:inner_semigradient}
If the terminal critic is a frozen, state-only target \(V^{\pi}_\theta(s_H)=V^{+}(s_H)\), then \(\partial^2_{\theta s}V^{\pi}_\theta(s_H)=0\) and~\eqref{eq:inner_ift} reduces to the semi-gradient
\begin{equation}
\label{eq:inner_semigradient}
\hat\kappa_\theta
=-\big(\nabla^2_{\phi\phi}J^{\pi,+}_H\big)^{-1}\nabla^2_{\theta\phi}J^{\pi,+}_H,
\end{equation}
computed by the same recursions with the explicit critic--design term \(\gamma^H\partial^2_{\theta s}V^{\pi}_\theta(s_H)\) omitted from the terminal costate sensitivity (Appendix~\ref{app:inner_sensitivity}). This \(\hat\kappa_\theta\) is the inner-loop analogue of the \ac{shac} semi-gradient of Corollary~\ref{cor:shac_semigradient}.
\end{corollary}

The full recursions for the two second-order terms in Lemma~\ref{lem:inner_sensitivity}, the policy Hessian \(\nabla^2_{\phi\phi}J^{\pi}_H\) and the cross-term \(\nabla^2_{\theta\phi}J^{\pi}_H\), and for the semi-gradient version of Corollary~\ref{cor:inner_semigradient}, together with their proofs, are given in Appendix~\ref{app:inner_sensitivity}.

\paragraph{Implementation details.}
Both terms in~\eqref{eq:inner_ift} are second derivatives of the rollout objective, available as matrix-free operators from forward-mode \ac{ad} of the first-order gradient \(F=\nabla_\phi J^{\pi}_H\): the policy Hessian--vector product \(A\,v=\nabla^2_{\phi\phi}J^{\pi}_H\,v\) and the cross-term \(B^\top w=(\nabla^2_{\theta\phi}J^{\pi}_H)^\top w\). The \ac{shac} semi-gradient of Corollary~\ref{cor:inner_semigradient} is obtained from the same \ac{ad} computation, without the term \(\gamma^H\partial^2_{\theta s}V^{\pi}_\theta(s_H)\). We never form \(\kappa_\theta\) explicitly but only \emph{apply} it to a vector \(u\) : from~\eqref{eq:inner_ift},
\begin{equation}
\label{eq:kappa_apply}
\kappa_\theta^\top u=-B^\top A^{-1}u,
\end{equation}
the product is computed as one linear solve \(Ay=u\) followed by the single \ac{vjp} \(-B^\top y\); in the co-design gradient, \(u=\nabla_\phi J^{\mathrm{D}}\). The solve acts in the policy-parameter dimension: for moderate \(\dim\phi\) (up to a few thousand), \(A\) can simply be formed and factorized directly, as in our inventory experiments (Appendix~\ref{app:example}); for larger policy networks, a few matrix-free Krylov iterations on Hessian--vector products \(A\,v\) suffice (conjugate gradients when \(-A\) is positive definite, otherwise MINRES/GMRES), so that neither \(\kappa_\theta\), \(A\), nor \(A^{-1}\) is ever formed. Either way, the computation needs only first- and second-order derivatives of \(f_\theta\), \(g_\phi\), \(R_\theta\), and \(V^{\pi}_\theta\) along one sampled rollout. In practice the policy Hessian \(A=\nabla^2_{\phi\phi}J^{\pi}_H\) can be ill-conditioned and indefinite, so the solve \(Ay=u\) in~\eqref{eq:kappa_apply} is replaced by a diagonally damped one,
\begin{equation}
\label{eq:inner_ift_damped}
\kappa_\theta^\top u=-B^\top\big(A-\varepsilon I\big)^{-1}u,
\qquad \varepsilon>0\,.
\end{equation}
At a strict local maximum \(A\) is negative definite and the shift moves its spectrum away from zero; when \(A\) is indefinite at a partially converged solution, choosing \(\varepsilon\) larger than the largest positive eigenvalue of \(A\) makes \(A-\varepsilon I\) negative definite. The damping then trades a small bias in \(\kappa_\theta\) for a well-posed, numerically stable system. When \(A\) is formed densely, a rank-\(k\) truncated eigendecomposition regularizes the solve in the same way: the smallest eigenvalues, which carry the least reliable curvature, are discarded (Appendix~\ref{app:protocols}). In practice \(\varepsilon\) is set from the conditioning of \(A\) (larger when the inner solution is less converged). Algorithm~\ref{alg:codesign} summarizes the resulting bilevel co-design procedure; further implementation details are in Appendix~\ref{app:implementation}.

\paragraph{Convergence Analysis.}
A convergence analysis of Algorithm~\ref{alg:codesign} is provided in Appendix~\ref{app:convergence}. Under local regularity (Assumption~\ref{ass:convergence}): a locally strongly concave inner problem along a continuous \(\phi^\star(\theta)\) and a contracting inner solver, and for a sufficiently small outer step size \(\alpha\), the average over \(N\) design updates satisfies,
\[
\frac1N\sum_{k<N}\mathbb{E}\big\|\nabla_\theta J^{\mathrm{D}}_H(\theta_k)\big\|^2
\le\mathcal{O}\Big(\frac{1}{\alpha N}\Big)+\mathcal{O}(\alpha\sigma^2)
+\mathcal{O}\big(\bar e^{\,2}+\Lambda^2\delta_\kappa^2+\delta_H^2\big)
\]
(Theorem~\ref{thm:convergence}). The first two terms are those of stochastic gradient ascent on a smooth nonconvex objective. The third comes from the bias of the gradient estimate and does not decrease with \(N\): the inner tracking error \(\bar e\), which shrinks with the outer step size under warm starts, so the inner loop need not be re-converged at every design; the error \(\delta_\kappa\) of the damped semi-gradient sensitivity solve~\eqref{eq:inner_ift_damped}, which enters only through the envelope factor \(\Lambda=\sup_\theta\|\nabla_\phi J^{\mathrm{D}}_H(\phi^\star(\theta),\theta)\|\) and therefore vanishes under matched objectives (Lemma~\ref{lem:envelope}); and the critic term \(\delta_H\), which carries the factor \(\gamma^H\). The design updates thus reach an approximately stationary design, with the accuracy set by these three terms; the bound applies to the average over the iterations, hence to the best iterate, rather than to the limit of the sequence.

\begin{algorithm}[t]
\caption{Gradient-based co-design with a pathwise (\ac{shac}-style) inner solver}
\label{alg:codesign}
\small
\begin{algorithmic}[1]
\Require design \(\theta_0\), policy \(\phi_0\), critic; damping \(\varepsilon>0\); step size \(\alpha\){; design set \(\Theta\), projection \(\Pi_\Theta\)}
\For{\(k=0,1,2,\dots\)}
  \State \textbf{inner solve:} warm-started \ac{shac} on \(J^{\pi}_H(\phi,\theta_k)\)~\eqref{eq:Jsvg_inner} until \(\nabla_\phi J^{\pi}_H\approx0\)~\eqref{eq:inner_stationarity}, yielding \(\phi^\star(\theta_k)\)
  \State \textbf{direct design effect:} \(g_{\mathrm{dir}}\gets\partial_\theta J^{\mathrm{D}}_H(\phi^\star,\theta_k)\) (Proposition~\ref{prop:svg_codesign_gradient}, Corollary~\ref{cor:shac_semigradient}), reverse-mode \ac{ad} of~\eqref{eq:svg_objective}
  \If{\(R^{\mathrm{D}}_\theta=R^{\pi}_\theta\)}\Comment{matched objectives: term vanishes (Lemma~\ref{lem:envelope})}
    \State \(g_{\mathrm{sens}}\gets0\)
  \Else
    \State \(u\gets\nabla_\phi J^{\mathrm{D}}_H(\phi^\star,\theta_k)\), one reverse-mode \ac{ad} pass in \(\phi\)
    \State \textbf{policy-sensitivity effect:} \(g_{\mathrm{sens}}\gets\kappa_\theta^\top u=-B^\top(A-\varepsilon I)^{-1}u\), the damped solve~\eqref{eq:inner_ift_damped} of Lemma~\ref{lem:inner_sensitivity}
  \EndIf
  \State \textbf{design update:} \(\theta_{k+1}\gets{\Pi_\Theta\big(}\theta_k+\alpha\,(g_{\mathrm{dir}}+g_{\mathrm{sens}}){\big)}\)\Comment{{projected }ascent on~\eqref{eq:outer_grad_chainrule}}
\EndFor
\end{algorithmic}
\end{algorithm}

%% file: sections/example.tex
\section{Experiments}
\label{sec:benchmark_diffusion_codesign}

\begin{wrapfigure}{r}{0.42\linewidth}
\vspace{-12pt}
\centering
\scalebox{0.8}{\input{figures/fig_inventory}}
\caption{A node in inventory--routing network}
\label{fig:inventory_diagram}
\vspace{-8pt}
\end{wrapfigure}
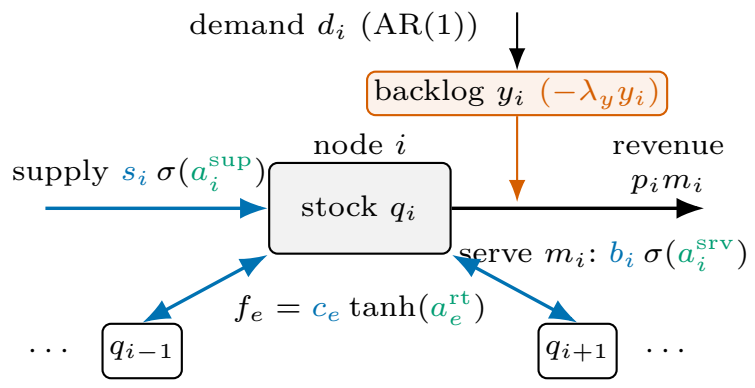
We evaluate the co-design \ac{rl} algorithm on a scalable in design space, fully differentiable inventory--routing network (Figure~\ref{fig:inventory_diagram}): a graph whose nodes hold stock that serves a stochastic local demand, and whose edges are transport links along which stock can be redistributed. Each node can draw new stock from a supply source, and demand it does not serve immediately accumulates as backlog. At every step, a policy decides how much stock each node draws from supply, how much demand it serves, and how stock is routed along the edges. The design parameters \(\theta\) determine the network \emph{infrastructure}; the supply, service, and routing capacities, each with an associated capital cost. Formally, the graph has \(N\) nodes and set of edges \(E\); the state \(x_t=(q_t,y_t,d_t)\in\mathbb{R}^{3N}\) includes current stock \(q_{i,t}\), backlog \(y_{i,t}\), and current demand \(d_{i,t}\) at each node. The design \(\theta=\{c_e,\,s_i,\,b_i\}\in\mathbb{R}^{n_\theta}\), with \(n_\theta=E+2N\), contains routing capacities \(c_e\ge0\) on edges, and supply and service capacities \(s_i,b_i\ge0\) at nodes. The policy outputs \(a_t=(a^{\mathrm{sup}}_t,a^{\mathrm{srv}}_t,a^{\mathrm{rt}}_t)\in\mathbb{R}^{2N+E}\), mapped to bounded controls through the design,
\begin{equation}
\label{eq:inv_action_map}
u^{\mathrm{sup}}_i=s_i\,\sigma(a^{\mathrm{sup}}_i),\qquad
u^{\mathrm{srv}}_i=b_i\,\sigma(a^{\mathrm{srv}}_i),\qquad
f_e=c_e\tanh(a^{\mathrm{rt}}_e),
\end{equation}
with \(\sigma\) the logistic function and \(f_e\) a signed edge flow: the policy sets a bounded utilization of each capacity (\(\sigma\) for one-directional supply and service, \(\tanh\) for bidirectional routing), so controls respect the design limits while staying differentiable in \((\phi,\theta)\). Shipments are limited by the node stock as well as by the edge capacity. Let \(s_{ie}=+1\) if edge \(e\) leaves node \(i\) and \(s_{ie}=-1\) if it enters it, so that \(\lceil s_{ie}f_e\rceil_{+}\) is the quantity node \(i\) ships on \(e\). All shipments of a node are scaled to the stock it holds,
\begin{equation}
\label{eq:inv_outflow_cap}
\eta_i=\mathrm{smin}\Big(1,\;\frac{q_i+u^{\mathrm{sup}}_i}{\sum_{e\ni i}\lceil s_{ie}f_e\rceil_{+}}\Big),\qquad
\tilde f_e=\eta_i\,\lceil f_e\rceil_{+}-\eta_j\,\lceil -f_e\rceil_{+}\quad\text{for } e=(i\to j),
\end{equation}
with \(\eta_i=1\) when node \(i\) ships nothing. Since \(\mathrm{smin}\) never exceeds the hard minimum, node \(i\) ships at most \(q_i+u^{\mathrm{sup}}_i\), so stock is conserved along the edges up to the smoothing offset of \(\mathrm{smin}\) (Appendix~\ref{app:inv_spec}). With the node inflow \(z_i=q_i+u^{\mathrm{sup}}_i+\sum_{e\in\mathrm{in}(i)}\tilde f_e-\sum_{e\in\mathrm{out}(i)}\tilde f_e\), the served quantity \(m_i\) and the state update are
\begin{equation}
\label{eq:inv_dynamics}
m_i=\mathrm{smin}\big(\lceil z_i\rceil_{+},\,y_i+d_i,\,u^{\mathrm{srv}}_i\big),\qquad
q_i^{+}=\lceil z_i-m_i\rceil_{+},\qquad
y_i^{+}=\lceil y_i+d_i-m_i\rceil_{+},
\end{equation}
where \(\lceil\cdot\rceil_{+}\) and \(\mathrm{smin}\) are softplus-smoothed \(\max(\cdot,0)\) and two- and three-way \(\min\) operators (Appendix~\ref{app:example}): under a hard \(\min\) only the binding constraint receives gradient---a capacity that is \emph{almost} binding gets exactly zero; the smooth versions share gradient among all near-binding arguments, preserving the design signal and the second derivatives \(\kappa_\theta\) requires. Demand is assumed to be an AR(1) process~\citep{lee1997information} passed through a softplus, \(d_i^{+}=\mathrm{softplus}\big((1-\nu_i)\mu_i+\nu_i d_i+\sigma_d\,\xi_i\big)\), \(\xi_i\sim\mathcal{N}(0,1)\), where \(\mu_i\) sets the average demand, \(\nu_i\in(0,1)\) the correlation between consecutive steps, and \(\sigma_d\) the noise scale. All maps are smooth in \((x,a,\theta)\) and the noise is exogenous (Assumption~\ref{ass:differentiable_simulator}); prices \(p_i\sim\mathcal{U}(0.5,2.0)\), holding costs, demand levels, and capital-cost weights are drawn once per environment instance and held fixed (Appendix~\ref{app:inv_spec}).

The policy is trained on an operational reward; the design objective adds a backlog penalty and subtracts a capital cost:
\begin{equation}
\label{eq:inv_rewards}
R^{\pi}_\theta = \sum_i p_i m_i-\lambda_a\overline{a^2}-\sum_i\lambda_{h,i}q_i^{2},
\qquad
R^{\mathrm{D}}_\theta = R^{\pi}_\theta-\lambda_y\sum_i y_i,
\qquad
\rho(\theta)=\lambda_{\mathrm{cap}}\,C(\theta),
\end{equation}
where \(p_i\) is the unit revenue for served demand, \(\lambda_a\) penalizes control effort through the mean squared action \(\overline{a^2}\), \(\lambda_{h,i}\) penalizes stored stock, \(\lambda_y\) penalizes backlog, and \(C(\theta)=\sum_e\alpha_e c_e^{2}+\sum_i\alpha^{s}_i s_i^{2}+\sum_i\alpha^{b}_i b_i^{2}\) is the quadratic cost of the installed capacities. The objectives are the discounted sums of~\eqref{eq:J_policy} and~\eqref{eq:J_design} with \(\gamma=0.99\) over \(T=200\)-step episodes. The backlog penalty is policy-dependent, so \(R^{\mathrm{D}}_\theta\neq R^{\pi}_\theta\) in exactly the way that activates the policy-sensitivity term; the design-only capital cost \(\rho\) contributes, by Lemma~\ref{lem:envelope}, \emph{only} to the direct design effect. The example is used in two configurations, \emph{misaligned} and \emph{default}, introduced with the corresponding experiments below (Appendix~\ref{app:example}).

Every experiment uses 5 environment instances per \(N\), with 3 inner-loop seeds per instance in the misaligned-configuration experiments (15 trials) and 5 seeds per method in the comparison study (25 trials); results report mean \(\pm\) SE across cells. The policy is a Gaussian \ac{nn} (one hidden layer, 32 units, normalized observations), trained by \ac{shac} on 512 parallel environments and warm-started across outer iterations; the policy-sensitivity term is estimated with a rank-\(k\) truncation of the \ac{ift} solve~\eqref{eq:kappa_apply}, with \(k\approx3\%\) of \(\dim\phi\) (\(k=50/100/200\) at \(N=5/10/20\); implementation details in Appendix~\ref{app:protocols}). We refer to the variation that retains this term as the \emph{sensitivity-enabled} gradient. The outer update is Adam with gradient-norm clipping rather than the plain step of Algorithm~\ref{alg:codesign}. The black-box baselines are \ac{bo}, a Gaussian-process surrogate with expected improvement, via BoTorch~\citep{balandat2020botorch} and a \((1{+}1)\)-\ac{es} with 1/5th-rule step-size adaptation~\citep{hansen2015evolution}.

\begin{figure}[t]
\centering
\includegraphics{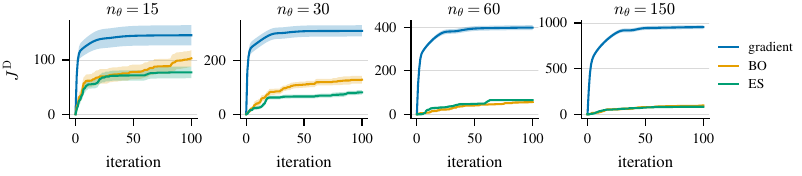}
\caption{\textbf{Comparison with \ac{bo} and \ac{es}} (with \(\lambda_y=0\), where \(\kappa_\theta=0\) by Lemma~\ref{lem:envelope}). Learning curves for \ac{bo}, \ac{es} and the gradient-based co-design method starting from the same initial design, for different design parameter sizes, over a budget of 100 design evaluations. Curves show the best co-design performance found so far relative to the shared initial design, \(J^{\mathrm{D}}-J^{\mathrm{D}}(\theta_0)\), with mean \(\pm\) SE over 25 trials; the percentage gaps quoted in the text are of the absolute final \(J^{\mathrm{D}}\) (Appendix~\ref{app:protocols}).}
\label{fig:benchmark}
\end{figure}

\paragraph{Comparison with BO and ES.}
We compare our gradient method with \ac{bo} and \ac{es} under the evaluation protocol of Appendix~\ref{app:protocols}: every method evaluates a design by training its policy with \ac{shac} and rolling out 200 steps, in a setting with only the direct design effect. This comparison is done to show the effect of the direct term in the gradient-based approach; therefore it was done on the configuration without any misalignment (\(\lambda_y=0\), so both design and policy rewards match and the objectives differ only by \(\rho\)), where the policy-sensitivity term is zero (Lemma~\ref{lem:envelope}). Figure~\ref{fig:benchmark} shows the results; the gradient method outperforms \ac{bo} and \ac{es} substantially, and this is more pronounced as \(N\) increases.

\begin{figure}[t]
\centering
\includegraphics[width=\linewidth]{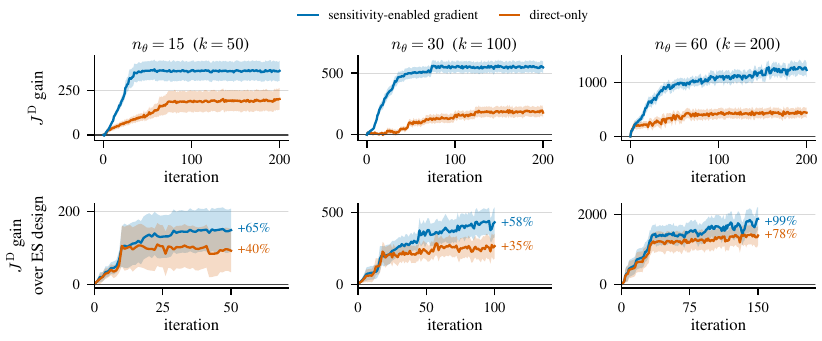}
\caption{{Top:} \textbf{Effect of the policy-sensitivity term} (misaligned configuration, \(\lambda_y=0.1\)). Learning curves for the gradient-based co-design with the {sensitivity-enabled gradient} and the direct-only (\(\kappa_\theta=0\)). The curves show the resulting co-design performance \(J^{\mathrm{D}}\) with mean \(\pm\) SE across the 15 trials. {Bottom:} \textbf{Refining the best \ac{es} design} (misaligned configuration, \(\lambda_y=0.1\)). Learning curves for the gradient-based co-design starting from the design found by a converged \ac{es} search, with the sensitivity-enabled gradient and the direct-only. The curves show the \(J^{\mathrm{D}}\) improvement over the \ac{es} design (the zero line), with mean \(\pm\) SE across the 15 trials; the annotated values are the final gains as \% of the mean \ac{es} performance.}
\label{fig:effect}
\label{fig:refinement}
\end{figure}

\paragraph{Effect of the policy-sensitivity term.}
Now we demonstrate the effect of the policy-sensitivity term \(\nabla_\phi J^{\mathrm{D}}\,\kappa_\theta\) in~\eqref{eq:outer_grad_chainrule} on the \emph{misaligned} configuration (\(\lambda_y=0.1\), plus a utilization cost and modified action maps that couple the optimal policy strongly to the design; Appendix~\ref{app:inv_spec}). The direct-only case differs only by setting \(\kappa_\theta=0\), and both cases start from the same initial conditions with shared design instances and seeds. The results in Figure~\ref{fig:effect} show that the policy-sensitivity term improves the co-design performance at every design dimension, and that its benefit grows with the dimension. More details in Appendices~\ref{app:protocols} and~\ref{app:ablation_tables}.

\paragraph{Refining the ES design.}
We next evaluate whether the proposed gradient method can refine designs found by black-box search, on the same misaligned configuration (\(\lambda_y=0.1\)). We chose \ac{es} as the baseline black-box search here because it produced better results than \ac{bo} in the misaligned setting: at \(N=5\) (600 evaluations each, same warm-started protocol), the best \ac{es} design outperforms \ac{bo}'s (\(+284\pm69\)). Each run starts from the best design of a converged \ac{es} search and its corresponding optimal policy. Figure~\ref{fig:refinement} reports the 50, 100 and 150 iterations of design refinement for \(N=5,10,20\), respectively. Both direct-only and {sensitivity-enabled} cases improve significantly on the \ac{es} designs at every design dimension, and {sensitivity-enabled gradient} outperforming direct-only (Appendix~\ref{app:ablation_tables}). This also serves as the comparison against black-box search under objective misalignment: the \ac{es} baseline is run to saturation before the gradient starts, so the gains reported here are obtained on top of a converged black-box search rather than against it under an equal budget. The difference between the {sensitivity-enabled gradient} and direct-only is smaller relative to the gains than in Figure~\ref{fig:effect}, which is expected: \ac{es} has already moved the design into the region where the policy has adapted, so part of the advantage of the policy-sensitivity term has been optimized by the \ac{es}.

\begin{wraptable}{r}{0.5\linewidth}
\vspace{-10pt}
\centering
\footnotesize
\caption{{Wall-clock per design evaluation (seconds).}}
\label{tab:cost}
{\begin{tabular}{lrrr}
\toprule
& \(n_\theta=15\) & \(30\) & \(60\) \\
\midrule
\ac{bo} & {2.7} & {2.8} & {3.2} \\
\ac{es} & {2.2} & {2.2} & {2.3} \\
direct-only & {2.0} & {2.0} & {2.1} \\
sensitivity-enabled, dense rank-\(k\) & {8.6} & {11.6} & {24.7} \\
sensitivity-enabled, matrix-free  & {6.4} & {6.4} & {6.7} \\
\bottomrule
\end{tabular}}
\vspace{-8pt}
\end{wraptable}
\paragraph{Computational cost.}
Table~\ref{tab:cost} reports the wall-clock per design evaluation of every method on the same GPU. Each evaluation trains the policy and scores the design; \ac{bo} adds its surrogate fit and \ac{es} nothing, so the black-box methods and the direct-only gradient has similar cost, as it is only one reverse-mode pass through the rollout. The policy-sensitivity term when built by the dense rank-\(k\) eigendecomposition used in Figure~\ref{fig:effect}, it costs {\(4\)--\(12\times\)} direct-only from \(n_\theta=15\) to \(60\); the matrix-free {solve~\eqref{eq:inner_ift_damped} of Algorithm~\ref{alg:codesign}, stopped after 200 matrix--vector products,} instead stays at about {\(6.5\)}\,s at every size, matches the dense variant's co-design performance over direct-only up to \(n_\theta=30\), and at \(n_\theta=60\) costs {a quarter} as much while recovering about three quarters of that performance (Appendix~\ref{app:protocols}).

%% file: figures/fig_inventory.tex
\begin{tikzpicture}[>=Latex, line width=0.5pt,
  every node/.style={font=\scriptsize, inner sep=1.5pt}]
\definecolor{figblue}{HTML}{0072B2}
\definecolor{figgreen}{HTML}{009E73}
\definecolor{figred}{HTML}{D55E00}

\node[draw, rounded corners=2pt, fill=black!5, minimum width=1.5cm,
      minimum height=0.75cm] (nd) at (0,0) {stock $q_i$};
\node[anchor=south] at (nd.north) {node $i$};

\draw[->, thick, figblue] (-2.6,0) -- (nd.west);
\node[anchor=south] at (-1.82,0.08)
  {supply $\textcolor{figblue}{s_i}\,\sigma(\textcolor{figgreen}{a^{\mathrm{sup}}_i})$};

\draw[->, thick] (nd.east) -- (2.85,0);
\node[anchor=north] at (2.02,-0.20)
  {serve $m_i$:\ $\textcolor{figblue}{b_i}\,\sigma(\textcolor{figgreen}{a^{\mathrm{srv}}_i})$};
\node[anchor=south, align=center] at (2.55,0.06) {revenue\\ $p_i m_i$};

\node[draw=figred, rounded corners=2pt, fill=figred!8] (bl) at (1.30,0.95)
  {backlog $y_i$ \textcolor{figred}{($-\lambda_y y_i$)}};
\draw[->] (1.30,1.62) -- (bl.north);
\node[anchor=east] at (1.05,1.50) {demand $d_i$ (AR(1))};
\draw[->, figred] (bl.south) -- (1.30,0.05);

\node[draw, rounded corners=2pt, minimum height=0.45cm] (jm) at (-1.80,-1.18) {$q_{i-1}$};
\node[draw, rounded corners=2pt, minimum height=0.45cm] (jp) at (1.80,-1.18) {$q_{i+1}$};
\node at (-2.55,-1.18) {$\cdots$};
\node at (2.55,-1.18) {$\cdots$};
\draw[<->, thick, figblue] (jm.north) -- (nd.south west);
\draw[<->, thick, figblue] (jp.north) -- (nd.south east);
\node at (0,-0.82)
  {$f_e=\textcolor{figblue}{c_e}\tanh(\textcolor{figgreen}{a^{\mathrm{rt}}_e})$};
\end{tikzpicture}

%% file: sections/discussion.tex
\section{Discussion}\label{sec:discussion}

The policy-sensitivity term, \(\kappa_\theta\) in the co-design gradient~\eqref{eq:outer_grad_chainrule}, is the central object of this work. It is the \emph{first-order} sensitivity of the optimal policy to the design. As shown in Lemma~\ref{lem:envelope} this term vanishes in the \emph{matched} case, where the design and policy rewards coincide up to a design-only cost \(\rho(\theta)\), and is generically nonzero in the \emph{mismatched} case, where they differ in their policy-dependent part. The experiments in Section~\ref{sec:benchmark_diffusion_codesign} confirm the effect of this policy-sensitivity term. With mismatched objectives, retaining the \(\kappa_\theta\) term produces better designs than the direct design effect alone, both in the ablation from a common initial design and when refining a design found by \ac{es}. Note that even in the matched case, the policy still adapts to the design, but its effect on the design objective is of second order, which we do not consider here because they are difficult to estimate in practice.

Lemma~\ref{lem:inner_sensitivity} assumes an exactly stationary inner solution with a nonsingular policy Hessian, whereas in practice \(\phi^\star(\theta)\) is never exactly stationary and the policy Hessian \(A\) is often ill-conditioned or indefinite, since overparameterized policies have flat or redundant parameter directions. The damped solve~\eqref{eq:inner_ift_damped}, or the rank truncation~\eqref{eq:kappa_apply} that plays the same role in our experiments, restores a well-posed system at the cost of a controllable bias in \(\kappa_\theta\) that grows with the damping \(\varepsilon\). These errors enter the design gradient linearly, so a partially converged inner policy optimization could still provide usable co-design gradient. An important practical implication is that the policy does \emph{not} need to be re-converged at every design update: warm-started tracking with finitely many inner steps is sufficient, and near a stationary design, where inner accuracy matters most, the outer steps are small and the target \(\phi^\star(\theta)\) moves slowly.

Like any first-order gradient method, the proposed co-design method offers no global-optimality guarantee, in exchange for sample efficiency and scalability to high-dimensional design spaces where \ac{bo} and \ac{es} degrade. The difference lies in what each method extracts from an evaluation. Black-box search extracts a single scalar from each design evaluation, whether to compare candidates as in \ac{es} or to condition a surrogate as in \ac{bo}, so the information gained per evaluation does not scale with the design dimension. The proposed gradient-based method instead differentiates the simulator and obtains a separate signal for every design parameter from the same rollouts, which is scalable with the design dimension. The two are therefore complementary, and combining them is the practical route to co-design at scale: a black-box search provides a good initial design, and the gradient refines it locally. The refinement experiment in Section~\ref{sec:benchmark_diffusion_codesign} demonstrates this combination, improving considerably on the design returned by a converged \ac{es} search.

The co-design gradient itself is not tied to pathwise \ac{rl}. The same decomposition into a direct design effect and a policy-sensitivity effect applies when the inner policy is trained with \ac{spg} methods, as demonstrated for morphology co-design with Stackelberg \ac{ppo}~\citep{dai2026stackelberg} and for simulator adaptation in sim-to-real \ac{rl}~\citep{anand2026bilevel}. The sensitivity estimate is then more challenging, since the stationarity condition contains score functions and a critic whose own sensitivities must be estimated, whereas in the differentiable-simulator setting the direct design effect is a single adjoint (reverse-mode) pass through the rollout, whose cost does not grow with the design dimension, and the second derivatives needed for \(\kappa_\theta\) are exact \ac{ad} operators on the same rollout.

\subsection{Limitations}\label{sec:limitations}

This paper provides the foundations of gradient-based co-design, the co-design gradient, the sensitivity of the inner solution, and their convergence analysis. However, Algorithm~\ref{alg:codesign} is only a proof of concept that shows the basic mechanism at work on a simple example. Turning it into a practical method for large policies and design spaces requires further algorithmic development, and the limitations below indicate where. The approach requires a differentiable simulator satisfying Assumption~\ref{ass:differentiable_simulator}: discrete design choices must be relaxed through smooth surrogate parameterizations, and hard contact or switching dynamics require smoothed models. Even with the \ac{svg}-style truncation, very stiff or chaotic systems can degrade the \ac{pd}. The sensitivity analysis rests on a well-conditioned inner optimum, which Appendix~\ref{app:convergence} formalizes as local strong concavity of the policy objective; overparameterized policies can violate this, which is why the solve for \(\kappa_\theta\) must be regularized in practice. Solving the inner loop repeatedly is the largest computational burden. Moreover, the sample efficiency that gradient-based co-design shows over black-box search does not necessarily translate into computational efficiency: the gradient adds second-order passes and a linear solve to each policy training, so one design evaluation could cost several times the wall-clock of a black-box evaluation.

%% file: sections/conclusion.tex
\section{Conclusion}\label{sec:conclusion}

We formulated co-design with a learned \ac{rl} policy as a bilevel problem and derived a gradient-based method for it. The co-design gradient splits into a direct design effect at a fixed policy and a policy-sensitivity effect through the design-dependent optimal policy. The latter follows from applying the \ac{ift} to the stationarity condition of a pathwise policy-gradient method. Based on this sensitivity, we developed a proof-of-concept co-design \ac{rl} algorithm that follows the \ac{svg}-style approach, computes the design gradient by adjoint (reverse-mode) differentiation of the short rollout. We prove that the resulting algorithm {reaches an approximately stationary design}. We demonstrated the effectiveness of the proposed method on a differentiable networked-system benchmark with high-dimensional design spaces, where it outperforms \ac{bo} and \ac{es} and further refines the designs they find. Scaling this proof of concept to practical applications is the main future work.

%% file: sections/appendix_codesign_grad.tex
\section{Derivations for the Pathwise Co-design Gradient}
\label{app:codesign_gradients}

Throughout this appendix, the closed-loop trajectory is generated by the reparameterized dynamics and policy of Assumption~\ref{ass:differentiable_simulator} at \(\phi=\phi^\star(\theta)\),
\[
a_t=g_{\phi^\star(\theta)}(s_t,\varepsilon_t),
\qquad
s_{t+1}=f_\theta(s_t,a_t,\xi_t),
\]
and the exogenous noise \(\omega=(s_0,\xi_{0:\infty},\varepsilon_{0:\infty})\) is independent of \(\theta\). We represent \(\kappa_\theta=\nabla_\theta\phi^\star(\theta)\) and use the total derivatives \(S_t=\mathrm{d}s_t/\mathrm{d}\theta\) and \(A_t=\mathrm{d}a_t/\mathrm{d}\theta\) along the sampled trajectory. The explicit (partial) derivative of a quantity with respect to \(\theta\) at fixed \((s,a)\) is represented as \(\partial_\theta(\cdot)\). All derivatives follow the row/Jacobian layout declared in Section~\ref{sec:outer_gradient_general} (rows indexed by outputs, so a scalar's derivative is a row vector); the adjoint displays below are written in the transposed column (\ac{ad}) form, as licensed there.

\subsection{Proof of Proposition~\ref{prop:pd_codesign_gradient}}
\label{app:pd_codesign_gradient}

We consider the reparameterized objective, which expectation over exogenous noise. Fix a finite horizon \(T\); the infinite-horizon result follows by \(T\to\infty\). Collect the randomness up to \(T\) in
\[
\omega_T \triangleq (s_0,\xi_0,\ldots,\xi_{T-1},\varepsilon_0,\ldots,\varepsilon_T),
\qquad
\mu_T(\mathrm{d}\omega_T)=\rho^0(\mathrm{d}s_0)\prod_{t=0}^{T-1}p_\xi(\mathrm{d}\xi_t)\prod_{t=0}^{T}p_\varepsilon(\mathrm{d}\varepsilon_t),
\]
where \(\rho^0\) is the initial-state law. Under the reparameterized maps of Assumption~\ref{ass:differentiable_simulator}, \((s_t,a_t)_{t=0}^{T}\) is a deterministic function of \((\theta,\omega_T)\). Define the truncated design return and objective
\[
G^{\mathrm{D}}_\theta(\omega_T)\triangleq\sum_{t=0}^{T}\gamma^t R^{\mathrm{D}}_\theta(s_t,a_t),
\qquad
J^{\mathrm{D}}_T(\theta)\triangleq\mathbb{E}_{\omega_T\sim\mu_T}\!\big[G^{\mathrm{D}}_\theta(\omega_T)\big].
\]
Since \(\mu_T\) does not depend on \(\theta\), dominated convergence permits interchange of differentiation and expectation:
\begin{equation}
\label{eq:app_pd_exchange}
\nabla_\theta J^{\mathrm{D}}_T(\theta)=\mathbb{E}_{\omega_T\sim\mu_T}\!\big[\nabla_\theta G^{\mathrm{D}}_\theta(\omega_T)\big].
\end{equation}
This is the pathwise estimator: it differentiates the trajectory map itself, not the log-density of the trajectory law as in the \ac{lr} estimator.

\paragraph{Forward-mode sensitivities.}
Differentiating the return along a sampled trajectory yields,
\begin{equation}
\label{eq:app_pd_grad_expand}
\nabla_\theta G^{\mathrm{D}}_\theta(\omega_T)
=\sum_{t=0}^{T}\gamma^t\Big(
\partial_\theta R^{\mathrm{D}}_\theta(s_t,a_t)
+\nabla_s R^{\mathrm{D}}_\theta(s_t,a_t)\,S_t
+\nabla_a R^{\mathrm{D}}_\theta(s_t,a_t)\,A_t\Big).
\end{equation}
Differentiating the policy and the dynamics of Assumption~\ref{ass:differentiable_simulator} gives the coupled recursions
\begin{align}
\label{eq:app_pd_action_rec}
A_t&=\nabla_s g_{\phi^\star(\theta)}(s_t,\varepsilon_t)\,S_t
+\nabla_\phi g_\phi(s_t,\varepsilon_t)\big|_{\phi^\star(\theta)}\,\kappa_\theta,\\
\label{eq:app_pd_state_rec}
S_{t+1}&=\partial_\theta f_\theta(s_t,a_t,\xi_t)
+\nabla_s f_\theta(s_t,a_t,\xi_t)\,S_t
+\nabla_a f_\theta(s_t,a_t,\xi_t)\,A_t,
\end{align}
with \(S_0=\mathrm{d}s_0/\mathrm{d}\theta\) (\(=0\) when \(\rho^0\) is \(\theta\)-independent). Equation~\eqref{eq:app_pd_action_rec} shows that the only bilevel-specific term is \(\kappa_\theta\); all other factors are standard Jacobians of the differentiable simulator and policy. Substituting~\eqref{eq:app_pd_action_rec}--\eqref{eq:app_pd_state_rec} into~\eqref{eq:app_pd_grad_expand} and inserting the result into~\eqref{eq:app_pd_exchange} yields the forward-mode statement of Proposition~\ref{prop:pd_codesign_gradient}. In forward mode, one simulates the trajectory and its sensitivities jointly: initialize \((s_0,S_0)\); at each \(t\) compute \((a_t,A_t)\) from~\eqref{eq:app_pd_action_rec}, advance to \(s_{t+1}\), propagate \(S_{t+1}\) by~\eqref{eq:app_pd_state_rec}, and accumulate~\eqref{eq:app_pd_grad_expand} in a single forward sweep. This sweep carries the matrix \(S_t\) (one column per design parameter), so its cost grows linearly with \(\dim\theta\); the adjoint form below avoids this dependence.

\paragraph{Reverse-mode / adjoint form.}
For high-dimensional \(\theta\), reverse-mode (backpropagation-through-time) is preferable. At each step the reparameterized trajectory is fixed by \emph{two} relations; the dynamics \(s_{t+1}=f_\theta(s_t,a_t,\xi_t)\) and the feedback policy \(a_t=g_{\phi^\star(\theta)}(s_t,\varepsilon_t)\). Hence, reverse mode carries \emph{two} adjoints, one attached to each of them: a costate and an action multiplier. Each is, by definition, the total sensitivity of the return to perturbing that quantity, equivalently the shadow price (in the Lagrangian view, the multiplier of that constraint). We accordingly define the costate \(\lambda_t\triangleq\nabla_{s_t}G^{\mathrm{D}}_\theta(\omega_T)=\mathrm{d}G^{\mathrm{D}}_\theta/\mathrm{d}s_t\), the sensitivity to the state, with \(\lambda_{T+1}\triangleq0\), and the action multiplier
\begin{equation}
\label{eq:app_pd_psi}
\psi_t\triangleq\gamma^t\nabla_a R^{\mathrm{D}}_\theta(s_t,a_t)+\big(\nabla_a f_\theta(s_t,a_t,\xi_t)\big)^{\!\top}\lambda_{t+1},
\end{equation}
which is the total sensitivity \(\mathrm{d}G^{\mathrm{D}}_\theta/\mathrm{d}a_t\) (the action \(a_t\) affects the return through the reward \(R^{\mathrm{D}}_\theta\) and the next state \(s_{t+1}\)). Since \(s_t\) influences the return directly through \(R^{\mathrm{D}}_\theta\), through the feedback action \(a_t=g_{\phi^\star(\theta)}(s_t,\varepsilon_t)\), and through \(s_{t+1}\), the costate satisfies, for \(t=T,\ldots,0\),
\begin{equation}
\label{eq:app_pd_costate}
\lambda_t=\gamma^t\nabla_s R^{\mathrm{D}}_\theta(s_t,a_t)
+\big(\nabla_s f_\theta(s_t,a_t,\xi_t)\big)^{\!\top}\lambda_{t+1}
+\big(\nabla_s g_{\phi^\star(\theta)}(s_t,\varepsilon_t)\big)^{\!\top}\psi_t.
\end{equation}
At the terminal step the dynamics Jacobians multiply \(\lambda_{T+1}=0\) and drop, so~\eqref{eq:app_pd_psi}--\eqref{eq:app_pd_costate} reduce to \(\psi_T=\gamma^T\nabla_a R^{\mathrm{D}}_\theta(s_T,a_T)\) and \(\lambda_T=\gamma^T\nabla_s R^{\mathrm{D}}_\theta(s_T,a_T)+(\nabla_s g_{\phi^\star(\theta)})^{\!\top}\psi_T\), which carry no \(f_\theta\) or \(\xi_T\) (no transition leaves \(s_T\)); the \(f_\theta\)-coupled recursions then run for \(t=T-1,\ldots,0\).
Each recursion is just the chain rule applied to the nodes its quantity feeds.

Equivalently, in terms of the per-step Hamiltonian \(H_t\triangleq\gamma^t R^{\mathrm{D}}_\theta(s_t,a_t)+\lambda_{t+1}^{\!\top}f_\theta(s_t,a_t,\xi_t)\), the two adjoints are \(\psi_t=\partial H_t/\partial a_t\) and \(\lambda_t=\partial H_t/\partial s_t+(\nabla_s g_{\phi^\star(\theta)})^{\!\top}\psi_t\): the closed-loop term \((\nabla_s g_{\phi^\star(\theta)})^{\!\top}\psi_t\) is what distinguishes the feedback policy from open-loop optimal control, where the action is not a function of the state (\(\nabla_s g=0\)), so this feedback is absent and \(\lambda_t=\partial H_t/\partial s_t\) (the action gradient \(\psi_t=\partial H_t/\partial a_t\) still exists as it is the control gradient, but no longer feeds back into the costate). Note that the action multiplier is the only adjoint beyond the costate, and it is \emph{eliminable}, substituting~\eqref{eq:app_pd_psi} into~\eqref{eq:app_pd_costate} yields a single \(\lambda\)-recursion. But we keep it explicit because the action is a node reused in both the costate and the policy-parameter gradient below; the two multipliers correspond to the two constraints of the rollout, the dynamics and the policy.

\paragraph{Adjoint-mode gradient.}
The costate \(\lambda_t=\mathrm{d}G^{\mathrm{D}}_\theta/\mathrm{d}s_t\) and action multiplier \(\psi_t=\mathrm{d}G^{\mathrm{D}}_\theta/\mathrm{d}a_t\) defined above are the \emph{total} sensitivities of the return to the state and action at time \(t\): each aggregates the entire downstream effect of a perturbation, propagated through all future states, actions, and rewards by the backward recursion~\eqref{eq:app_pd_psi}--\eqref{eq:app_pd_costate}. This is exactly the content of reverse-mode (adjoint) differentiation: once these node sensitivities are available, the total derivative of the return with respect to any parameter is obtained by summing, over every node at which the parameter enters \emph{explicitly}, the local partial derivative of that node's generating map contracted with the costate of the node it produces. Along the trajectory \(\theta\) enters explicitly at exactly three kinds of node:
\begin{itemize}
\item the reward \(R^{\mathrm{D}}_\theta(s_t,a_t)\), through \(\partial_\theta R^{\mathrm{D}}_\theta\), which enters the return directly with weight \(\gamma^t\);
\item the dynamics \(f_\theta(s_t,a_t,\xi_t)\), through \(\partial_\theta f_\theta\), which produces the next state \(s_{t+1}\) and is therefore weighted by its costate \(\lambda_{t+1}\);
\item the action \(a_t=g_{\phi^\star(\theta)}(s_t,\varepsilon_t)\), through its fixed-state design sensitivity \(\delta a_t=\nabla_\phi g_\phi(s_t,\varepsilon_t)\big|_{\phi^\star(\theta)}\,\kappa_\theta\) (the policy depends on \(\theta\) only via \(\phi^\star(\theta)\)), which produces \(a_t\) and is therefore weighted by its multiplier \(\psi_t\).
\end{itemize}
Summing the three contributions over \(t\) gives
\begin{align}
\label{eq:app_pd_decomp}
\nabla_\theta G^{\mathrm{D}}_\theta(\omega_T)
&=\underbrace{\sum_{t=0}^{T}\gamma^t\,\partial_\theta R^{\mathrm{D}}_\theta(s_t,a_t)
+\sum_{t=0}^{T-1}\big(\partial_\theta f_\theta(s_t,a_t,\xi_t)\big)^{\!\top}\lambda_{t+1}}_{\text{direct design effect}}\nonumber\\
&\quad+\underbrace{\sum_{t=0}^{T}\Big(\nabla_\phi g_\phi(s_t,\varepsilon_t)\big|_{\phi^\star(\theta)}\,\kappa_\theta\Big)^{\!\top}\psi_t}_{\text{policy-sensitivity effect}},
\end{align}
the \(\kappa_\theta\)-free terms forming the direct design effect and the \(\psi_t\)-term the policy-sensitivity effect. A further boundary convention is implicit here: we take the initial law \(\rho^0\) design-independent, so \(S_0=0\) and no initial-state term appears. For a reparameterized \(\theta\)-dependent initial state \(s_0=h_\theta(\chi)\) the forward form starts at \(S_0=\partial_\theta h_\theta\) and the reverse form starts from the term \((\partial_\theta h_\theta)^{\!\top}\lambda_0\).

\paragraph{Consistency with the chain rule.}
Taking \(\mathbb{E}_{\omega_T\sim\mu_T}\) of~\eqref{eq:app_pd_decomp} and comparing with the chain rule~\eqref{eq:outer_grad_chainrule} identifies the first group with the direct effect \(\partial_\theta J^{\mathrm{D}}_T(\phi^\star,\theta)\) and the second with \(\nabla_\phi J^{\mathrm{D}}_T(\phi^\star,\theta)\,\kappa_\theta\). Indeed, repeating the adjoint argument while differentiating with respect to \(\phi\) (which enters only through the actions) gives \(\nabla_\phi J^{\mathrm{D}}_T(\phi^\star,\theta)=\mathbb{E}_{\omega_T}\!\big[\sum_{t}(\nabla_\phi g_\phi(s_t,\varepsilon_t)|_{\phi^\star})^{\!\top}\psi_t\big]\), so the policy-sensitivity term is exactly \(\nabla_\phi J^{\mathrm{D}}_T(\phi^\star,\theta)\,\kappa_\theta\). This is the basis for the envelope property of Lemma~\ref{lem:envelope}: when \(R^{\mathrm{D}}_\theta=R^{\pi}_\theta\) and \(\phi^\star\) is stationary for \(J_\pi\), this expectation vanishes.

\paragraph{Infinite-horizon limit.}
Under the domination condition of Proposition~\ref{prop:pd_codesign_gradient}, which justifies interchanging the limit, the expectation, and differentiation, \(J^{\mathrm{D}}(\theta)=\lim_{T\to\infty}J^{\mathrm{D}}_T(\theta)-\rho(\theta)\) satisfies
\[
\nabla_\theta J^{\mathrm{D}}(\theta)=\mathbb{E}_{\omega\sim\mu}\!\big[\nabla_\theta G^{\mathrm{D}}_\theta(\omega)\big]-\nabla_\theta\rho(\theta),
\]
with \(\nabla_\theta G^{\mathrm{D}}_\theta(\omega)\) given by~\eqref{eq:app_pd_grad_expand} (forward) or~\eqref{eq:app_pd_decomp} (reverse) in the limit \(T\to\infty\). This proves Proposition~\ref{prop:pd_codesign_gradient}.

\subsection{Proof of Proposition~\ref{prop:svg_codesign_gradient} and Corollary~\ref{cor:shac_semigradient}}
\label{app:svg_codesign_gradient}

For the truncated objective~\eqref{eq:svg_objective}, collect \(\omega_H=(s_0,\xi_{0:H-1},\varepsilon_{0:H-1})\) with \(\theta\)-independent \(\mu_H\), and define the bootstrapped return
\[
\mathcal{G}_H^{\mathrm{D}}(\theta;\omega_H)=\sum_{t=0}^{H-1}\gamma^t R^{\mathrm{D}}_\theta(s_t,a_t)+\gamma^H V^{\mathrm{D}}_\theta(s_H).
\]
As in~\eqref{eq:app_pd_exchange}, \(\nabla_\theta J_H^{\mathrm{D}}(\theta)=\mathbb{E}_{\omega_H}\!\big[\nabla_\theta\mathcal{G}_H^{\mathrm{D}}(\theta;\omega_H)\big]-\nabla_\theta\rho(\theta)\); the design cost \(\rho(\theta)\) is deterministic and contributes the additive \(-\nabla_\theta\rho(\theta)\), which we omit in the per-sample expressions below and include back in the final gradient (as for the infinite-horizon objective in Proposition~\ref{prop:pd_codesign_gradient}).

\paragraph{Forward-mode.}
With \(S_t,A_t\) from~\eqref{eq:app_pd_action_rec}--\eqref{eq:app_pd_state_rec}, and using the total derivative of the terminal value \(\mathrm{d}V^{\mathrm{D}}_\theta(s_H)/\mathrm{d}\theta=\partial_\theta V^{\mathrm{D}}_\theta(s_H)+\nabla_s V^{\mathrm{D}}_\theta(s_H)\,S_H\),
\begin{align}
\label{eq:app_svg_forward}
\nabla_\theta\mathcal{G}_H^{\mathrm{D}}(\theta;\omega_H)
&=\sum_{t=0}^{H-1}\gamma^t\Big(
\partial_\theta R^{\mathrm{D}}_\theta(s_t,a_t)
+\nabla_s R^{\mathrm{D}}_\theta(s_t,a_t)\,S_t
+\nabla_a R^{\mathrm{D}}_\theta(s_t,a_t)\,A_t\Big)\nonumber\\
&\quad+\gamma^H\Big(\partial_\theta V^{\mathrm{D}}_\theta(s_H)+\nabla_s V^{\mathrm{D}}_\theta(s_H)\,S_H\Big),
\end{align}
which is the forward form of Proposition~\ref{prop:svg_codesign_gradient}.

\paragraph{Reverse-mode.}
Because the truncated return contains no action at \(t=H\) (the sum stops at \(H-1\)), the state \(s_H\) enters only through the terminal critic, giving the terminal costate
\[
\lambda_H=\gamma^H\nabla_s V^{\mathrm{D}}_\theta(s_H),
\]
and for \(t=H-1,\ldots,0\) the recursions~\eqref{eq:app_pd_psi}--\eqref{eq:app_pd_costate} hold with horizon \(H\). The gradient then decomposes as
\begin{align}
\label{eq:app_svg_reverse}
\nabla_\theta\mathcal{G}_H^{\mathrm{D}}(\theta;\omega_H)
&=\underbrace{\sum_{t=0}^{H-1}\gamma^t\,\partial_\theta R^{\mathrm{D}}_\theta(s_t,a_t)
+\sum_{t=0}^{H-1}\big(\partial_\theta f_\theta(s_t,a_t,\xi_t)\big)^{\!\top}\lambda_{t+1}
}_{\text{direct design effect}} +\gamma^H\partial_\theta V^{\mathrm{D}}_\theta(s_H)\nonumber\\
&\quad+\underbrace{\sum_{t=0}^{H-1}\Big(\nabla_\phi g_\phi(s_t,\varepsilon_t)\big|_{\phi^\star(\theta)}\,\kappa_\theta\Big)^{\!\top}\psi_t}_{\text{policy-sensitivity effect}}.
\end{align}
Taking \(\mathbb{E}_{\omega_H}\) and appending the design-cost term \(-\nabla_\theta\rho(\theta)\) through the identity \(\nabla_\theta J_H^{\mathrm{D}}=\mathbb{E}_{\omega_H}[\nabla_\theta\mathcal{G}_H^{\mathrm{D}}]-\nabla_\theta\rho\) above proves Proposition~\ref{prop:svg_codesign_gradient}. The decomposition matches the chain rule~\eqref{eq:outer_grad_chainrule}, now with the explicit critic--design term \(\gamma^H\partial_\theta V^{\mathrm{D}}_\theta(s_H)\) appearing in the direct effect; this is the only structural difference from the infinite-horizon case~\eqref{eq:app_pd_decomp}, and it reflects the design dependence of the bootstrapped tail value.

\paragraph{\ac{shac}-style semi-gradient.}
Replacing the terminal critic by a frozen, state-only target \(V^{+}(s_H)\)---the surrogate objective \(J^{\mathrm{D},+}_H\) of Corollary~\ref{cor:shac_semigradient}---removes the explicit design derivative, \(\partial_\theta V^{+}\equiv0\). The terminal contribution then reduces to \(\gamma^H\nabla_s V^{+}(s_H)\,S_H\) in the forward form~\eqref{eq:app_svg_forward} and to \(\lambda_H=\gamma^H\nabla_s V^{+}(s_H)\) in the reverse form~\eqref{eq:app_svg_reverse}, yielding the semi-gradient of Corollary~\ref{cor:shac_semigradient}. Gradients propagate through the terminal state but not through the explicit design derivative \(\partial_\theta V^{\mathrm{D}}_\theta\) of the terminal value function, consistent with practical \ac{shac} implementations.

\subsection{Critic sensitivity: proof of Lemma~\ref{lem:critic_design_sensitivity}}
\label{app:critic_design}

The terminal critic \(V^{\mathrm{D}}_\theta\) in~\eqref{eq:svg_objective} satisfies the reparameterized Bellman equation
\begin{equation}
\label{eq:app_design_bellman}
V^{\mathrm{D}}_\theta(s)
=
\mathbb{E}_{\varepsilon}\!\left[
R^{\mathrm{D}}_\theta(s,a)
+\gamma\,\mathbb{E}_{\xi}\!\left[V^{\mathrm{D}}_\theta(s')\right]
\right],
\qquad
a=g_{\phi^\star(\theta)}(s,\varepsilon),
\quad
s'=f_\theta(s,a,\xi),
\end{equation}
in which \(\theta\) enters through the design reward \(R^{\mathrm{D}}_\theta\), the dynamics \(f_\theta\), and the policy parameters \(\phi^\star(\theta)\). Differentiating this fixed point with respect to \(\theta\) at fixed state \(s\) yields the design sensitivity of the critic.

\begin{proof}[Proof of Lemma~\ref{lem:critic_design_sensitivity}]
Differentiate~\eqref{eq:app_design_bellman} with respect to \(\theta\) at fixed \(s\). The noise laws are \(\theta\)-independent, so no measure-derivative (score) term arises, and differentiation passes inside \(\mathbb{E}_{\varepsilon}\) and \(\mathbb{E}_{\xi}\) under the usual differentiation-under-the-integral-sign regularity: a locally uniform integrable bound on the \(\theta\)-derivative of the integrand, as in the  condition of Proposition~\ref{prop:pd_codesign_gradient}. The integrand depends on \(\theta\) explicitly through \(R^{\mathrm{D}}_\theta\) and through the design-parameterization of \(V^{\mathrm{D}}_\theta(s')\), and implicitly through the action \(a\) and next state \(s'\). Using \(\delta a=\nabla_\phi g_\phi(s,\varepsilon)\big|_{\phi^\star(\theta)}\,\kappa_\theta\) (the action depends on \(\theta\) only through \(\phi^\star(\theta)\)), \(\delta s'=\partial_\theta f_\theta(s,a,\xi)+\nabla_a f_\theta(s,a,\xi)\,\delta a\), and the total derivative \(\tfrac{\mathrm{d}}{\mathrm{d}\theta}V^{\mathrm{D}}_\theta(s')=\nabla_s V^{\mathrm{D}}_\theta(s')\,\delta s'+w_\theta(s')\), we obtain
\[
w_\theta(s)
=
\mathbb{E}_{\varepsilon}\!\left[
\partial_\theta R^{\mathrm{D}}_\theta(s,a)+\nabla_a R^{\mathrm{D}}_\theta(s,a)\,\delta a
+\gamma\,\mathbb{E}_{\xi}\!\left[\nabla_s V^{\mathrm{D}}_\theta(s')\,\delta s'+w_\theta(s')\right]
\right].
\]
Collecting the terms that do not contain \(w_\theta\) into the source \(b_\theta(s)\) gives the Bellman fixed point~\eqref{eq:critic_sensitivity_bellman}; rearranging \((I-\gamma P^{\pi}_\theta)w_\theta=b_\theta\) gives its operator form. The solve is well posed for \(\gamma<1\), since \(I-\gamma P^{\pi}_\theta\) is the same nonsingular operator that defines \(V^{\mathrm{D}}_\theta\).
\end{proof}

The term \(b_\theta\) contains the policy sensitivity \(\kappa_\theta\) through \(\delta a=\nabla_\phi g_\phi|_{\phi^\star(\theta)}\,\kappa_\theta\) (and \(\delta s'\)); consequently \(w_\theta=\partial_\theta V^{\mathrm{D}}_\theta\) is \emph{not} purely a direct effect. Therefore the critic term \(\gamma^H\partial_\theta V^{\mathrm{D}}_\theta(s_H)\) is not grouped with the direct effect in~\eqref{eq:app_svg_reverse}.

The fixed point~\eqref{eq:critic_sensitivity_bellman} is the co-design analogue of the stochastic value-gradient recursion of \citet{heess2015svg} (their Eq.~8): the recursive term \(\gamma\,\mathbb{E}[w_\theta(s')]\) is their \(\gamma V'_\theta\). As noted with Lemma~\ref{lem:critic_design_sensitivity}, \(w_\theta\) can therefore be learned by a  critic sensitivity function approximator \(W_\zeta\) on bootstrapped targets \(b_\theta(s)+\gamma W_\zeta(s')\).

\paragraph{Monte Carlo (unrolled)/ function approximator free form.}
Since \(P^{\pi}_\theta\) is a \(\gamma\)-contraction for \(\gamma<1\), and  the Neumann series is: \((I-\gamma P^{\pi}_\theta)^{-1}=\sum_{t\ge0}(\gamma P^{\pi}_\theta)^t\), and as \((P^{\pi}_\theta)^t b_\theta(s_0)=\mathbb{E}[b_\theta(s_t)\mid s_0]\), we get:
\begin{equation}
\label{eq:app_critic_mc}
w_\theta(s_0)=\sum_{t\ge0}(\gamma P^{\pi}_\theta)^t b_\theta(s_0)=\mathbb{E}\Big[\sum_{t\ge0}\gamma^t b_\theta(s_t)\;\Big|\;s_0\Big],
\end{equation}
which is the Monte Carlo form~\eqref{eq:critic_sensitivity_mc}. It is the same solution of the same fixed point~\eqref{eq:critic_sensitivity_bellman} that the bootstrapped critic sensitivity \(W_\zeta\) estimates, so the two estimators are consistent: the temporal-difference route solves the fixed point by regression on one-step targets, the Monte Carlo route by unrolling it along a sampled trajectory. Read as a backward recursion, \eqref{eq:app_critic_mc} gives the discounted stepwise accumulation
\begin{equation}
\label{eq:app_critic_reverse}
w_t=b_\theta(s_t)+\gamma\,w_{t+1},
\end{equation}
with \(w_N=0\) at the terminal state. The recursion sweeps a rolled-out episode once and returns \(w_t\approx\partial_\theta V^{\mathrm{D}}_\theta(s_t)\) at every visited state simultaneously. Truncating the sum after \(n\) steps and replacing the residual tail \(\gamma^n w_\theta(s_{t+n})\) by the learned critic sensitivity \(\gamma^n W_\zeta(s_{t+n})\) yields the \(n\)-step estimator
\begin{equation}
\label{eq:app_critic_nstep}
w_\theta(s_t)\approx\sum_{k=0}^{n-1}\gamma^k b_\theta(s_{t+k})+\gamma^n W_\zeta(s_{t+n}),
\end{equation}
which combines the bootstrapped and Monte Carlo approaches: \(n=0\) is the pure bootstrap \(W_\zeta(s_t)\), \(n\to\infty\) is the pure sum~\eqref{eq:app_critic_mc}, and the \ac{shac} semi-gradient case is \(W_\zeta\equiv0\) that drops the tail. Because \(b_\theta\) is built from the one-step \(\delta s'\) and the value-critic gradient \(\nabla_s V^{\mathrm{D}}_\theta\) rather than a product of state Jacobians, the variance of~\eqref{eq:app_critic_mc} is controlled by the discount, in contrast to the full pathwise sensitivity of Proposition~\ref{prop:pd_codesign_gradient}.

%% file: sections/appendix_shac_sensitivity.tex
\section{Sensitivity of SVG-style RL agent}
\label{app:inner_sensitivity}

This appendix proves Lemma~\ref{lem:inner_sensitivity} and Corollary~\ref{cor:inner_semigradient} by computing the two second-order derivatives of the \ac{svg}-style inner objective~\eqref{eq:Jsvg_inner} along a sampled rollout. The closed-loop maps are the reparameterized dynamics and policy of Assumption~\ref{ass:differentiable_simulator}, \(a_t=g_\phi(s_t,\varepsilon_t)\) and \(s_{t+1}=f_\theta(s_t,a_t,\xi_t)\), with policy reward \(R_\theta\equiv R^{\pi}_\theta\) and design-dependent critic \(V^{\pi}_\theta\). Again the noise is independent of \((\phi,\theta)\), so differentiation commutes with \(\mathbb{E}_\omega\). We define the sampled bootstrapped return
\[
G_H(\omega;\phi,\theta)=\sum_{t=0}^{H-1}\gamma^t R_\theta(s_t,a_t)+\gamma^H V^{\pi}_\theta(s_H),
\qquad
J^{\pi}_H(\phi,\theta)=\mathbb{E}_\omega[G_H].
\]
All Jacobians below are evaluated along the sampled trajectory; \(\partial_\theta(\cdot)\) denotes the explicit partial derivative at fixed \((s,a)\), and mixed second derivatives are written e.g. \(\nabla^2_{sa}R_\theta\), \(\partial^2_{\theta a}f_\theta\). All derivatives follow the row/Jacobian layout declared in Section~\ref{sec:outer_gradient_general} (rows indexed by outputs, so a scalar's derivative is a row vector); the adjoint displays below are written in the transposed column (\ac{ad}) form, as licensed there.

\subsection{First-order actor gradient}
\label{app:inner_first_order}
Define the policy-parameter sensitivities \(S^\phi_t\triangleq \mathrm{d}s_t/\mathrm{d}\phi\) and \(A^\phi_t\triangleq\mathrm{d}a_t/\mathrm{d}\phi\), with \(S^\phi_0=0\). Differentiating the policy and dynamics of Assumption~\ref{ass:differentiable_simulator} with respect to \(\phi\),
\begin{align}
\label{eq:in_Aphi}
A^\phi_t&=\nabla_s g_\phi(s_t,\varepsilon_t)\,S^\phi_t+\nabla_\phi g_\phi(s_t,\varepsilon_t),\\
\label{eq:in_Sphi}
S^\phi_{t+1}&=\nabla_s f_\theta(s_t,a_t,\xi_t)\,S^\phi_t+\nabla_a f_\theta(s_t,a_t,\xi_t)\,A^\phi_t.
\end{align}
The forward-mode actor gradient is
\begin{equation}
\label{eq:in_gradGphi_fwd}
\nabla_\phi G_H=\sum_{t=0}^{H-1}\gamma^t\big(\nabla_s R_\theta(s_t,a_t)\,S^\phi_t+\nabla_a R_\theta(s_t,a_t)\,A^\phi_t\big)+\gamma^H\nabla_s V^{\pi}_\theta(s_H)\,S^\phi_H .
\end{equation}
Equivalently, in reverse mode, the costate \(\lambda_t=\mathrm{d}G_H/\mathrm{d}s_t\) and action multiplier \(\psi_t=\mathrm{d}G_H/\mathrm{d}a_t\) are the same two adjoints (one for the dynamics, one for the policy) as in Appendix~\ref{app:pd_codesign_gradient}. By the same reasoning they obey, with terminal costate \(\lambda_H=\gamma^H\nabla_s V^{\pi}_\theta(s_H)\) and, for \(t=H-1,\ldots,0\),
\begin{align}
\label{eq:in_psi}
\psi_t&=\gamma^t\nabla_a R_\theta(s_t,a_t)+\big(\nabla_a f_\theta(s_t,a_t,\xi_t)\big)^{\!\top}\lambda_{t+1},\\
\label{eq:in_lambda}
\lambda_t&=\gamma^t\nabla_s R_\theta(s_t,a_t)+\big(\nabla_s f_\theta(s_t,a_t,\xi_t)\big)^{\!\top}\lambda_{t+1}+\big(\nabla_s g_\phi(s_t,\varepsilon_t)\big)^{\!\top}\psi_t.
\end{align}
The actor gradient now follows directly. The policy parameters \(\phi\) enter the rollout \emph{only} through the policy map \(g_\phi\): the reward \(R_\theta\), dynamics \(f_\theta\), and critic \(V^{\pi}_\theta\) carry no \(\phi\), so the only nodes \(\phi\) produces are the actions \(a_t\). By the adjoint parameter-gradient rule, the gradient with respect to a parameter is the sum, over the nodes that parameter directly feeds, of the explicit local Jacobian contracted with that node's adjoint. Hence the actor gradient is the policy Jacobian \(\nabla_\phi g_\phi\) contracted with the action adjoint \(\psi_t\), summed over the rollout. The state costate \(\lambda_t\) does \emph{not} appear explicitly, because \(\phi\) produces no state directly; it enters only \emph{indirectly}, through its presence in \(\psi_t\) via~\eqref{eq:in_psi}. Taking the expectation over the noise,
\begin{equation}
\label{eq:in_residual}
\nabla_\phi J^{\pi}_H(\phi,\theta)
=\mathbb{E}_\omega\Big[\sum_{t=0}^{H-1}\big(\nabla_\phi g_\phi(s_t,\varepsilon_t)\big)^{\!\top}\psi_t\Big]
\;\triangleq\;F(\phi,\theta),
\end{equation}
the reverse-mode equivalent of the forward actor gradient~\eqref{eq:in_gradGphi_fwd}. The converged inner actor \(\phi^\star(\theta)\) is a stationary point of the inner objective, i.e.\ \(\nabla_\phi J^{\pi}_H(\phi^\star(\theta),\theta)=0\); we call \(F\triangleq\nabla_\phi J^{\pi}_H\) the \emph{policy-gradient residual} because it vanishes exactly at \(\phi^\star(\theta)\) and measures the deviation from first-order optimality condition. The inner sensitivity \(\kappa_\theta\) is obtained by differentiating the optimality equation \(F(\phi^\star(\theta),\theta)=0\), and its two Jacobians \(\nabla_\phi F=\nabla^2_{\phi\phi}J^{\pi}_H\) and \(\nabla_\theta F=\nabla^2_{\theta\phi}J^{\pi}_H\) are the second-order terms computed in the next two subsections.

\subsection{Actor Hessian \texorpdfstring{\(\nabla^2_{\phi\phi}J^{\pi}_H\)}{the actor Hessian}}
\label{app:inner_hessphi}
Differentiating~\eqref{eq:in_residual} with respect to \(\phi\),
\begin{equation}
\label{eq:in_hessphi}
\nabla^2_{\phi\phi}J^{\pi}_H=\mathbb{E}_\omega\Big[\sum_{t=0}^{H-1}\Big(\nabla^2_{\phi\phi}g_\phi\, \psi_t+\big(\nabla^2_{s\phi}g_\phi\,S^\phi_t\big)^{\!\top}\psi_t+\big(\nabla_\phi g_\phi\big)^{\!\top}\nabla_\phi\psi_t\Big)\Big],
\end{equation}
 and the second-order adjoints are given by:
\begin{align}
\label{eq:in_dpsi_dphi}
\nabla_\phi\psi_t&=\gamma^t\big(\nabla^2_{sa}R_\theta\,S^\phi_t+\nabla^2_{aa}R_\theta\,A^\phi_t\big)+\big(\nabla^2_{sa}f_\theta\,S^\phi_t+\nabla^2_{aa}f_\theta\,A^\phi_t\big)^{\!\top}\lambda_{t+1}+\big(\nabla_a f_\theta\big)^{\!\top}\nabla_\phi\lambda_{t+1},\\
\label{eq:in_dlambda_dphi}
\nabla_\phi\lambda_t&=\gamma^t\big(\nabla^2_{ss}R_\theta\,S^\phi_t+\nabla^2_{as}R_\theta\,A^\phi_t\big)+\big(\nabla^2_{ss}f_\theta\,S^\phi_t+\nabla^2_{as}f_\theta\,A^\phi_t\big)^{\!\top}\lambda_{t+1}+\big(\nabla_s f_\theta\big)^{\!\top}\nabla_\phi\lambda_{t+1}\nonumber\\
&\quad+\big(\nabla^2_{ss}g_\phi\,S^\phi_t+\nabla^2_{s\phi}g_\phi\big)^{\!\top}\psi_t+\big(\nabla_s g_\phi\big)^{\!\top}\nabla_\phi\psi_t,
\end{align}
with terminal costate \(\nabla_\phi\lambda_H=\gamma^H\nabla^2_{ss}V^{\pi}_\theta(s_H)\,S^\phi_H\).

\subsection{Cross term \texorpdfstring{\(\nabla^2_{\theta\phi}J^{\pi}_H\)}{the cross block}}
\label{app:inner_cross}
Define the design parameter (\(\theta\)) sensitivities with fixed \(\phi\), \(S^\theta_t\triangleq\mathrm{d}s_t/\mathrm{d}\theta\) and \(A^\theta_t\triangleq\mathrm{d}a_t/\mathrm{d}\theta\),
\begin{align}
\label{eq:in_Atheta}
A^\theta_t&=\nabla_s g_\phi(s_t,\varepsilon_t)\,S^\theta_t,\\
\label{eq:in_Stheta}
S^\theta_{t+1}&=\partial_\theta f_\theta(s_t,a_t,\xi_t)+\nabla_s f_\theta(s_t,a_t,\xi_t)\,S^\theta_t+\nabla_a f_\theta(s_t,a_t,\xi_t)\,A^\theta_t,
\end{align}
with \(S^\theta_0=0\) (or \(S^\theta_0=\partial_\theta h_\theta(\chi)\) if the initial state is reparameterized as \(s_0=h_\theta(\chi)\)). Note \(A^\theta_t\) has no explicit term, since \(g_\phi\) carries no explicit \(\theta\). Differentiating the policy-gradient residual~\eqref{eq:in_residual} with respect to \(\theta\) at fixed \(\phi\),
\begin{equation}
\label{eq:in_cross}
\nabla^2_{\theta\phi}J^{\pi}_H=\mathbb{E}_\omega\Big[\sum_{t=0}^{H-1}\Big(\big(\nabla^2_{s\phi}g_\phi\,S^\theta_t\big)^{\!\top}\psi_t+\big(\nabla_\phi g_\phi\big)^{\!\top}\nabla_\theta\psi_t\Big)\Big],
\end{equation}
where, now including the explicit design dependence of \(R_\theta\) and \(f_\theta\),
\begin{align}
\label{eq:in_dpsi_dtheta}
\nabla_\theta\psi_t&=\gamma^t\big(\partial^2_{\theta a}R_\theta+\nabla^2_{sa}R_\theta\,S^\theta_t+\nabla^2_{aa}R_\theta\,A^\theta_t\big)\nonumber\\
&\quad+\big(\partial^2_{\theta a}f_\theta+\nabla^2_{sa}f_\theta\,S^\theta_t+\nabla^2_{aa}f_\theta\,A^\theta_t\big)^{\!\top}\lambda_{t+1}+\big(\nabla_a f_\theta\big)^{\!\top}\nabla_\theta\lambda_{t+1},\\
\label{eq:in_dlambda_dtheta}
\nabla_\theta\lambda_t&=\gamma^t\big(\partial^2_{\theta s}R_\theta+\nabla^2_{ss}R_\theta\,S^\theta_t+\nabla^2_{as}R_\theta\,A^\theta_t\big)\nonumber\\
&\quad+\big(\partial^2_{\theta s}f_\theta+\nabla^2_{ss}f_\theta\,S^\theta_t+\nabla^2_{as}f_\theta\,A^\theta_t\big)^{\!\top}\lambda_{t+1}+\big(\nabla_s f_\theta\big)^{\!\top}\nabla_\theta\lambda_{t+1}\nonumber\\
&\quad+\big(\nabla^2_{ss}g_\phi\,S^\theta_t\big)^{\!\top}\psi_t+\big(\nabla_s g_\phi\big)^{\!\top}\nabla_\theta\psi_t,
\end{align}
with terminal costate gradient
\begin{equation}
\label{eq:in_terminal_dtheta}
\nabla_\theta\lambda_H=\gamma^H\Big(\underbrace{\partial^2_{\theta s}V^{\pi}_\theta(s_H)}_{\text{critic--design term}}+\nabla^2_{ss}V^{\pi}_\theta(s_H)\,S^\theta_H\Big).
\end{equation}

\subsection{Proof of Lemma~\ref{lem:inner_sensitivity}}
\begin{proof}
By definition of the converged inner-loop actor, \(F(\phi^\star(\theta),\theta)=0\) with \(F=\nabla_\phi J^{\pi}_H\). Under the standard regularity conditions, \(F\) continuously differentiable, the stationary point locally isolated, and \(\nabla_\phi F=\nabla^2_{\phi\phi}J^{\pi}_H\) nonsingular, the \ac{ift} provides the sensitivity of the inner-loop actor parameters \(\phi^\star\) with respect to the design parameters as
\[
\nabla_\theta\phi^\star(\theta)=-\big(\partial_\phi F\big)^{-1}\partial_\theta F=-\big(\nabla^2_{\phi\phi}J^{\pi}_H\big)^{-1}\nabla^2_{\theta\phi}J^{\pi}_H,
\]
evaluated at \(\phi=\phi^\star(\theta)\). The actor Hessian is given by~\eqref{eq:in_hessphi} with recursions~\eqref{eq:in_dpsi_dphi}--\eqref{eq:in_dlambda_dphi}; the cross term is given by~\eqref{eq:in_cross} with recursions~\eqref{eq:in_dpsi_dtheta}--\eqref{eq:in_terminal_dtheta}. The explicit design terms \(\partial_\theta f_\theta\), \(\partial_\theta R_\theta\) (via \(\partial^2_{\theta a},\partial^2_{\theta s}\)) and the critic--design term \(\partial^2_{\theta s}V^{\pi}_\theta(s_H)\) appear only in the cross term, since the actor Hessian holds \(\theta\) fixed. This proves~\eqref{eq:inner_ift}.
\end{proof}

\subsection{Proof of Corollary~\ref{cor:inner_semigradient}}
\begin{proof}
Freezing the critic to a state-only target \(V^{\pi}_\theta(s_H)=V^{+}(s_H)\) as in \ac{shac} removes its explicit design dependence, so \(\partial^2_{\theta s}V^{\pi}_\theta(s_H)=0\) and the terminal condition~\eqref{eq:in_terminal_dtheta} reduces to \(\nabla_\theta\lambda_H=\gamma^H\nabla^2_{ss}V^{+}(s_H)\,S^\theta_H\). Similarly, the terminal costate of the first-order pass becomes \(\lambda_H=\gamma^H\nabla_s V^{+}(s_H)\), and the \(\phi\)-Hessian terminal becomes \(\nabla_\phi\lambda_H=\gamma^H\nabla^2_{ss}V^{+}(s_H)S^\phi_H\).
\end{proof}

\begin{remark}[Parallel with the outer gradient]
\label{rem:inner_outer_parallel}
The omitted term \(\gamma^H\partial^2_{\theta s}V^{\pi}_\theta(s_H)\) is the inner-loop counterpart of the outer-loop critic--design term \(\gamma^H\partial_\theta V^{\mathrm{D}}_\theta(s_H)\) of Proposition~\ref{prop:svg_codesign_gradient}: both are kept in the general \ac{svg} form and dropped in the \ac{shac} semi-gradient. As in Appendix~\ref{app:critic_design}, the design sensitivity of a bootstrapped value function can be recovered, by implicit differentiation of its Bellman fixed point if needed, but at the cost of additional computational complexity.
\end{remark}

\subsection{Implementation}\label{sec:shac_sensitivity_implementation}
Although the recursions above are written out in full for transparency, none of them is implemented by hand. In our JAX implementation the policy-gradient residual \(F=\nabla_\phi J^{\pi}_H\) is a single reverse-mode pass over the rollout; the actor Hessian \(A=\nabla^2_{\phi\phi}J^{\pi}_H\) and the cross block \(B=\nabla^2_{\theta\phi}J^{\pi}_H\) need never be formed densely: \(A\,v\) is available as a forward-over-reverse Hessian--vector product, and \(B\,v\), \(B^\top w\) as a \ac{jvp} and \ac{vjp} of \(F\) in the design direction. The only quantity that is \emph{not} a direct automatic-differentiation output is the sensitivity itself: \(\kappa_\theta=-A^{-1}B\) requires solving a linear system in \(A\), not forming \(A^{-1}\). For moderate size of \(\dim\phi\) (up to a few thousand), a standard direct factorization or truncated eigendecomposition of the densely formed \(A\) is practical, as in our inventory experiments (Appendix~\ref{app:example}); for larger policy networks, the damped solve~\eqref{eq:inner_ift_damped} is matrix-free on the operators \(A\) and \(B\). For a scalar outer objective, \(\kappa_\theta^\top u\) is assembled as one solve \(Ay=u\) followed by \(-B^\top y\), so neither \(\kappa_\theta\) nor \(A^{-1}\) is ever formed. More details on implementation are given in Appendix~\ref{app:implementation}.

%% file: sections/convergence_analysis.tex
\section{Convergence Analysis of the Bilevel Co-design Iteration}
\label{app:convergence}

In this appendix, we analyze the convergence of the bilevel co-design procedure of Algorithm~\ref{alg:codesign}. We consider that at each outer iteration \(k\), an inner solver warm-started from the previous policy returns \(\phi_k\approx\phi^\star(\theta_k)\), the co-design gradient estimate \(\hat g_k\) is formed at \((\phi_k,\theta_k)\), and the design is updated as \(\theta_{k+1}=\theta_k+\alpha\,\hat g_k\) (in the experiments the update is followed by the projection onto the design box; Remark~\ref{rem:conv_projection}). The analysis covers this plain step; the experiments instead use an adaptive outer optimizer, which changes how the gradient is turned into a step but leaves the direct design effect and the sensitivity solve unchanged (Section~\ref{sec:benchmark_diffusion_codesign}). We follow the viewpoint of the bilevel-optimization literature~\citep{ghadimi2018approximation,chen2021closing}, in which the outer iteration is a stochastic gradient method on the \emph{reduced} objective \(J^{\mathrm{D}}_H(\theta)=J^{\mathrm{D}}_H(\phi^\star(\theta),\theta)\) of~\eqref{eq:svg_objective}, and the inner loop only affects the gradient estimate through the error of the gradient estimate. The estimate \(\hat g_k\) differs from the co-design gradient~\eqref{eq:outer_grad_chainrule} by zero-mean sampling noise and by a \emph{bias} from three sources: the inner policy is not exactly at \(\phi^\star(\theta_k)\), the policy sensitivity is obtained from an approximate (damped) solver, and the terminal critics are used in semi-gradient form (Corollaries~\ref{cor:shac_semigradient} and~\ref{cor:inner_semigradient}). We first state the setting and assumptions in Appendix~\ref{app:conv_setting}, then establish the regularity of the reduced objective and bound each source of bias in Appendix~\ref{app:conv_bias}, and finally combine these bounds into a convergence result in Appendix~\ref{app:conv_theorem}. The coupling of inner and outer loops is provided in the tracking recursion of Lemma~\ref{lem:conv_tracking}: the outer step moves the inner target \(\phi^\star(\theta_k)\), and the inner error feeds back into the outer bias. This nested structure is why we use a tracking recursion rather than two-timescale stochastic approximation~\citep{borkar2008stochastic,hong2023twotimescale}, which suits interleaved single-step updates.

\subsection{Setting and Assumptions}
\label{app:conv_setting}

\paragraph{Notation.}
As in Section~\ref{appendix:shac_sensitivity_full}, we write \(A(\phi,\theta)\triangleq\nabla^2_{\phi\phi}J^{\pi}_H(\phi,\theta)\) for the policy Hessian and \(B(\phi,\theta)\triangleq\nabla^2_{\theta\phi}J^{\pi}_H(\phi,\theta)\) for the cross term, so that \(\kappa_\theta=-A^{-1}B\) at \(\phi=\phi^\star(\theta)\)~\eqref{eq:inner_ift}. The terminal critics \(V^{\pi}_\theta\) and \(V^{\mathrm{D}}_\theta\) are the given design-dependent functions that define the surrogates~\eqref{eq:Jsvg_inner} and~\eqref{eq:svg_objective}; their fitting is not analyzed. If \(V^{\mathrm{D}}_\theta\) is the exact value function of the closed loop \((\phi^\star(\theta),\theta)\) under \(R^{\mathrm{D}}_\theta\), then, with exogenous noise independent across steps and an integrable discounted return, the Markov and tower properties give \(J^{\mathrm{D}}_H(\theta)=J^{\mathrm{D}}(\phi^\star(\theta),\theta)\) of~\eqref{eq:J_design} for every \(H\): the outer truncation itself introduces no bias, and the results below then concern the design objective~\eqref{eq:J_design} at the inner solution of the \(H\)-step policy surrogate. The horizon still enters through that inner surrogate, through the constants of Assumption~\ref{ass:convergence}, and through the factor \(\gamma^H\) in the bias constants \(\delta_\kappa\) and \(\delta_H\) of Assumption~\ref{ass:convergence}(d). The designs are restricted to a compact convex set \(\Theta\). For \(\theta\in\Theta\), we denote by \(\mathbb{B}_r(\theta)\triangleq\{\phi:\|\phi-\phi^\star(\theta)\|\le r\}\) a ball around the inner solution and collect the admissible pairs in \(\mathcal{K}\triangleq\{(\phi,\theta):\theta\in\Theta,\ \phi\in\mathbb{B}_r(\theta)\}\). Following the \ac{ad} column convention of Section~\ref{sec:outer_gradient_general}, we identify row and column vectors and write \(\langle\cdot,\cdot\rangle\) for the Euclidean inner product; \(\|\cdot\|\) denotes the Euclidean norm of a vector and the spectral norm of a matrix. The inner tracking error is \(e_k\triangleq\phi_k-\phi^\star(\theta_k)\), and \(\mathcal{F}_k\) denotes the history of the algorithm up to and including the inner solve that produced \(\phi_k\).

\paragraph{Mean-field estimator.}
The gradient estimate \(\hat g_k\) of Algorithm~\ref{alg:codesign} is a sample average of pathwise derivatives. Its conditional mean given \(\mathcal{F}_k\) is
\begin{equation}
\label{eq:conv_meanfield}
\bar g(\phi,\theta)\triangleq
\underbrace{\partial_\theta J^{\mathrm{D}}_H(\phi,\theta)-\gamma^H\,\mathbb{E}_{\omega_H}\big[(w_\theta-W_\zeta)(s_H)\big]}_{\text{direct design effect as computed}}
\;+\;
\underbrace{\nabla_\phi J^{\mathrm{D}}_H(\phi,\theta)\,\tilde\kappa(\phi,\theta)}_{\text{policy-sensitivity effect as computed}},
\end{equation}
evaluated at \((\phi_k,\theta_k)\), where the rollout is generated by the current pair \((\phi,\theta)\). Here \(w_\theta=\partial_\theta V^{\mathrm{D}}_\theta\) is the critic sensitivity of Lemma~\ref{lem:critic_design_sensitivity} and \(W_\zeta\) its estimate, with \(W_\zeta\equiv0\) in the \ac{shac} semi-gradient of Corollary~\ref{cor:shac_semigradient}, and \(\tilde\kappa(\phi,\theta)\) is the policy sensitivity returned by the approximate solve at the current \((\phi,\theta)\), e.g.\ the damped solve~\eqref{eq:inner_ift_damped}, possibly with the semi-gradient cross term of Corollary~\ref{cor:inner_semigradient}. The first term is the direct design effect as it is actually computed, with the explicit design derivative of the critic replaced by its estimate, and the second term is the policy-sensitivity effect as actually computed. At \(\phi=\phi^\star(\theta)\), with \(W_\zeta=w_\theta\) and \(\tilde\kappa=\kappa_\theta\), \eqref{eq:conv_meanfield} reduces to the exact gradient~\eqref{eq:svg_codesign_gradient}.

\paragraph{Assumptions.}
The convergence analysis is local, it holds on the set \(\mathcal{K}\) of designs in \(\Theta\) and policies within distance \(r\) of the inner solution, where the inner objective is assumed to be strongly concave. The following assumption collects the required conditions.

\begin{assumption}[Regularity of the bilevel co-design problem]
\label{ass:convergence}
There exist constants \(r,m>0\), \(L_1,L_2,L_B,L_g,\delta_\kappa,\delta_H,\sigma^2,\nu\ge0\), and \(q\in(0,1)\) such that:
\begin{enumerate}
\item[(a)] \emph{Smoothness.} On an open set containing \(\mathcal{K}\), the design surrogate \(J^{\mathrm{D}}_H\) is \(C^1\) with an \(L_1\)-Lipschitz gradient in \((\phi,\theta)\), and the policy surrogate \(J^{\pi}_H\) is \(C^2\) with an \(L_2\)-Lipschitz Hessian in \((\phi,\theta)\) and \(\|B\|\le L_B\).
\item[(b)] \emph{Inner local strong stability.} For every \(\theta\in\Theta\), there is a stationary point \(\phi^\star(\theta)\) of \(J^{\pi}_H(\cdot,\theta)\)~\eqref{eq:inner_stationarity}, the map \(\theta\mapsto\phi^\star(\theta)\) is continuous on \(\Theta\), and \(A(\phi,\theta)\preceq-mI\) for all \(\phi\in\mathbb{B}_r(\theta)\).
\item[(c)] \emph{Inner solver.} Started from \(\phi\in\mathbb{B}_r(\theta)\) at the design \(\theta\), the inner solver returns a policy \(\phi'\) with \(\mathbb{E}\big[\|\phi'-\phi^\star(\theta)\|^2\,\big|\,\mathcal{G}\big]\le q\,\|\phi-\phi^\star(\theta)\|^2+\nu\), where \(\mathcal{G}\) is the history of the algorithm up to the start of the inner run, which determines \(\phi\) and \(\theta\).
\item[(d)] \emph{Outer estimator.} The estimate satisfies \(\hat g_k=\bar g(\phi_k,\theta_k)+M_{k+1}\) with \(\mathbb{E}[M_{k+1}\mid\mathcal{F}_k]=0\) and \(\mathbb{E}[\|M_{k+1}\|^2\mid\mathcal{F}_k]\le\sigma^2\). The mean field \(\bar g(\cdot,\theta)\) is \(L_g\)-Lipschitz on \(\mathbb{B}_r(\theta)\), and the approximate sensitivity satisfies \(\|\tilde\kappa(\phi^\star(\theta),\theta)-\kappa_\theta\|\le\delta_\kappa\) for all \(\theta\in\Theta\); and the critic-sensitivity estimate satisfies \(\gamma^H\sup_{\theta\in\Theta}\mathbb{E}_{\omega_H}\|(w_\theta-W_\zeta)(s_H)\|\le\delta_H\), with the rollout generated by \((\phi^\star(\theta),\theta)\).
\item[(e)] \emph{Localization.} All iterates, including the intermediate iterates of the inner solver, remain in \(\mathcal{K}\).
\end{enumerate}
\end{assumption}

Under Assumption~\ref{ass:differentiable_simulator}, condition (a) holds whenever \(f_\theta\), \(g_\phi\), \(R^{\pi}_\theta\), \(R^{\mathrm{D}}_\theta\), \(V^{\pi}_\theta\), and \(V^{\mathrm{D}}_\theta\) have bounded derivatives up to third order along the \(H\)-step rollout and the pathwise sensitivities have finite second moments, which is the finite-horizon form of the boundedness condition of Proposition~\ref{prop:pd_codesign_gradient}. Condition (b) adds to the nonsingular policy Hessian of Lemma~\ref{lem:inner_sensitivity} to negative definiteness on a neighborhood of the inner solution. The continuity of \(\phi^\star\) in (b) fixes one branch of local maximizers when the inner problem has several; it is the branch that the warm start of Algorithm~\ref{alg:codesign} follows in practice. It is the local form of the inner strong-convexity assumption that is standard in bilevel optimization~\citep{ghadimi2018approximation}, and it is what makes the inner problem contractive. Condition (c) abstracts the inner policy optimization, actor and critic updates together, into a contraction toward the current inner solution~\citep{bottou2018optimization}. In condition (d), \(\sigma^2\) is the second moment of the per-rollout pathwise sensitivities divided by the number of rollouts. Condition (d) also idealizes the estimate as conditionally unbiased for \(\bar g\). This is exact for the pathwise derivatives of the return under Assumption~\ref{ass:differentiable_simulator}, but the sensitivity \(\tilde\kappa\) is a nonlinear function of the sampled \(A\) and \(B\), so it carries an additional finite-sample bias that decreases with the number of rollouts and that we do not track. Conditions (c) and (d) are assumptions on the inner solver and on the estimator, not properties we prove for \ac{shac}; Theorem~\ref{thm:convergence} applies to any instance of Algorithm~\ref{alg:codesign} that satisfies them.  Condition (e) is the usual localization; in practice it is enforced for the design by projecting \(\theta\) onto the design box, and Remark~\ref{rem:conv_projection} states what the projection changes in the result below. Finally, we define the \emph{envelope factor}
\[
\Lambda\triangleq\sup_{\theta\in\Theta}\big\|\nabla_\phi J^{\mathrm{D}}_H(\phi^\star(\theta),\theta)\big\|,
\]
which is finite by (a) and the continuity in (b), and which vanishes under matched objectives as shown in Lemma~\ref{lem:envelope}.

\subsection{Regularity of the Reduced Objective and Bias Bounds}
\label{app:conv_bias}

We first show that the reduced objective is smooth, so that a gradient method on it makes progress. This is where the inner problem enters the outer objective: the inner solution \(\phi^\star(\theta)\) must be differentiable with a bounded sensitivity \(\kappa_\theta\), which is given by Lemma~\ref{lem:inner_sensitivity} and the additional condition (b).

\begin{lemma}[Regularity of the inner solution and of the reduced objective]
\label{lem:conv_regularity}
Let Assumption~\ref{ass:convergence}(a)--(b) hold. Then:
(i) \(\phi^\star(\theta)\) is the unique stationary point of \(J^{\pi}_H(\cdot,\theta)\) in \(\mathbb{B}_r(\theta)\), and it is a strict local maximizer;
(ii) the inner solution \(\theta\mapsto\phi^\star(\theta)\) is \(C^1\) on \(\Theta\) with \(\kappa_\theta=-A^{-1}B\) and \(\|\kappa_\theta\|\le L_\star\triangleq L_B/m\), so that \(\|\phi^\star(\theta)-\phi^\star(\theta')\|\le L_\star\|\theta-\theta'\|\);
(iii) the reduced objective \(J^{\mathrm{D}}_H(\theta)\) is \(C^1\) with the gradient~\eqref{eq:outer_grad_chainrule},
\(\nabla_\theta J^{\mathrm{D}}_H(\theta)=\partial_\theta J^{\mathrm{D}}_H(\phi^\star,\theta)+\nabla_\phi J^{\mathrm{D}}_H(\phi^\star,\theta)\,\kappa_\theta\), and this gradient is \(L_J\)-Lipschitz on \(\Theta\) with
\[
L_J=(1+L_\star)\Big[L_1(1+L_\star)+\Lambda L_2\,\frac{m+L_B}{m^2}\Big].
\]
\end{lemma}
\begin{proof}
\emph{(i)} On the convex ball \(\mathbb{B}_r(\theta)\), the condition \(A\preceq-mI\) gives \(\langle\nabla_\phi J^{\pi}_H(\phi,\theta)-\nabla_\phi J^{\pi}_H(\phi',\theta),\phi-\phi'\rangle\le-m\|\phi-\phi'\|^2\). Thus \(J^{\pi}_H(\cdot,\theta)\) is \(m\)-strongly concave on the ball, has at most one stationary point there, and that point is its maximizer.

\emph{(ii)} Fix \(\theta_0\in\Theta\). Since \(A(\phi^\star(\theta_0),\theta_0)\) is nonsingular, the \ac{ift}~\citep{krantz2002implicit} applied to the stationarity condition~\eqref{eq:inner_stationarity}, as in Lemma~\ref{lem:inner_sensitivity}, gives a neighborhood of \(\theta_0\) and a \(C^1\) map \(\psi\) on it with \(\psi(\theta_0)=\phi^\star(\theta_0)\) and \(F(\psi(\theta),\theta)=0\). By continuity of \(\phi^\star\) and \(\psi\), for \(\theta\) close to \(\theta_0\) both \(\psi(\theta)\) and \(\phi^\star(\theta)\) lie in \(\mathbb{B}_r(\theta)\), where the stationary point is unique by (i); hence \(\psi=\phi^\star\) near \(\theta_0\), so \(\phi^\star\in C^1\) on \(\Theta\) with \(\kappa_\theta=-A^{-1}B\). The bound follows from \(\|A^{-1}\|\le1/m\) and \(\|B\|\le L_B\), and the Lipschitz estimate from the mean-value inequality on the convex set \(\Theta\).

\emph{(iii)} The chain rule gives the gradient. For \(\theta,\theta'\in\Theta\), let \(d\triangleq\|\theta-\theta'\|\). By (ii), the points \((\phi^\star(\theta),\theta)\) and \((\phi^\star(\theta'),\theta')\) are at most \((1+L_\star)d\) apart, so the two partial gradients of \(J^{\mathrm{D}}_H\) change by at most \(L_1(1+L_\star)d\) between them. Writing \(A^{-1}B-A'^{-1}B'=A^{-1}(B-B')+A^{-1}(A'-A)A'^{-1}B'\) and using \(\|A^{-1}\|,\|A'^{-1}\|\le1/m\), the sensitivity changes by at most \(L_2(1+L_\star)(1/m+L_B/m^2)\,d\). Combining these three estimates with \(\|\kappa_{\theta'}\|\le L_\star\) and \(\|\nabla_\phi J^{\mathrm{D}}_H(\phi^\star,\theta)\|\le\Lambda\) yields \(L_J\).
\end{proof}

The factor \(1+L_\star\) in \(L_J\) records the two routes through which the design acts on the objective: directly through the environment, and through the re-optimized policy with weight \(\|\kappa_\theta\|\le L_\star\). The next lemma bounds the error of the approximate sensitivity solve.

\begin{lemma}[Approximate sensitivity: damping and semi-gradient]
\label{lem:conv_damping}
Let \((\phi^\star(\theta),\theta)\in\mathcal{K}\) and \(\varepsilon\ge0\). (i) The damped sensitivity \(\kappa_\varepsilon\triangleq-(A-\varepsilon I)^{-1}B\) of the solve~\eqref{eq:inner_ift_damped} satisfies
\[
\|\kappa_\varepsilon-\kappa_\theta\|\le\frac{\varepsilon\,L_B}{m\,(m+\varepsilon)} .
\]
(ii) If in addition the cross term is replaced by the semi-gradient \(\hat B\) of Corollary~\ref{cor:inner_semigradient}, with the frozen critic \(V^{+}\) equal to \(V^{\pi}_\theta\) at the current design so that \(A\) is unchanged, the resulting \(\tilde\kappa\triangleq-(A-\varepsilon I)^{-1}\hat B\) satisfies
\[
\|\tilde\kappa-\kappa_\theta\|\le\frac{\varepsilon\,L_B}{m\,(m+\varepsilon)}+\frac{\gamma^H c_V}{m+\varepsilon},
\qquad
c_V\triangleq\sup_{\theta\in\Theta}\gamma^{-H}\big\|B-\hat B\big\|,
\]
with \(B\) and \(\hat B\) evaluated at \((\phi^\star(\theta),\theta)\). Here \(B-\hat B\) is the critic--design term \(\gamma^H\partial^2_{\theta s}V^{\pi}_\theta(s_H)\) propagated backward through the adjoint recursions~\eqref{eq:in_dpsi_dtheta}--\eqref{eq:in_terminal_dtheta}. Writing \(\Phi_t\triangleq\nabla_s f_\theta+\nabla_a f_\theta\,\nabla_s g_\phi\) for the closed-loop Jacobian at step \(t\), so that \(\partial s_H/\partial s_{t+1}=\Phi_{H-1}\cdots\Phi_{t+1}\), this gives
\[
c_V\le C_1\sum_{t=0}^{H-1}\sup_{\omega_H}\big\|\Phi_{H-1}\cdots\Phi_{t+1}\big\|,
\]
where \(C_1\) bounds \(\|\nabla_\phi g_\phi\|\,\|\nabla_a f_\theta\|\,\|\partial^2_{\theta s}V^{\pi}_\theta\|\) on \(\mathcal{K}\) and the empty product is the identity. If the state-transition Jacobians satisfy \(\|\Phi_{H-1}\cdots\Phi_{t+1}\|\le1\) along the rollouts, then \(c_V\le C_1H\). If they grow geometrically, so does \(c_V\); this is the long-rollout issue of Section~\ref{sec:outer_gradient_general}. Hence \(\delta_\kappa\) in Assumption~\ref{ass:convergence}(d) can be taken as the right-hand side above.
\end{lemma}
\begin{proof}
\emph{(i)} We have \(\kappa_\varepsilon-\kappa_\theta=\big[A^{-1}-(A-\varepsilon I)^{-1}\big]B=A^{-1}\big[(A-\varepsilon I)-A\big](A-\varepsilon I)^{-1}B=-\varepsilon A^{-1}(A-\varepsilon I)^{-1}B\). Since \(A\preceq-mI\), we have \(\|A^{-1}\|\le1/m\) and \(\|(A-\varepsilon I)^{-1}\|\le1/(m+\varepsilon)\), which gives the bound.

\emph{(ii)} With \(V^{+}=V^{\pi}_\theta\) at the current design, the two surrogates coincide as functions of \(\phi\), so \(A\) is unchanged and \(\tilde\kappa-\kappa_\varepsilon=-(A-\varepsilon I)^{-1}(\hat B-B)\), whose norm is at most \(\|B-\hat B\|/(m+\varepsilon)\le\gamma^H c_V/(m+\varepsilon)\); adding (i) gives the bound. By Corollary~\ref{cor:inner_semigradient}, the only term dropped from \(B\) is the critic--design term \(\gamma^H\partial^2_{\theta s}V^{\pi}_\theta(s_H)\) of the terminal condition~\eqref{eq:in_terminal_dtheta}. Since the recursions~\eqref{eq:in_dpsi_dtheta}--\eqref{eq:in_terminal_dtheta} are linear in this terminal term and the first-order adjoints are the same for both cross terms, the differences \(\Delta\lambda_t\triangleq\nabla_\theta\lambda_t-\nabla_\theta\hat\lambda_t\) and \(\Delta\psi_t\triangleq\nabla_\theta\psi_t-\nabla_\theta\hat\psi_t\) satisfy \(\Delta\lambda_H=\gamma^H\partial^2_{\theta s}V^{\pi}_\theta(s_H)\), \(\Delta\psi_t=(\nabla_a f_\theta)^{\!\top}\Delta\lambda_{t+1}\), and \(\Delta\lambda_t=(\nabla_s f_\theta)^{\!\top}\Delta\lambda_{t+1}+(\nabla_s g_\phi)^{\!\top}\Delta\psi_t=\Phi_t^{\!\top}\Delta\lambda_{t+1}\). By~\eqref{eq:in_cross}, \(\Delta\lambda_{t+1}=(\Phi_{H-1}\cdots\Phi_{t+1})^{\!\top}\Delta\lambda_H\) and \(B-\hat B=\mathbb{E}_{\omega_H}\big[\sum_{t=0}^{H-1}(\nabla_\phi g_\phi)^{\!\top}(\nabla_a f_\theta)^{\!\top}(\Phi_{H-1}\cdots\Phi_{t+1})^{\!\top}\gamma^H\partial^2_{\theta s}V^{\pi}_\theta(s_H)\big]\), and taking norms gives the bound on \(c_V\).
\end{proof}

The bias of the mean-field estimator~\eqref{eq:conv_meanfield} can now be decomposed into its three sources. As in the error analysis of approximate implicit differentiation~\citep{grazzi2020iteration}, the bias is linear in the error of the inner solution and in the error of the linear solve.

\begin{lemma}[Bias decomposition]
\label{lem:conv_bias}
Under Assumption~\ref{ass:convergence}, the bias \(b_k\triangleq\bar g(\phi_k,\theta_k)-\nabla_\theta J^{\mathrm{D}}_H(\theta_k)\) of the estimate satisfies
\[
\|b_k\|\le L_g\|e_k\|+\Lambda\,\delta_\kappa+\delta_H,
\]
with \(\delta_\kappa\) and \(\delta_H\) the constants of Assumption~\ref{ass:convergence}(d).
\end{lemma}
\begin{proof}
We split the bias as \(b_k=\big[\bar g(\phi_k,\theta_k)-\bar g(\phi^\star(\theta_k),\theta_k)\big]+\big[\bar g(\phi^\star(\theta_k),\theta_k)-\nabla_\theta J^{\mathrm{D}}_H(\theta_k)\big]\). The first bracket is the effect of the inexact inner solution and is bounded by \(L_g\|e_k\|\) by Assumption~\ref{ass:convergence}(d). By~\eqref{eq:conv_meanfield} and Lemma~\ref{lem:conv_regularity}(iii), the second bracket equals \(-\gamma^H\mathbb{E}_{\omega_H}[(w_\theta-W_\zeta)(s_H)]+\nabla_\phi J^{\mathrm{D}}_H(\phi^\star,\theta)\big(\tilde\kappa(\phi^\star,\theta)-\kappa_\theta\big)\), whose norm is at most \(\delta_H+\Lambda\delta_\kappa\) by Assumption~\ref{ass:convergence}(d).
\end{proof}

For \(W_\zeta\equiv0\), the term \(\gamma^H\mathbb{E}_{\omega_H}[w_\theta(s_H)]\) is exactly the gap between the \ac{shac} semi-gradient of Corollary~\ref{cor:shac_semigradient} and the gradient~\eqref{eq:svg_codesign_gradient}. If the source \(b_\theta\) of Lemma~\ref{lem:critic_design_sensitivity} is bounded, the Neumann series~\eqref{eq:critic_sensitivity_bellman} gives \(\|w_\theta\|_\infty\le\|b_\theta\|_\infty/(1-\gamma)\), so that \(\delta_H\) can be taken as \(\gamma^H\|b_\theta\|_\infty/(1-\gamma)\). The omitted critic term is thus discounted \(H\) times, which is the mechanism that makes the semi-gradient accurate.

Now we analyze the coupling between the two loops. The inner error \(e_k\) drives the outer bias through Lemma~\ref{lem:conv_bias}, and the outer step moves the inner target \(\phi^\star(\theta_k)\). The following lemma shows that the inner solver keeps tracking this moving target.

\begin{lemma}[Inner tracking under warm starts]
\label{lem:conv_tracking}
Let Assumption~\ref{ass:convergence} hold, let \(\Gamma^2\triangleq\sup_{\mathcal{K}}\|\bar g\|^2+\sigma^2\), which bounds \(\mathbb{E}\|\hat g_k\|^2\), and let \(\tilde q\triangleq(1+q)/2\). Then
\[
\mathbb{E}\|e_{k+1}\|^2\le\tilde q\,\mathbb{E}\|e_k\|^2+c\,\alpha^2+\nu,
\qquad c\triangleq\frac{q(1+q)}{1-q}\,L_\star^2\Gamma^2,
\]
and consequently \(\mathbb{E}\|e_k\|^2\le\tilde q^{\,k}\,\mathbb{E}\|e_0\|^2+\bar e^2\) with \(\bar e^2\triangleq2(c\,\alpha^2+\nu)/(1-q)\).
\end{lemma}
\begin{proof}
The policy \(\phi_{k+1}\) is returned by the inner solver run at the new design \(\theta_{k+1}\) from the warm start \(\phi_k\). By Assumption~\ref{ass:convergence}(c) and (e), conditionally on \(\mathcal{F}_k\) and \(M_{k+1}\), which determine \(\theta_{k+1}\), we have \(\mathbb{E}\|e_{k+1}\|^2\le q\|\phi_k-\phi^\star(\theta_{k+1})\|^2+\nu\). The target has moved: by Lemma~\ref{lem:conv_regularity}(ii), \(\|\phi_k-\phi^\star(\theta_{k+1})\|\le\|e_k\|+L_\star\alpha\|\hat g_k\|\). Young's inequality \(\|x+y\|^2\le(1+\eta)\|x\|^2+(1+1/\eta)\|y\|^2\) with \(\eta=(1-q)/(2q)\) gives \(q(1+\eta)=\tilde q\) and \(q(1+1/\eta)=q(1+q)/(1-q)\). Taking expectations and using \(\mathbb{E}\|\hat g_k\|^2=\mathbb{E}\|\bar g(\phi_k,\theta_k)\|^2+\mathbb{E}\|M_{k+1}\|^2\le\Gamma^2\) yields the recursion. Unrolling it with \(\sum_{j\ge0}\tilde q^{\,j}=2/(1-q)\) gives the bound.
\end{proof}

The \(\alpha^2\) term is the price of the moving target, and it is the bilevel content of the lemma. The inner loop need not be re-converged from scratch at every design: warm starts with a fixed contraction \(q\) keep the tracking error at a level that shrinks with the outer step size and with the inner noise floor \(\nu\).

\subsection{Convergence Result}
\label{app:conv_theorem}

Combining the smoothness of the reduced objective with the bias bounds gives the following convergence result for Algorithm~\ref{alg:codesign}.

\begin{theorem}[Convergence of the bilevel co-design iteration]
\label{thm:convergence}
Let Assumption~\ref{ass:convergence} hold and let the outer step size satisfy \(\alpha\le1/(4L_J)\). Then, for every \(N\ge1\),
\begin{equation}
\label{eq:conv_main_bound}
\frac1N\sum_{k=0}^{N-1}\mathbb{E}\big\|\nabla_\theta J^{\mathrm{D}}_H(\theta_k)\big\|^2
\;\le\;
\frac{4\big(J^\star-J^{\mathrm{D}}_H(\theta_0)\big)}{\alpha N}
+2L_J\alpha\sigma^2
+9\Big[L_g^2\Big(\bar e^2+\frac{2\,\mathbb{E}\|e_0\|^2}{(1-q)N}\Big)+\Lambda^2\delta_\kappa^2+\delta_H^2\Big],
\end{equation}
where \(J^\star\triangleq\sup_{\Theta}J^{\mathrm{D}}_H\).
\end{theorem}
\begin{proof}
We abbreviate \(J\triangleq J^{\mathrm{D}}_H\) and \(\nabla J_k\triangleq\nabla_\theta J(\theta_k)\), and write the estimate as \(\hat g_k=\nabla J_k+b_k+M_{k+1}\), where the bias \(b_k\) is \(\mathcal{F}_k\)-measurable. Since \(\nabla J\) is \(L_J\)-Lipschitz on the convex set \(\Theta\) by Lemma~\ref{lem:conv_regularity}, the ascent lemma gives \(J(\theta_{k+1})\ge J(\theta_k)+\alpha\langle\nabla J_k,\hat g_k\rangle-\tfrac{L_J}{2}\alpha^2\|\hat g_k\|^2\). Taking the conditional expectation given \(\mathcal{F}_k\) and using Assumption~\ref{ass:convergence}(d), we get \(\mathbb{E}[\langle\nabla J_k,\hat g_k\rangle\mid\mathcal{F}_k]=\|\nabla J_k\|^2+\langle\nabla J_k,b_k\rangle\ge\tfrac12\|\nabla J_k\|^2-\tfrac12\|b_k\|^2\) and \(\mathbb{E}[\|\hat g_k\|^2\mid\mathcal{F}_k]\le2\|\nabla J_k\|^2+2\|b_k\|^2+\sigma^2\). Hence, for \(\alpha\le1/(4L_J)\),
\[
\mathbb{E}[J(\theta_{k+1})\mid\mathcal{F}_k]\ \ge\ J(\theta_k)+\tfrac{\alpha}{4}\|\nabla J_k\|^2-\tfrac{3\alpha}{4}\|b_k\|^2-\tfrac{L_J}{2}\alpha^2\sigma^2 .
\]
Taking full expectations, summing over \(k<N\), telescoping with \(J(\theta_N)\le J^\star\), and dividing by \(\alpha N/4\) gives
\[
\frac1N\sum_{k<N}\mathbb{E}\|\nabla J_k\|^2\le\frac{4(J^\star-J(\theta_0))}{\alpha N}+2L_J\alpha\sigma^2+\frac3N\sum_{k<N}\mathbb{E}\|b_k\|^2 .
\]
By Lemma~\ref{lem:conv_bias}, \(\|b_k\|^2\le3\big(L_g^2\|e_k\|^2+\Lambda^2\delta_\kappa^2+\delta_H^2\big)\), and by Lemma~\ref{lem:conv_tracking}, \(\frac1N\sum_{k<N}\mathbb{E}\|e_k\|^2\le\bar e^2+\frac{2\,\mathbb{E}\|e_0\|^2}{(1-q)N}\). Substituting these two bounds gives~\eqref{eq:conv_main_bound}.
\end{proof}

The first two terms of~\eqref{eq:conv_main_bound} are those of stochastic gradient ascent on a smooth nonconvex objective~\citep{ghadimi2013stochastic}: the optimization term decays as \(1/(\alpha N)\), and the variance term is proportional to \(\alpha\). The third term is the bias floor of a stochastic gradient method with biased gradients~\citep{ajalloeian2020convergence}. Here it has three named sources: the inner tracking error \(\bar e^2\), which shrinks with \(\alpha^2\) and with the inner noise floor \(\nu\); the approximate sensitivity \(\delta_\kappa\le\varepsilon L_B/(m(m+\varepsilon))+\gamma^H c_V/(m+\varepsilon)\) by Lemma~\ref{lem:conv_damping}; and the semi-gradient critic term \(\delta_H\), which carries the factor \(\gamma^H\) by Lemma~\ref{lem:conv_bias}. Thus the design updates reach an approximately stationary design of the surrogate \(J^{\mathrm{D}}_H\), with the accuracy set by these three terms: the bound applies to the average over the iterations, and hence to the best iterate, rather than to the limit of the sequence. At a fixed horizon \(H\), so that the constants of Assumption~\ref{ass:convergence} are fixed, the tracking term vanishes as \(\alpha\to0\) and \(\nu\to0\) with \(q\) held fixed (for the gradient-ascent instance of (c), \(\tau\to0\) with \(K\propto1/\tau\)), the sensitivity term vanishes as \(\varepsilon\to0\) with the full cross term, and the critic term vanishes as \(W_\zeta\to w_\theta\); the horizon enters only through the factor \(\gamma^H\), and the limit \(H\to\infty\) is not taken here because the constants of Assumption~\ref{ass:convergence} depend on \(H\). If all three bias terms are zero, an idealization that requires an exact inner solution, an exact sensitivity solve, and the exact critic sensitivity of Lemma~\ref{lem:critic_design_sensitivity}, then the choice \(\alpha=1/\sqrt N\) for \(N\ge16L_J^2\) gives \(\min_{k<N}\mathbb{E}\|\nabla_\theta J^{\mathrm{D}}_H(\theta_k)\|^2\le\big(4(J^\star-J^{\mathrm{D}}_H(\theta_0))+2L_J\sigma^2\big)/\sqrt N\), the standard rate for nonconvex stochastic gradient methods, and Algorithm~\ref{alg:codesign} is then an ordinary stochastic gradient method on the reduced co-design objective.

\begin{remark}[]
\label{rem:conv_envelope}
At the inner solution, errors in the policy sensitivity enter~\eqref{eq:conv_main_bound} only through \(\Lambda\delta_\kappa\), i.e.\ multiplied by \(\|\nabla_\phi J^{\mathrm{D}}_H(\phi^\star,\theta)\|\). Away from it they also enter through the tracking term, since \(\tilde\kappa\) is part of the mean field whose Lipschitz constant is \(L_g\); that contribution vanishes with the inner error. Under matched objectives with coinciding critics, Lemma~\ref{lem:envelope} gives \(\Lambda=0\), so damping, truncation, or an inexact solve for \(\kappa_\theta\) have no first-order effect at the inner solution, and the bias floor reduces to the tracking and semi-gradient terms. In general, \(\Lambda\) is zero when the policy-dependent parts of the two objectives coincide and is generically positive otherwise. It measures how much the design objective would still gain from moving the policy at the policy optimum, and it sets the accuracy that the sensitivity solve must deliver. This is the quantitative form of the rule of Section~\ref{sec:discussion} on when to compute the policy sensitivity.
\end{remark}

\begin{remark}[Projected update]
\label{rem:conv_projection}
The experiments keep the design in a box \(\Theta\) (Appendix~\ref{app:inv_spec}) and take the projected step \(\theta_{k+1}=\Pi_\Theta(\theta_k+\alpha\,\hat g_k)\). Since \(\Theta\) is compact and convex and \(\Pi_\Theta\) is nonexpansive, the iterates stay in \(\Theta\); this provides the design part of Assumption~\ref{ass:convergence}(e) and the compact set over which \(\Lambda\), \(c_V\) and \(J^\star\) are taken. Theorem~\ref{thm:convergence} is stated for the unconstrained step and does not carry over to the projected one unchanged. The stationarity measure for the projected step is the gradient mapping \(G_\alpha(\theta)\triangleq\alpha^{-1}\big(\Pi_\Theta(\theta+\alpha\nabla_\theta J^{\mathrm{D}}_H(\theta))-\theta\big)\), which equals the gradient where the projection is inactive. The analysis of projected stochastic gradient methods on smooth nonconvex objectives~\citep{ghadimi2016minibatch} bounds its average squared norm with the same optimization term and the same three bias terms as~\eqref{eq:conv_main_bound}, but with a variance term proportional to \(\sigma^2\) without the factor \(\alpha\). At an active bound the projection removes the noise that points out of the box and keeps the noise that points inward, so the noise does not average out; the variance floor is lowered by averaging more rollouts per estimate, not by a smaller step. In our experiments the lower bound was never active and the upper bounds were active on 9--18\% of the design coordinates.
\end{remark}

%% file: sections/appendix_implementation.tex
\section{Implementation Notes}
\label{app:implementation}

This appendix summarizes how the quantities of Sections~\ref{sec:codesign}--\ref{appendix:shac_sensitivity_full} are computed. Everything except the inner-solution sensitivity \(\kappa_\theta\) is a direct application of \acf{ad} to the differentiable rollout surrogate; \(\kappa_\theta\) additionally requires one linear solve.

\paragraph{Surrogate and estimation.}
The critic-bootstrapped surrogates \(J^{\pi}_H\)~\eqref{eq:Jsvg_inner} and \(J^{\mathrm{D}}_H\)~\eqref{eq:svg_objective} are estimated as sample means over \(E\) parallel rollouts. The exogenous noise \(\omega\) is drawn once per outer iteration and used by every term of that gradient, so all terms are derivatives of the same sampled rollouts. We write a single horizon \(H\) throughout for readability, but the outer design-return rollout~\eqref{eq:svg_objective} and the inner sensitivity solve~\eqref{eq:Jsvg_inner} that defines \(\kappa_\theta\) may use \emph{different} horizons in practice; each surrogate and its gradient are exact for its own horizon, and the two need not coincide. The semi-gradient case \(\partial_\theta V^{+}\equiv0\) (Corollaries~\ref{cor:shac_semigradient} and~\ref{cor:inner_semigradient}) holds by construction when the terminal critic is state-only (it carries no design argument); a \texttt{stop\_gradient} on the critic parameters additionally holds it fixed, so no separate implementation is needed to drop that term. Unless stated otherwise, the experiments reported in this paper use this \ac{shac} semi-gradient with a frozen, state-only terminal critic, so that the sensitivity critic is \(W_\zeta\equiv0\), \(\partial_\theta V^{+}\equiv0\), and the explicit critic--design term of Proposition~\ref{prop:svg_codesign_gradient} is used. The full \ac{svg} term can be beneficial when the terminal value's design dependence is significant (Corollary~\ref{cor:shac_semigradient}).

\paragraph{Matrix-free operators.}
\ac{ad} does not return the matrices \(A\) and \(B\) themselves but their transformation over a vector: each product in Table~\ref{tab:impl} (\(A\,v\), \(B\,v\), \(B^\top w\)) is a single forward- or reverse-mode pass over the rollout, costing a small multiple of one gradient evaluation, so the matrices never need to be stored. The actor Hessian \(A\) is symmetric but, at finite inner convergence, possibly indefinite (hence the damped solve~\eqref{eq:inner_ift_damped}). For moderate \(\dim\phi\), \(A\) can be formed from these products and factorized directly, the truncated eigendecomposition used in the inventory experiments (Appendix~\ref{app:example}); for larger policy networks, the solve runs matrix-free as a Krylov iteration on the products alone: conjugate gradients when \(\varepsilon I-A\) is positive definite, and a symmetric-indefinite (MINRES) or \ac{gmres}-type solver otherwise.

\paragraph{Outer-gradient construction.}
Write \(J^{\mathrm{D}}_H=\hat J^{\mathrm{D}}_H-\rho\), where \(\hat J^{\mathrm{D}}_H(\theta)=\mathbb{E}_{\omega_H}\!\big[\sum_{t=0}^{H-1}\gamma^t R^{\mathrm{D}}_\theta(s_t,a_t)+\gamma^H V^{\mathrm{D}}_\theta(s_H)\big]\) is the truncated \emph{return} surrogate that \ac{ad} differentiates and \(\rho(\theta)\) is the \(\theta\)-only design cost. Since \(\rho\) is a known closed-form function of \(\theta\), its gradient \(\nabla_\theta\rho\) is computed directly from that expression and added to the total gradient, instead of including \(\rho\) in the rollout return and backpropagating it through the simulator. The full co-design gradient is then the sum of three separately-accounted terms,
\begin{equation}
\label{eq:impl_outer_sum}
\nabla_\theta J^{\mathrm{D}}_H(\theta)
=
\underbrace{\partial_\theta \hat J^{\mathrm{D}}_H}_{\text{\ac{ad}, }\phi\text{ fixed}}
\;+\;
\underbrace{\kappa_\theta^\top\big(\nabla_\phi \hat J^{\mathrm{D}}_H\big)}_{\text{damped solve}~\eqref{eq:inner_ift_damped}}
\;-\;
\underbrace{\nabla_\theta\rho(\theta)}_{\text{design cost}},
\end{equation}
where the indirect term is evaluated as \(-B^\top(A-\varepsilon I)^{-1}\nabla_\phi \hat J^{\mathrm{D}}_H\) (with \(\nabla_\phi\rho=0\), so \(\nabla_\phi \hat J^{\mathrm{D}}_H=\nabla_\phi J^{\mathrm{D}}_H\)), so \(\kappa_\theta\) is never formed explicitly. The first two terms are the \ac{ad} direct and indirect effects of the chain rule~\eqref{eq:outer_grad_chainrule} restricted to the return \(\hat J^{\mathrm{D}}_H\); the design cost contributes the single additional term \(-\nabla_\theta\rho\), and is counted exactly once. By Lemma~\ref{lem:envelope}, the indirect term vanishes when the design and policy rewards coincide; it is generically nonzero, and potentially significant, when they differ.

\begin{table}[t]
\caption{Computation of each quantity. \ac{ad}: automatic differentiation of the rollout surrogate; HVP: Hessian--vector product. Only \(\kappa_\theta\) needs a solver beyond \ac{ad}.}
\label{tab:impl}
\centering
\begin{tabular}{lll}
\toprule
Quantity & Definition & Computed by \\
\midrule
Direct effect & \(\partial_\theta \hat J^{\mathrm{D}}_H\) (\(\phi\) fixed) & reverse-mode \ac{ad} in \(\theta\) \\
Inner gradient \(u\) & \(\nabla_\phi \hat J^{\mathrm{D}}_H\) & reverse-mode \ac{ad} in \(\phi\) \\
Residual \(F\) & \(\nabla_\phi J^{\pi}_H\) & reverse-mode \ac{ad} in \(\phi\) \\
\(A\,v\) & \(\nabla^2_{\phi\phi}J^{\pi}_H\,v\) & forward-over-reverse HVP (\ac{jvp} of \(F\)) \\
\(B\,v\) & \(\nabla^2_{\theta\phi}J^{\pi}_H\,v\) & \ac{jvp} of \(F\) in \(\theta\) \\
\(B^\top w\) & \((\nabla^2_{\theta\phi}J^{\pi}_H)^\top w\) & \ac{vjp} of \(F\) in \(\theta\) \\
\(\kappa_\theta^\top u\) & \(-B^\top A^{-1}u\) (exact) & damped solve \(-B^\top(A-\varepsilon I)^{-1}u\)~\eqref{eq:inner_ift_damped} \\
\bottomrule
\end{tabular}
\end{table}

%% file: sections/appendix_example.tex
\section{Experiment Details and Extended Results}
\label{app:example}

This appendix elaborates on Section~\ref{sec:benchmark_diffusion_codesign}. It gives the full specification of the inventory--routing example (Appendix~\ref{app:inv_spec}), the optimization protocols (Appendix~\ref{app:protocols}), and extended results and analyses (Appendices~\ref{app:ablation_tables}--\ref{app:rank}).

\subsection{Inventory--routing example}
\label{app:inv_spec}

\paragraph{Topology and dimensions.}
The network is a ring over \(N\) nodes (\(E=N\) directed edges), so \(n_\theta=\dim\theta=E+2N=3N\); the action dimension is also \(3N\) and the observation is the full state \(x_t\in\mathbb{R}^{3N}\). We use \(N\in\{5,10,20\}\) (\(n_\theta=15/30/60\)) for the misaligned-configuration experiments and \(N\in\{5,10,20,50\}\) for the comparison with \ac{bo} and \ac{es}.

\paragraph{Smoothing operators.}
The dynamics~\eqref{eq:inv_dynamics} and the outflow limit~\eqref{eq:inv_outflow_cap} use the softplus-smoothed positive part and smoothed two- and three-way minima,
\begin{equation}
\label{eq:inv_smoothing}
\lceil z\rceil_{+}=\tfrac{1}{k}\,\mathrm{softplus}(kz),
\qquad
\mathrm{smin}(u,v)=-\tfrac{1}{k}\,\mathrm{logsumexp}\!\big({-k}u,\,{-k}v\big),\qquad
\mathrm{smin}(u,v,w)=-\tfrac{1}{k}\,\mathrm{logsumexp}\!\big({-k}u,\,{-k}v,\,{-k}w\big),
\end{equation}
with sharpness \(k=5\) for both. These converge to \(\max(\cdot,0)\) and \(\min(\cdot,\cdot,\cdot)\) as \(k\to\infty\) while keeping the rollout smooth in \((x,a,\theta)\); the residual offset \(\log 2/k\approx0.14\) is small compared to typical stock levels. Each smooth minimum lies below the hard minimum by at most \(\ln 2/k\approx0.14\) (two arguments) or \(\ln 3/k\approx0.22\) (three), the gap being attained only when the arguments coincide and vanishing exponentially in their separation; since it never exceeds the hard minimum, \(m_i\) in~\eqref{eq:inv_dynamics} never exceeds the available stock and \(\eta_i\) in~\eqref{eq:inv_outflow_cap} never lets a node ship more than it holds. The undershoot has two consequences. \(m_i\) can be negative, by at most \(0.22\) and only when available stock, demand and service capacity are all near zero; a negative \(m_i\) raises \(q_i^{+}\) and \(y_i^{+}\) by at most that amount and lowers the revenue, so it is a bounded smoothing offset rather than a channel the policy can profit from. And \(\eta_i\) can dip to \(-\ln(1+e^{-k})/k\approx-1.3\cdot10^{-3}\) when the sender has no stock, which lets an edge carry a reverse flow of that relative size, so \(z_i\ge0\) holds up to a term of order \(10^{-3}\) of the neighbors' shipments, which the positive parts in~\eqref{eq:inv_dynamics} absorb.

\paragraph{Demand process and heterogeneity.}
Demand follows the softplus-AR(1) of Section~\ref{sec:benchmark_diffusion_codesign} with innovation scale \(\sigma_d=0.25\), and episodes are initialized from the stationary distribution. Each environment instance draws its parameters once per node: \(\mu_i\sim\mathcal{U}(0.2,2.0)\), \(\nu_i\sim\mathcal{U}(0.6,0.85)\), price \(p_i\sim\mathcal{U}(0.5,2.0)\), holding weight \(\lambda_{h,i}\sim\mathcal{U}(0.02,0.30)\), and capital-cost weights \(\alpha_e,\alpha^{s}_i,\alpha^{b}_i\sim\mathcal{U}(0.3,2.0)\). This heterogeneity makes both the optimal sizing and the optimal policy differ per node, so the design--policy coupling carries node-specific signal.

\paragraph{Objectives.}
The reward weights in~\eqref{eq:inv_rewards} are \(\lambda_a=0.05\) and \(\lambda_{\mathrm{cap}}=1.5\); episodes run \(T=200\) steps with \(\gamma=0.99\). The design box is \(c_e\in[0,2]\), \(s_i,b_i\in[0,4]\), initialized at \(c_e=1.0\), \(s_i=b_i=2.5\): capacities sized for mean rather than peak demand, so that prioritization during demand spikes is design-dependent.

\paragraph{Default and Misaligned environment configurations.}
The \emph{default} configuration is exactly as above with \(\lambda_y=0\): the policy-dependent parts of \(R^{\mathrm{D}}_\theta\) and \(R^{\pi}_\theta\) coincide, so by Lemma~\ref{lem:envelope} the policy-sensitivity term is zero and the comparison with \ac{bo} and \ac{es} isolates the direct term. The \emph{misaligned} configuration makes two changes. First, the  backlog penalty is added to the design objective (\(\lambda_y=0.1\)). Second, capacity is considered as operating costs rather than output scales: the action maps become \(u^{\mathrm{sup}}_i=\mathrm{softplus}(a^{\mathrm{sup}}_i)\), \(u^{\mathrm{srv}}_i=\mathrm{softplus}(a^{\mathrm{srv}}_i)\), \(f_e=a^{\mathrm{rt}}_e\) (the stock limit~\eqref{eq:inv_outflow_cap} on the outflows applies unchanged), and both rewards acquire the utilization cost
\begin{equation}
\label{eq:inv_util_cost}
-\,\rho_u\Big(\textstyle\sum_i \frac{(u^{\mathrm{sup}}_i)^2}{s_i+\epsilon}+\sum_i \frac{(u^{\mathrm{srv}}_i)^2}{b_i+\epsilon}+\sum_e \frac{f_e^{2}}{c_e+\epsilon}\Big),
\qquad \rho_u=0.5,\quad \epsilon=10^{-3},
\end{equation}
where the floor \(\epsilon\) keeps the cost finite at the lower edge of the design box. This removes the multiplicative redundancy of~\eqref{eq:inv_action_map}, in which a change of \(s_i\) can be exactly compensated inside the policy, and so strengthens the dependence of \(\phi^\star\) on \(\theta\). The trained controller deliberately under-serves on this configuration (served/demand \(0.67\)--\(0.85\)): backlog is accounted for only in \(R^{\mathrm{D}}_\theta\), so the backlog is part of the objective misalignment. The weight \(\rho_u\) controls this margin and simultaneously scales the misalignment, which bounds the regime in which the policy-sensitivity term matters.

\subsection{Optimization protocols and hyperparameters}
\label{app:protocols}

\paragraph{Inner loop.}
The policy is a Gaussian \ac{nn} with one hidden layer of 32 units, trained by \ac{shac} on \(R^{\pi}_\theta\) over 512 parallel environments with observation normalization (actor learning rate \(6.0\times10^{-4}\), critic learning rate \(7.5\times10^{-3}\), TD(\(\lambda\)) with \(\lambda=0.668\), gradient clip 0.458, 2 critic epochs; tuned once by random search on the default configuration). The first evaluation of a trial trains from scratch (4M environment steps at \(N\le10\), 5M at \(N=20\), 6M at \(N=50\)); every later design evaluation of the gradient method warm-starts from the current policy and critic (1.5M/2M/2.5M steps); \ac{bo}, \ac{es} and random search train a fresh policy from scratch at every query with the budget of the first evaluation, since their successive queries are unrelated designs and a warm start would make the oracle path-dependent. When a method starts from a design found by another optimizer, the policy is first re-converged at that design with 20 warm restarts, so that the score of the starting design is meaningful.

\paragraph{Outer loop (misaligned configuration).}
The outer optimizer is Adam with gradient-norm clip 30; the learning rate is \(0.02\) in the effect experiment and \(0.005\) in the refinement experiment (at \(0.02\) refinement is unstable for every variation). The direct design effect \(\partial_\theta J^{\mathrm{D}}\) is computed by reverse-mode \ac{ad} at fixed \(\phi^\star\).

\paragraph{Forming the policy-sensitivity term.}
At \(\dim\phi\approx1.5\)k--\(5.9\)k the policy Hessian is small enough to form densely, so the inventory experiments form \(\kappa_\theta^{\top}u\) by a rank-truncated eigendecomposition instead of the damped Krylov solve~\eqref{eq:inner_ift_damped}: symmetrize \(A\leftarrow\tfrac12(A+A^{\top})\), keep the \(k\) eigenpairs of largest magnitude, with \(k\) set to \(\approx3\%\) of \(\dim\phi\) (\(k=50/100/200\) at \(N=5/10/20\)), and apply the signed truncated inverse to \(u=\nabla_\phi J^{\mathrm{D}}\) before the final \(B^{\top}\)-\ac{vjp}. {Eigenvalues of either sign are retained and inverted with their own sign; only the near-zero ones, which fall outside the top \(k\) by magnitude, are discarded, and no damping \(\varepsilon\) is applied on top. The rank is fixed as a fraction of \(\dim\phi\) rather than as a constant, so that the captured share of the spectrum does not shrink as the policy grows; we did not sweep this fraction in the reported runs.} The matrices \(A\), \(B\) and the vector \(u\) are all formed at the operating point of the trained policy: the sensitivity rollouts feed the policy the same normalized observations that \ac{shac} trains it on and that \(J^{\mathrm{D}}\) evaluates it on, with the running normalizer frozen at the current statistics and applied once (the critic likewise). The truncation plays the same regularizing role as the damping \(\varepsilon\) in~\eqref{eq:inner_ift_damped} for an indefinite, imperfectly converged \(A\).  The direct-only variation sets \(\kappa_\theta=0\) exactly; it is \emph{not} the \ac{shac} semi-gradient of Corollary~\ref{cor:shac_semigradient}, which would retain \(\kappa_\theta\) inside the horizon.

\paragraph{Comparison protocol (Figure~\ref{fig:benchmark}, matched objectives).}
This is the comparison with \ac{bo} and \ac{es} of Section~\ref{sec:benchmark_diffusion_codesign}, run in the default configuration (\(\lambda_y=0\), where the policy-sensitivity term vanishes by Lemma~\ref{lem:envelope}). Three methods optimize the same \(J^{\mathrm{D}}\) with the same scoring rollout (200 steps, 32 environments) and the inner loop of the previous paragraph, under its warm (gradient method) and cold (\ac{bo}, \ac{es}) policy budgets, and an identical budget of 100 design evaluations per trial: the gradient method (direct design effect only); \ac{bo} with a Gaussian-process surrogate and expected improvement \citep{balandat2020botorch}; and a \((1{+}1)\)-\ac{es} with the 1/5th success rule~\citep{hansen2015evolution} (\(\sigma_0=0.15\) of the box). Trials are 5 environment instances \(\times\) 5 seeds per method and \(N\). The numbers reported are the gaps in the final co-design performance \(J^{\mathrm{D}}\) after the 100 evaluations, gradient method relative to \ac{bo} or \ac{es}, paired within each environment instance, which removes environment-difficulty variance (pooled standard deviations are \(5\)--\(10\times\) larger): \(+5.1\%/{+}8.8\%/{+}9.3\%/{+}9.9\%\) over \ac{bo} and \(+7.7\%/{+}11.5\%/{+}9.0\%/{+}10.0\%\) over \ac{es} at \(N=5/10/20/50\), positive in every instance. Figure~\ref{fig:benchmark} instead plots the best \(J^{\mathrm{D}}\) found so far minus the \(J^{\mathrm{D}}\) of the shared initial design, averaged over seeds and then over instances, which removes the environment-difficulty offset from the curves; at 100 evaluations these gains are \(+146/{+}312/{+}399/{+}955\) for the gradient method against \(+103/{+}129/{+}57/{+}95\) for \ac{bo} and \(+77/{+}82/{+}65/{+}82\) for \ac{es} at \(N=5/10/20/50\), where the initial designs score \(809/1738/3620/8646\). The percentages above are gaps in the absolute final \(J^{\mathrm{D}}\); the figure shows gains over the initial design.

\paragraph{{Computational cost (Table~\ref{tab:cost}).}}
All rows were measured on one RTX~5080 with one job at a time, each method under its own protocol: \ac{bo} and \ac{es} over the 100 evaluations of the comparison protocol (mean over evaluations 2--100; \ac{bo}'s cost grows with its surrogate, 3.4\,s over the last 50 evaluations against 3.2\,s over all of them at \(N=20\)), and the gradient variations over 30 iterations of the effect protocol (iterations 10--30, so compilation is excluded). Every evaluation scores the design with the same 200-step rollout over 32 environments. The black-box methods train a fresh policy at every query (4M/4M/5M steps at \(N=5/10/20\)), the gradient variations warm-start (1.5M/1.5M/2M steps); training is launch-bound on the GPU at these sizes, so this difference is worth about 0.2\,s per evaluation. Every program was compiled once, into a fresh compilation cache at the start of the pass, and that executable served every row; with XLA's autotuning the same program can compile to executables whose run time differs by up to \(2\times\) on these overhead-dominated rollouts, so the absolute times carry that uncertainty while the ratios between rows, all measured under one compilation, do not. The dense variant forms \(A\) by one Hessian-vector product per policy parameter (\(\dim\phi=1502/2972/5912\)) before the truncated eigendecomposition, and its cost grows accordingly; the matrix-free variant runs MINRES on Hessian-vector products~\eqref{eq:inner_ift_damped} without damping, stopped after 200 matrix--vector products, and never forms \(A\); the cap binds at every iteration (relative residual 3--9\%), so its direction is a truncated Krylov iterate, which regularizes the indefinite system in the same way the rank truncation does. Run on 5 trials per size with the same protocol as Figure~\ref{fig:effect}, the matrix-free variant's gain over direct-only at the common stopping point (iteration 200) is \(+172/+418/+715\) against \(+177/+428/+974\) for the dense variant (5/5 trials each): equal up to \(N=10\), and \(-259\pm109\) (2/5 trials) at \(N=20\), where the undamped solve on the indefinite \(A\) is the less accurate of the two.

\subsection{Extended tables and per-trial results}
\label{app:ablation_tables}

\begin{figure}[t]
\centering
\includegraphics{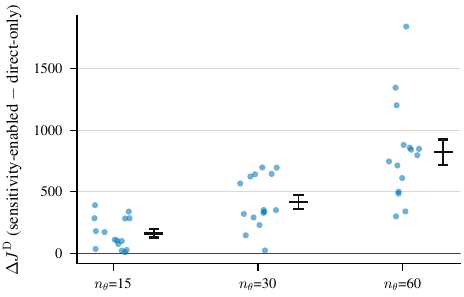}
\caption{\textbf{Per-trial paired differences} at the common stopping point (iteration 200): \(\Delta J^{\mathrm{D}}\) (sensitivity-enabled \(-\) direct-only, \(k=50/100/200\)); dots are the 15 trials per \(N\), black markers the mean \(\pm\) SE.}
\label{fig:ablation_cells}
\end{figure}

\paragraph{Scoring of the policy sensitivity effect experiment.}
The metric for the policy sensitivity effect experiment (Figure~\ref{fig:effect}) was fixed before the runs: both variations are stopped at the same iteration, 200, and scored by the mean \(J^{\mathrm{D}}\) over the last 10 iterations. Under this metric the paired contrasts (sensitivity-enabled \(-\) direct-only) are \(+161.5\pm32.7\), \(+416.9\pm55.0\) and \(+821.3\pm103.5\) at \(N=5/10/20\), each positive in 15/15 trials (\(p<10^{-3}\)); the gains over the branch point are \(+356\), \(+544\) and \(+1259\) for the {sensitivity-enabled gradient} against \(+195\), \(+127\) and \(+438\) for direct-only. The contrasts are as large by the plateau score (median of iterations 60--150 minus the branch point: \(+172\), \(+481\), \(+671\); 15/15 each) and by the rolling maximum (\(+159\), \(+393\), \(+813\); 15/15). Figure~\ref{fig:ablation_cells} shows the per-trial differences.

\paragraph{Scoring of the refinement experiment.}
Table~\ref{tab:refinement_stops} reports the refinement experiment at the stopping points displayed in Figure~\ref{fig:refinement}, 50, 100 and 150 iterations for \(N=5\), \(10\) and \(20\); larger \(N\) are given more iterations because their refinement takes longer to converge. The score is the mean \(J^{\mathrm{D}}\) of the last 10 iterations up to the stopping point, minus the \(J^{\mathrm{D}}\) of the starting design. In each trial, both variations start from the same point, the best design found by \ac{es} in that trial, with the policy re-converged at it over 20 warm restarts and a freshly initialized outer optimizer.

\begin{table}[t]
\caption{\textbf{Refinement from the best \ac{es} design} (misaligned configuration, \(k\approx3\%\) of \(\dim\phi\)), scored at the stopping point of Figure~\ref{fig:refinement}. Gain = \(J^{\mathrm{D}}\) improvement over the starting \ac{es} design for the sensitivity-enabled and the direct-only variation (in brackets, as \% of the mean \ac{es} performance, and wins over the 15 trials); contrast = sensitivity-enabled \(-\) direct-only, paired over the 15 trials.}
\label{tab:refinement_stops}
\centering
\footnotesize
\begin{tabular}{ccrrr}
\toprule
& iterations & sensitivity-enabled & direct-only & \multicolumn{1}{c}{difference} \\
\midrule
\(N=5\), \(k=50\)   & 50  & \(+148.3\) (\(65\%\), 15/15) & \(+91.2\) (\(40\%\)) & \(+57.1\pm23.0\) (13/15, \(p=0.026\)) \\
\(N=10\), \(k=100\) & 100 & \(+422.9\) (\(58\%\), 15/15) & \(+255.6\) (\(35\%\)) & \(+167.3\pm58.6\) (15/15, \(p=0.013\)) \\
\(N=20\), \(k=200\) & 150 & \(+1772.8\) (\(99\%\), 15/15) & \(+1396.3\) (\(78\%\)) & \(+376.5\pm128.9\) (13/15, \(p=0.011\)) \\
\bottomrule
\end{tabular}
\end{table}

\subsection{The truncation rank}
\label{app:rank}

The truncation rank is a size-dependent setting: \(k=50\) captures 3.3\% of the spectrum at \(N=5\) (\(\dim A=1502\)) but 0.85\% at \(N=20\) (\(\dim A=5912\)). We therefore set \(k\) to \(\approx3\%\) of \(\dim\phi\) (\(50/100/200\)); on three \(N=20\) trials in the effect setting, raising \(k\) from 50 to 200 adds \(+664\) to the performance difference at \(N=20\).

%% file: main.bbl
\begin{thebibliography}{36}
\providecommand{\natexlab}[1]{#1}
\providecommand{\url}[1]{\texttt{#1}}
\expandafter\ifx\csname urlstyle\endcsname\relax
  \providecommand{\doi}[1]{doi: #1}\else
  \providecommand{\doi}{doi: \begingroup \urlstyle{rm}\Url}\fi

\bibitem[Ajalloeian \& Stich(2020)Ajalloeian and Stich]{ajalloeian2020convergence}
Ahmad Ajalloeian and Sebastian~U. Stich.
\newblock On the convergence of {SGD} with biased gradients.
\newblock \emph{arXiv preprint arXiv:2008.00051}, 2020.

\bibitem[Anand et~al.(2026)Anand, Sawant, Parmas, Hoffmann, Reinhardt, and Gros]{anand2026bilevel}
Akhil~S Anand, Shambhuraj Sawant, Paavo Parmas, Jasper Hoffmann, Dirk Reinhardt, and Sebastien Gros.
\newblock Bi-level reinforcement learning pathway for sim-to-real optimality.
\newblock \emph{Reinforcement Learning Journal}, 7, 2026.

\bibitem[Balandat et~al.(2020)Balandat, Karrer, Jiang, Daulton, Letham, Wilson, and Bakshy]{balandat2020botorch}
Maximilian Balandat, Brian Karrer, Daniel~R. Jiang, Samuel Daulton, Benjamin Letham, Andrew~Gordon Wilson, and Eytan Bakshy.
\newblock {BoTorch}: A framework for efficient {Monte-Carlo} {Bayesian} optimization.
\newblock In \emph{Advances in Neural Information Processing Systems}, volume~33, 2020.

\bibitem[Bhatia et~al.(2021)Bhatia, Jackson, Tian, Xu, and Matusik]{bhatia2021evolutiongym}
Jagdeep Bhatia, Holly Jackson, Yunsheng Tian, Jie Xu, and Wojciech Matusik.
\newblock Evolution gym: A large-scale benchmark for evolving soft robots.
\newblock \emph{Advances in Neural Information Processing Systems}, 34:\penalty0 2201--2214, 2021.

\bibitem[Blondel et~al.(2022)Blondel, Berthet, Cuturi, Frostig, Hoyer, Llinares-L{\'o}pez, Pedregosa, and Vert]{blondel2022implicit}
Mathieu Blondel, Quentin Berthet, Marco Cuturi, Roy Frostig, Stephan Hoyer, Felipe Llinares-L{\'o}pez, Fabian Pedregosa, and Jean-Philippe Vert.
\newblock Efficient and modular implicit differentiation.
\newblock \emph{Advances in neural information processing systems}, 35:\penalty0 5230--5242, 2022.

\bibitem[Borkar(2008)]{borkar2008stochastic}
Vivek~S Borkar.
\newblock \emph{Stochastic approximation: a dynamical systems viewpoint}, volume~2.
\newblock Springer, 2008.

\bibitem[Bottou et~al.(2018)Bottou, Curtis, and Nocedal]{bottou2018optimization}
L{\'e}on Bottou, Frank~E. Curtis, and Jorge Nocedal.
\newblock Optimization methods for large-scale machine learning.
\newblock \emph{SIAM Review}, 60\penalty0 (2):\penalty0 223--311, 2018.

\bibitem[Chen et~al.(2021)Chen, Sun, and Yin]{chen2021closing}
Tianyi Chen, Yuejiao Sun, and Wotao Yin.
\newblock Closing the gap: Tighter analysis of alternating stochastic gradient methods for bilevel problems.
\newblock In \emph{Advances in Neural Information Processing Systems}, volume~34, pp.\  25294--25307, 2021.

\bibitem[Cheney et~al.(2014)Cheney, MacCurdy, Clune, and Lipson]{cheney2014unshackling}
Nick Cheney, Robert MacCurdy, Jeff Clune, and Hod Lipson.
\newblock Unshackling evolution: evolving soft robots with multiple materials and a powerful generative encoding.
\newblock \emph{ACM SIGEVOlution}, 7\penalty0 (1):\penalty0 11--23, 2014.

\bibitem[Dai et~al.(2026)Dai, Wang, Ashley, and Schmidhuber]{dai2026stackelberg}
Yanning Dai, Yuhui Wang, Dylan~R. Ashley, and J{\"u}rgen Schmidhuber.
\newblock Efficient morphology-control co-design via stackelberg proximal policy optimization.
\newblock In \emph{The Fourteenth International Conference on Learning Representations}, 2026.

\bibitem[Freeman et~al.(2021)Freeman, Frey, Raichuk, Girgin, Mordatch, and Bachem]{freeman2021brax}
C~Daniel Freeman, Erik Frey, Anton Raichuk, Sertan Girgin, Igor Mordatch, and Olivier Bachem.
\newblock Brax--a differentiable physics engine for large scale rigid body simulation.
\newblock In \emph{Thirty-fifth Conference on Neural Information Processing Systems Datasets and Benchmarks Track (Round 1)}, 2021.

\bibitem[Ghadimi \& Lan(2013)Ghadimi and Lan]{ghadimi2013stochastic}
Saeed Ghadimi and Guanghui Lan.
\newblock Stochastic first- and zeroth-order methods for nonconvex stochastic programming.
\newblock \emph{SIAM Journal on Optimization}, 23\penalty0 (4):\penalty0 2341--2368, 2013.

\bibitem[Ghadimi \& Wang(2018)Ghadimi and Wang]{ghadimi2018approximation}
Saeed Ghadimi and Mengdi Wang.
\newblock Approximation methods for bilevel programming.
\newblock \emph{arXiv preprint arXiv:1802.02246}, 2018.

\bibitem[Ghadimi et~al.(2016)Ghadimi, Lan, and Zhang]{ghadimi2016minibatch}
Saeed Ghadimi, Guanghui Lan, and Hongchao Zhang.
\newblock Mini-batch stochastic approximation methods for nonconvex stochastic composite optimization.
\newblock \emph{Mathematical Programming}, 155\penalty0 (1-2):\penalty0 267--305, 2016.

\bibitem[Grazzi et~al.(2020)Grazzi, Franceschi, Pontil, and Salzo]{grazzi2020iteration}
Riccardo Grazzi, Luca Franceschi, Massimiliano Pontil, and Saverio Salzo.
\newblock On the iteration complexity of hypergradient computation.
\newblock In \emph{International Conference on Machine Learning}, pp.\  3748--3758. PMLR, 2020.

\bibitem[Gupta et~al.(2021)Gupta, Savarese, Ganguli, and Fei-Fei]{gupta2021derl}
Agrim Gupta, Silvio Savarese, Surya Ganguli, and Li~Fei-Fei.
\newblock Embodied intelligence via learning and evolution.
\newblock \emph{Nature communications}, 12\penalty0 (1):\penalty0 5721, 2021.

\bibitem[Hansen et~al.(2015)Hansen, Arnold, and Auger]{hansen2015evolution}
Nikolaus Hansen, Dirk~V Arnold, and Anne Auger.
\newblock Evolution strategies.
\newblock In \emph{Springer Handbook of Computational Intelligence}, pp.\  871--898. Springer, 2015.

\bibitem[Heess et~al.(2015)Heess, Wayne, Silver, Lillicrap, Erez, and Tassa]{heess2015svg}
Nicolas Heess, Gregory Wayne, David Silver, Timothy Lillicrap, Tom Erez, and Yuval Tassa.
\newblock Learning continuous control policies by stochastic value gradients.
\newblock \emph{Advances in neural information processing systems}, 28, 2015.

\bibitem[Herber \& Allison(2019)Herber and Allison]{herber2017nested}
Daniel~R Herber and James~T Allison.
\newblock Nested and simultaneous solution strategies for general combined plant and control design problems.
\newblock \emph{Journal of Mechanical Design}, 141\penalty0 (1):\penalty0 011402, 2019.

\bibitem[Hong et~al.(2023)Hong, Wai, Wang, and Yang]{hong2023twotimescale}
Mingyi Hong, Hoi-To Wai, Zhaoran Wang, and Zhuoran Yang.
\newblock A two-timescale stochastic algorithm framework for bilevel optimization: Complexity analysis and application to actor-critic.
\newblock \emph{SIAM Journal on Optimization}, 33\penalty0 (1):\penalty0 147--180, 2023.

\bibitem[Hu et~al.(2020)Hu, Anderson, Li, Sun, Carr, Ragan-Kelley, and Durand]{hu2020difftaichi}
Yuanming Hu, Luke Anderson, Tzu-Mao Li, Qi~Sun, Nathan Carr, Jonathan Ragan-Kelley, and Fr{\'e}do Durand.
\newblock Difftaichi: Differentiable programming for physical simulation.
\newblock \emph{ICLR}, 2020.

\bibitem[Jang et~al.(2017)Jang, Gu, and Poole]{jang2017categorical}
Eric Jang, Shixiang Gu, and Ben Poole.
\newblock Categorical reparameterization with {Gumbel-Softmax}.
\newblock In \emph{International Conference on Learning Representations (ICLR)}, 2017.

\bibitem[Krantz \& Parks(2002)Krantz and Parks]{krantz2002implicit}
Steven~George Krantz and Harold~R Parks.
\newblock \emph{The implicit function theorem: history, theory, and applications}, volume 202.
\newblock Springer, 2002.

\bibitem[Lee et~al.(1997)Lee, Padmanabhan, and Whang]{lee1997information}
Hau~L Lee, Venkata Padmanabhan, and Seungjin Whang.
\newblock Information distortion in a supply chain: The bullwhip effect.
\newblock \emph{Management science}, 43\penalty0 (4):\penalty0 546--558, 1997.

\bibitem[Lorraine et~al.(2020)Lorraine, Vicol, and Duvenaud]{lorraine2020optimizing}
Jonathan Lorraine, Paul Vicol, and David Duvenaud.
\newblock Optimizing millions of hyperparameters by implicit differentiation.
\newblock In \emph{International conference on artificial intelligence and statistics}, pp.\  1540--1552. PMLR, 2020.

\bibitem[Milgrom \& Segal(2002)Milgrom and Segal]{milgrom2002envelope}
Paul Milgrom and Ilya Segal.
\newblock Envelope theorems for arbitrary choice sets.
\newblock \emph{Econometrica}, 70\penalty0 (2):\penalty0 583--601, 2002.

\bibitem[Mohamed et~al.(2020)Mohamed, Rosca, Figurnov, and Mnih]{mohamed2020monte}
Shakir Mohamed, Mihaela Rosca, Michael Figurnov, and Andriy Mnih.
\newblock Monte carlo gradient estimation in machine learning.
\newblock \emph{Journal of Machine Learning Research}, 21\penalty0 (132):\penalty0 1--62, 2020.

\bibitem[Nikishin et~al.(2022)Nikishin, Abachi, Agarwal, and Bacon]{nikishin2022control}
Evgenii Nikishin, Romina Abachi, Rishabh Agarwal, and Pierre-Luc Bacon.
\newblock Control-oriented model-based reinforcement learning with implicit differentiation.
\newblock In \emph{Proceedings of the AAAI Conference on Artificial Intelligence}, volume~36, pp.\  7886--7894, 2022.

\bibitem[Parmas et~al.(2018)Parmas, Rasmussen, Peters, and Doya]{parmas2018pipps}
Paavo Parmas, Carl~Edward Rasmussen, Jan Peters, and Kenji Doya.
\newblock Pipps: Flexible model-based policy search robust to the curse of chaos.
\newblock In \emph{International Conference on Machine Learning}, pp.\  4065--4074. PMLR, 2018.

\bibitem[Rajeswaran et~al.(2019)Rajeswaran, Finn, Kakade, and Levine]{rajeswaran2019imaml}
Aravind Rajeswaran, Chelsea Finn, Sham~M Kakade, and Sergey Levine.
\newblock Meta-learning with implicit gradients.
\newblock \emph{Advances in neural information processing systems}, 32, 2019.

\bibitem[Schaff et~al.(2019)Schaff, Yunis, Chakrabarti, and Walter]{schaff2019jointly}
Charles Schaff, David Yunis, Ayan Chakrabarti, and Matthew~R Walter.
\newblock Jointly learning to construct and control agents using deep reinforcement learning.
\newblock In \emph{2019 international conference on robotics and automation (ICRA)}, pp.\  9798--9805. IEEE, 2019.

\bibitem[Spielberg et~al.(2019)Spielberg, Zhao, Hu, Du, Matusik, and Rus]{spielberg2019learning}
Andrew Spielberg, Allan Zhao, Yuanming Hu, Tao Du, Wojciech Matusik, and Daniela Rus.
\newblock Learning-in-the-loop optimization: End-to-end control and co-design of soft robots through learned deep latent representations.
\newblock \emph{Advances in Neural Information Processing Systems}, 32, 2019.

\bibitem[Sutton \& Barto(2018)Sutton and Barto]{sutton2018reinforcement}
Richard~S Sutton and Andrew~G Barto.
\newblock \emph{Reinforcement learning: An introduction}.
\newblock MIT press, 2018.

\bibitem[Tobin \& Friesz(1988)Tobin and Friesz]{tobin1988sensitivity}
Roger~L Tobin and Terry~L Friesz.
\newblock Sensitivity analysis for equilibrium network flow.
\newblock \emph{Transportation Science}, 22\penalty0 (4):\penalty0 242--250, 1988.

\bibitem[Xu et~al.(2022)Xu, Makoviychuk, Narang, Ramos, Matusik, Garg, and Macklin]{xu2022shac}
Jie Xu, Viktor Makoviychuk, Yashraj Narang, Fabio Ramos, Wojciech Matusik, Animesh Garg, and Miles Macklin.
\newblock Accelerated policy learning with parallel differentiable simulation.
\newblock In \emph{International Conference on Learning Representations}, 2022.

\bibitem[Yuan et~al.(2022)Yuan, Song, Luo, Sun, and Kitani]{yuan2022transform2act}
Ye~Yuan, Yuda Song, Zhengyi Luo, Wen Sun, and Kris Kitani.
\newblock Transform2act: Learning a transform-and-control policy for efficient agent design.
\newblock In \emph{International Conference on Learning Representations}, 2022.

\end{thebibliography}
